\documentclass[10pt,twocolumn,letterpaper]{article}

\usepackage{authblk}
\usepackage[pagenumbers]{cvpr} 
\usepackage[accsupp]{axessibility}  

\usepackage{amsmath}
\usepackage{amssymb}
\usepackage{bm}
\usepackage{multirow}
\usepackage[ruled,vlined]{algorithm2e}
\usepackage{floatrow}

\usepackage{booktabs}
\usepackage{xspace}

\usepackage[pagebackref,breaklinks,colorlinks]{hyperref}
\usepackage[capitalise]{cleveref}
\usepackage{url}

\usepackage{graphicx}
\usepackage{lipsum}
\usepackage{microtype}
\usepackage{nicefrac}
\usepackage{verbatim}
\usepackage{bbding}

\makeatletter
\DeclareRobustCommand\onedot{\futurelet\@let@token\@onedot}
\def\@onedot{\ifx\@let@token.\else.\null\fi\xspace}
\def\eg{\emph{e.g}\onedot} 
\def\ie{\emph{i.e}\onedot}

\def\etal{\emph{et al}\onedot}
\renewcommand{\paragraph}{%
	\@startsection{paragraph}{4}{\z@}%
	{0.1em \@plus 0.5ex \@minus 0.2ex}{-1em}%
	{\normalsize\bf}%
}
\newcommand\rurl[1]{%
  \href{https://#1}{\nolinkurl{#1}}%
}
\makeatother

\newcommand{\bx}{\bm{x}}

\newcommand{\bh}{\bm{h}}

\crefname{section}{Sec.}{Secs.}
\Crefname{section}{Section}{Sections}
\Crefname{table}{Table}{Tables}
\crefname{table}{Tab.}{Tabs.}

\begin{document}

\title{Cross-domain Few-shot Learning with Task-specific Adapters}

\author[]{\vspace{-0.3cm}Wei-Hong Li}
\author[]{Xialei Liu\thanks{Xialei Liu is the corresponding author.}}
\author[]{Hakan Bilen\vspace{-0.25cm}}

\affil[]{VICO Group, University of Edinburgh, United Kingdom\vspace{-0.25cm}}
\affil[]{\small \rurl{github.com/VICO-UoE/URL}\vspace{-0.3cm}}

\maketitle

\begin{abstract}
    In this paper, we look at the problem of cross-domain few-shot classification that aims to learn a classifier from previously unseen classes and domains with few labeled samples. 
    Recent approaches broadly solve this problem by parameterizing their few-shot classifiers with task-agnostic and task-specific weights where the former is typically learned on a large training set and the latter is dynamically predicted through an auxiliary network conditioned on a small support set. 
    In this work, we focus on the estimation of the latter, and propose to learn task-specific weights from scratch directly on a small support set, in contrast to dynamically estimating them.
    In particular, through systematic analysis, we show that task-specific weights through parametric adapters in matrix form with residual connections to multiple intermediate layers of a backbone network significantly improves the performance of the state-of-the-art models in the Meta-Dataset benchmark with minor additional cost.
\end{abstract}

\section{Introduction}\label{sec:intro}

Deep learning methods have seen remarkable progress in various fields where large quantities of data and compute power are available.
However, the ability of deep networks to learn new concepts from small data remains limited.
Few-shot classification~\cite{lake2011one,miller2000learning} is inspired from this limitation and aims at learning a model that can be efficiently adapted to recognize unseen classes from few samples.
In particular, the standard setting for learning few-shot classifiers involves two stages: (i) learning a model, typically from a large training set, (ii) adapting this model to learn new classes from a given small support set. 
These two stages are called meta-training and meta-testing respectively.
The adapted model is finally evaluated on a query set where the task is to assign each query sample to one of the classes in the support set.

\begin{figure}
\begin{center}
\includegraphics[width=0.95\linewidth]{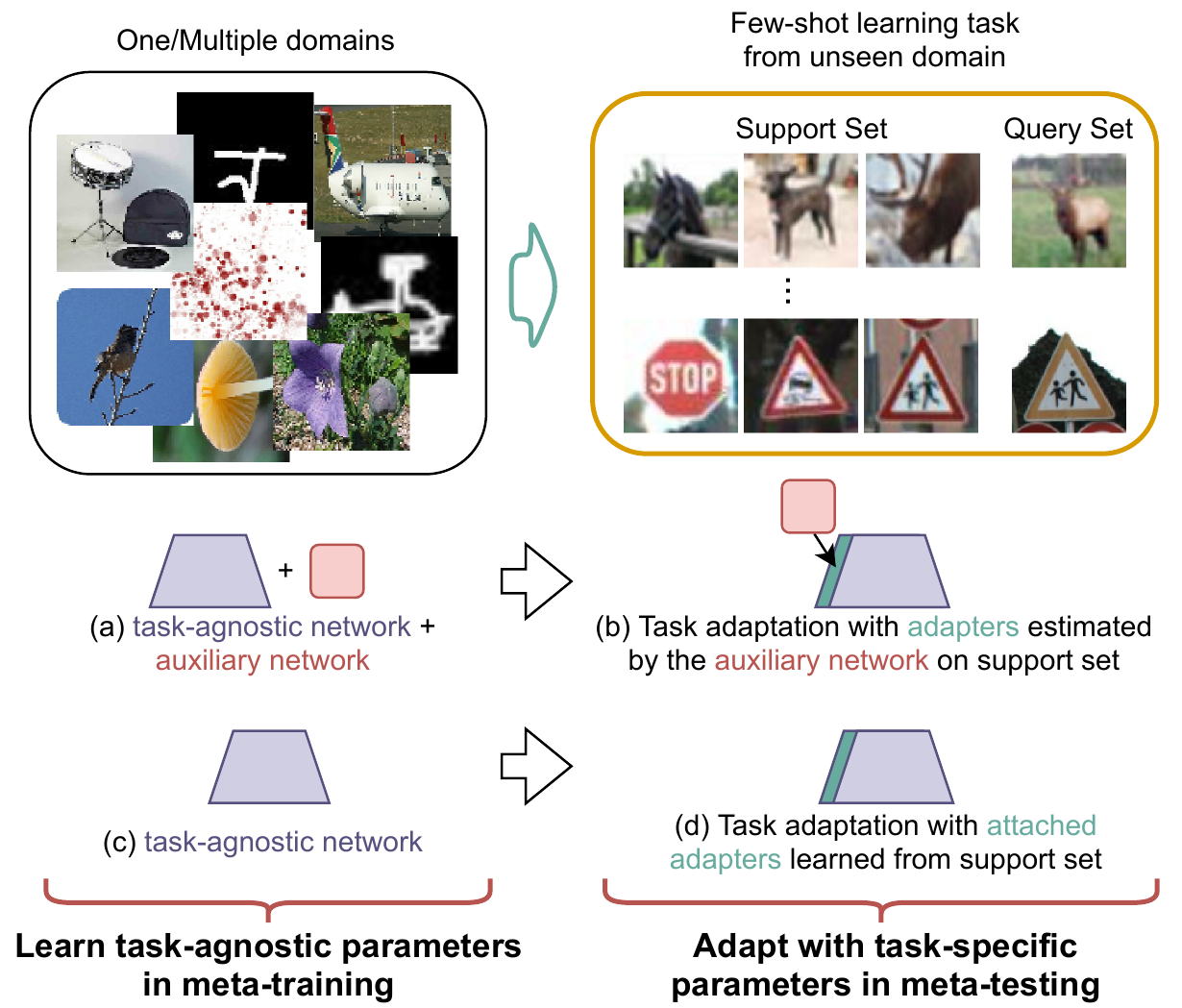}
\end{center}
\vspace{-0.45cm}
\caption{\textbf{Cross-domain Few-shot Learning} considers to learn a model from one or multiple domains to generalize to unseen domains with few samples. Prior works often learn a task-agnostic model with an auxiliary network during meta-training (a) and a set of adapters are generated by the auxiliary network to adapt to the given support set (b). While in this work, we propose to attach adapters directly to a pretrained task-agnostic model (c), which can be estimated from scratch during meta-testing (d). We also propose different architecture topologies of adapters and their efficient approximations. 
}
\label{fig:cdfsl}
\end{figure}

Early methods~\cite{vinyals2016matching,ravi2016optimization,finn2017model,oreshkin2018tadam,rusu2018meta,snell2017prototypical} pose the few-shot classification problem in a learning-to-learn formulation by training a deep network over a distribution of related tasks, which are sampled from the training set, and transfer this experience to improve its performance for learning new classes.
Concretely, Vinyals~\etal~\cite{vinyals2016matching} learn a feature encoder that is conditioned on the support set in meta-training and does not require any further training in meta-test thanks to its non-parametric classifier.
Ravi and Larochelle~\cite{ravi2016optimization} take the idea of learning a feature encoder in meta-train further by also learning an update rule through an LSTM that produces the updates for a classifier in meta-test.
Finn~\etal~\cite{finn2017model} pose the task as a meta-learning problem and learn the parameters of a deep network in meta-training such that a network initialized with the learned parameters can be efficiently finetuned on a new task.
We refer to \cite{wang2020generalizing, hospedales2020meta} for comprehensive review of early works.

Despite the significant progress, the scope of the early methods has been limited to a restrictive setting where training and test samples come from a single domain (or data distribution) such as Omniglot~\cite{Lake1332}, miniImageNet~\cite{vinyals2016matching} and tieredImageNet~\cite{ren2018meta}.
They perform poorly in the more challenging cross-domain few-shot tasks, where test data is sampled from an unknown or previously unseen domain~\cite{triantafillou2019meta}.
This setting poses an additional learning challenge, not only requires leveraging the limited information from the small support set for learning the target task but also \emph{selectively transferring relevant knowledge} from previously seen domains to the target task.

Broadly, recent approaches address this challenge by parameterizing deep networks with a large set of task-agnostic and a small set of task-specific weights that encode generic representations valid for multiple tasks and private representations are specific to the target task respectively.
While the task-agnostic weights are learned over multiple tasks, typically, from a large dataset in meta-training, the task-specific weights are estimated from a given small support set (\eg 5 images per category)~\cite{requeima2019fast,bateni2020improved,lee2019meta,dvornik2020selecting,liu2020universal,li2021universal,triantafillou2021flute}.
In the literature, the task-agnostic weights are used to parameterize a single network that is trained on large data from one domain~\cite{requeima2019fast,bateni2020improved,doersch2020crosstransformers} or on multiple domains~\cite{li2021universal}, or to be distributed over multiple networks, each trained on a different domain~\cite{dvornik2020selecting,liu2020universal,triantafillou2021flute} \footnote{Note that the task-agnostic weights can also be finetuned on the target task (\eg \cite{chen2020new,dhillon2019baseline}).}.
The task-specific weights are utilized to parameterize a linear classifier~\cite{lee2019meta}, a pre-classifier feature mapping~\cite{li2021universal} and an ensemble of classifiers at each layer of a deep neural network~\cite{adler2020cross}.

Recently, inspired from \cite{perez2018film}, \emph{task-specific adapters}~\cite{requeima2019fast,bateni2020improved}, small capacity transformations that are applied to multiple layers of a deep network, have been successfully used to steer the few-shot classifiers to new tasks and domains.
Their weights are often estimated dynamically through an auxiliary network conditioned on the support set~\cite{requeima2019fast,bateni2020improved,liu2020universal,triantafillou2021flute} (see \cref{fig:cdfsl}.(a,b)), in a similar spirit to~\cite{bertinetto2016learning,jia2016dynamic}.
As the auxiliary network is trained on multiple tasks in meta-training, the premise of estimating the task-specific adapter weights with it is based on the principle of transfer learning such that it can transfer the knowledge from the previous tasks to better estimate them for unseen tasks.
However, learning an accurate auxiliary network is a challenging task due to two reasons.
First, it has to generalize to previously unseen tasks and especially to significantly different unseen domains.
Second, learning to predict high-dimensional weights where each corresponds to a dimension of a highly nonlinear feature space is a difficult learning problem too.

Motivated by this shortcoming, as shown in \cref{fig:cdfsl}, we propose to employ a set of light-weight task-specific adapters along with the task-agnostic weights for adapting the few-shot classifier to the tasks from unseen domains.
Unlike the prior work, we learn the weights of these adapters from scratch by directly optimizing them on a small support set (see \cref{fig:cdfsl}.(c,d)).
Moreover, we systematically study various combinations of several design choices for task-specific adaptation, which have not been explored before, including adapter connection types (serial or residual), parameterizations (matrix and its decomposed variations, channelwise operations) and estimation of task-specific parameters.
Extensive experiments demonstrate that attaching parameteric adapters in matrix form to convolutional layers with residual connections significantly boosts the state-of-the-art performance in most domains, especially resulting in superior performance in unseen domains on Meta-Dataset with negligible increase in computations.

\paragraph{More related work.} 
Here we provide more detailed discussion of the most related work.
Both CNAPS~\cite{requeima2019fast} and Simple CNAPS~\cite{bateni2020improved} employ task-specific adapters via FiLM layers (which uses a channelwise affine transformation and connected to the backbone in a serial way)~\cite{perez2018film} to adapt their feature extractors to the target task and estimate them via an auxiliary network. 
Compared to them, we propose learning residual adapters in matrix form directly on the support set.
SUR~\cite{dvornik2020selecting} and URT~\cite{liu2020universal} learn an attention mechanism to select/fuse features from multiple domain-specific models in meta-train respectively. 
As we build on a single multi-domain feature extractor, our method does not require such attention but we attach task-specific adapters to the feature extractor to adapt the features to unseen tasks.
URL~\cite{li2021universal} learns a pre-classifier feature mapping to adapt the feature from a single task-agnostic model learned from multiple domains for unseen tasks. 
While we build on their feature extractor and pre-classifier alignment, the pre-classifier alignment provides very limited capacity for task adaptation, which we address by adapting the feature extractor with adapters at multiple layers.
FLUTE~\cite{triantafillou2021flute} follows a hybrid three step approach that first learns the parameters of domain-specific FiLM layers so called templates, employs an auxiliary network to initialize the parameters of a new FiLM layer for unseen task by combining the templates and finetunes them on the small support set.
Different from FLUTE, our method learns such adaptation in a single step by learning residual adapters in meta-test.

There are also methods (\eg \cite{saikia2020optimized,doersch2020crosstransformers}) that do not fit into task-agnostic and task-specific parameterization grouping.
BOHB~\cite{saikia2020optimized} proposes to use multi-domain data as validation objective for hyper-parameter optimization such that the feature learned on ImageNet with the optimized hyper-parameter generalizes well to multi-domain. 
CTX~\cite{doersch2020crosstransformers} proposes to learn spatial correspondences from ImageNet and evaluates on the remaining (unseen) domains.
We also compare our method to them in the setting where we use a standard single domain learning network learned from ImageNet and adapt its representations through residual adapters.

\section{Method}\label{sec:method}

\begin{figure*}[t]
\includegraphics[width=0.9\linewidth]{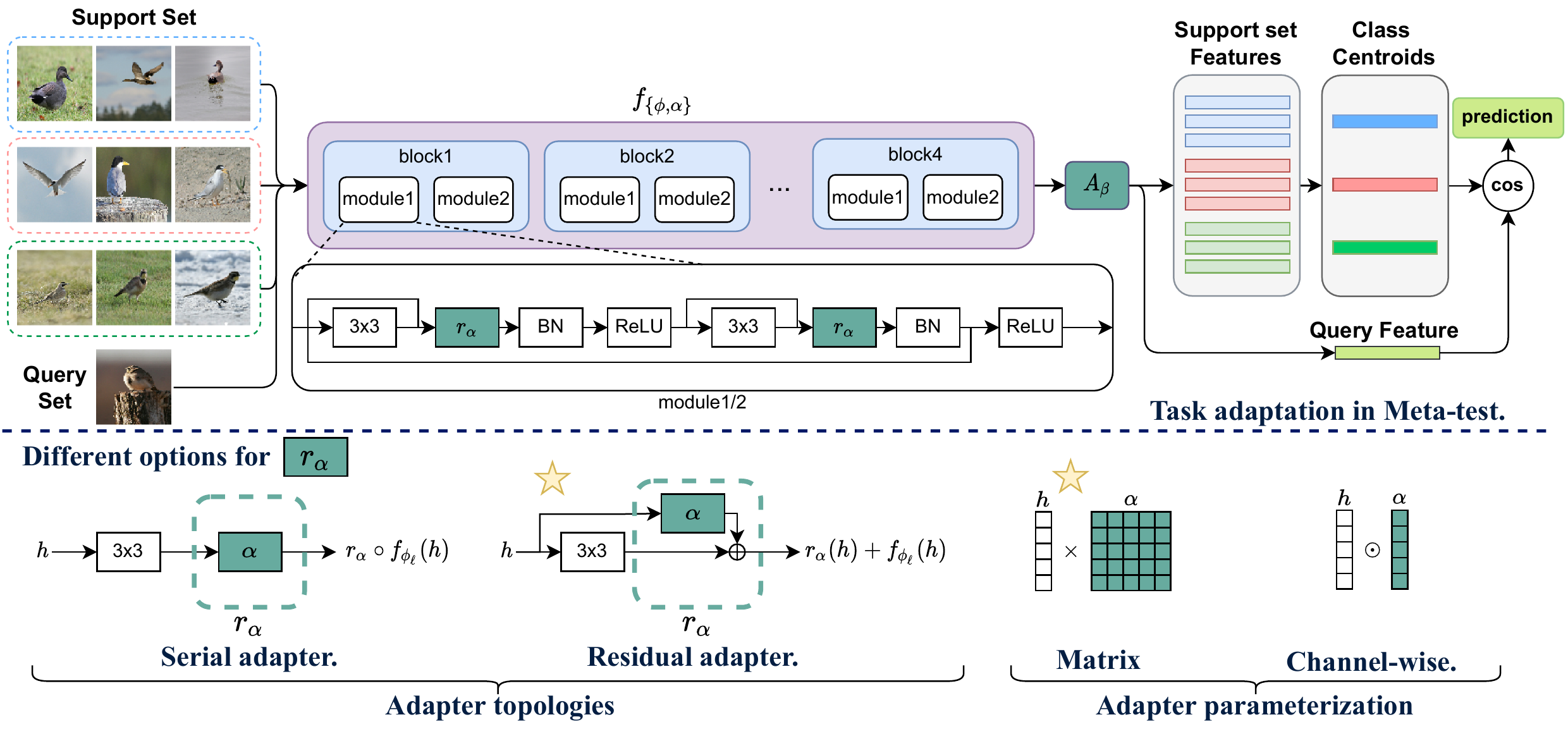}
\vspace{-0.35cm}
\caption{Illustration of our task adaptation for cross-domain few-shot learning. In meta-test stage (a), our method first attaches a parametric transformation $r_{\alpha}$ to each layer, where $\alpha$ can be constructed by (b) a serial or (c) a residual topology. They can be parameterized with matrix multiplication (d) or  channel-wise scaling (e).
We found that (c) is the best configuration with matrix parameterization which is further improved by attaching a linear transformation $A_{\beta}$ to the end of the network. We adapt the network for a given task by optimizing $\alpha$ and $A_{\beta}$ on a few labeled images from the support set, then map query images to the task-specific space and assign them to the nearest class center.}
\label{fig:framework}
\end{figure*}

Few-shot classification aims at learning to classify samples of new categories efficiently from few samples only. 
Each few-shot learning task consists of a support set $\mathcal{S}=\{(\bx_i, y_i)\}_{i=1}^{\rvert \mathcal{S} \lvert}$ with $\rvert \mathcal{S} \lvert$ sample and label pairs respectively and a query set $\mathcal{Q}=\{(\bx_j)\}_{j=1}^{\lvert \mathcal{Q} \rvert }$ with $\lvert \mathcal{Q} \rvert$ samples to be classified.
The goal is to learn a classifier on $\mathcal{S}$ that accurately predicts the labels of $\mathcal{Q}$.
Note that this paper focuses on few-shot image classification problem, \ie $\bx$ and $y$ denote an image and its label.

As in \cite{dvornik2020selecting,liu2020universal,li2021universal}, we solve this problem in two steps involving i) representation learning where we learn a task-agnostic feature extractor $f$ from a large dataset $\mathcal{D}_b$, ii) task adaptation where we adapt the task-agnostic representations through various task-specific weights to the target tasks $(\mathcal{S},\mathcal{Q})$ that are sampled from another large dataset $\mathcal{D}_{t}$ by taking the subsets of the dataset to build $\mathcal{S}$ and $\mathcal{Q}$.
Note that $\mathcal{D}_{b}$ and $\mathcal{D}_{t}$ contain mutually exclusive classes.

\subsection{Task-agnostic representation learning}
Learning task-agnostic or universal representations~\cite{bilen2017universal} has been key to the success of cross-domain generalization.
Representations learned from a large diverse dataset such as ImageNet~\cite{deng2009imagenet} can be considered as universal and successfully transferred to tasks in different domains with minor adaptations~\cite{rebuffi2017learning,liu2020universal,dvornik2020selecting}.
We denote this setting as single domain learning (SDL).

More powerful and diverse representations can be obtained by training a single network over multiple domains.
Let $\mathcal{D}_{b}=\{\mathcal{D}_{k}\}_{k=1}^{K}$ consists of $K$ subdatasets, each sampled from a different domain. 
The vanilla multi-domain learning (MDL) strategy jointly optimizes network parameters over the images from all $K$ subdatasets:
\begin{equation}\label{eq:mtl}
    \min_{\phi, \psi_{k}} \sum_{k=1}^{K} \frac{1}{|\mathcal{D}_{k}|} \sum_{\bx, y \in \mathcal{D}_{k}} \ell(g_{\psi_{k}} \circ f_{\phi}(\bx), y),
\end{equation} where $\ell$ is cross-entropy loss, $f$ is feature extractor that takes an image as input and outputs a $d$ dimensional feature.
$f$ is parameterized by $\phi$ which is shared across $K$ domains. 
$g_{\psi_{k}}$ is the classifier for domain $k$ and parameterized by $\psi_k$ which is discarded in meta-test.
We denote this setting as MDL.
The challenge in MDL is to allow efficiently sharing the knowledge across the domains while preventing negative transfer between them and also carefully balancing the individual loss functions ( \cite{chen2018gradnorm}). 
URL~\cite{li2021universal}, a variant of MDL, mitigates these challenges by first training individual domain-specific networks offline and then distilling their knowledge into a \emph{single} multi-domain network.
We refer to \cite{li2021universal} for more details.

Another way of obtaining multi-domain representations is to employ multiple domain-specific feature extractors, one for each domain, and adaptively ``fuse'' their features for each task \cite{dvornik2020selecting,liu2021multi,triantafillou2021flute}.
While these methods are effective, they require computing features for each image through multiple feature extractors and are thus computationally expensive.
Due to its simplicity and effectiveness, we conduct experiments with the feature extractor of URL~\cite{li2021universal} along with the SDL one.

\subsection{Task-specific weight learning}
A good task-agnostic feature extractor $f_{\phi}$ is expected to produce representations that generalize to many previously unseen tasks and domains.
However this gets more challenging when there is a large domain gap between the training set $\mathcal{D}_b$ and test set $\mathcal{D}_t$ which requires further adaptation to the target task.
In this work, we propose to incorporate additional capacity to the task-agnostic feature extractor by adding task-specific weights to adapt the representations to the target task by using the support set. Specifically, we directly attach task-specific weights to a learned task-agnostic model, and estimate them from scratch given the support set.
We denote the task-specific weights with $\vartheta$ and task-adapted classifier with $p_{(\phi,\vartheta)}$ that outputs a softmax probability vector whose dimensionality equals to the number of categories in the support set $\mathcal{S}$.

To obtain the task-specific weights, we freeze the task-agnostic weights $\phi$ and minimize the cross-entropy loss $\ell$ over the support samples in meta-test w.r.t. the task-specific weights $\vartheta$ \cite{dvornik2020selecting,tian2020rethinking,li2021universal}:
\begin{equation}
    \label{eq:learn-tsw}
\min_{\vartheta}\frac{1}{|\mathcal{S}|}\sum_{(\bx, y) \in \mathcal{S}} \ell(p_{(\phi,\vartheta)}(\bx),y),
\end{equation} where $\mathcal{S}$ is sampled from the test set $\mathcal{D}_t$. 
Most previous works freeze the task-agnostic weights but estimate the task-specific weights through an auxiliary network (or a task encoder)~\cite{requeima2019fast,bateni2020improved,li2021universal,triantafillou2021flute}, where inaccurate prediction of parameters can lead to noisy adaptation and wrong prediction.

\subsection{Task-specific adapter parameterization ($\vartheta$)}
Task adaptation techniques can be broadly grouped into two categories that aims
to adapt the feature extractor or classifier to a given target task.
We use $\alpha$ and $\beta$ to denote task-specific weights for adapting the feature extractor and classifier respectively where $\vartheta=\{\alpha, \beta\}$.

\paragraph{Feature extractor adaptation.} A simple method to adapt $f_{\phi}$ is finetuning its parameters on the support set \cite{chen2020new,dhillon2019baseline}. 
However, this strategy tends to suffer from the unproportionate optimization, \ie updating very high-dimensional weights from a small number of support samples.
In this paper, we propose to attach task-specific adapters directly to the existing task-agnostic model, \eg we attach the adapters to each module of a ResNet backbone in \cref{fig:framework} (a), and the adapters can be efficiently learned/estimated from few samples.
Concretely, let $f_{\phi_l}$ denote the $l$-th layer of the feature extractor $f_{\phi}$ (\ie a convolutional layer) with the weights $\phi_l$. 
Given a support set $\mathcal{S}$, the task-specific adapters $r_{\alpha}$ parameterized by $\alpha$, can be incorporated to the output of the layer $f_{\phi_l}$ as
\begin{equation}\label{eq:ra}
	f_{\{\phi_l,\alpha\}}(\bh) = r_{\alpha}(f_{\phi_l}(\bh),\bh)
\end{equation} where $\bh\in\mathrm{R}^{W\times H \times C}$ is the input tensor, $f_{\phi_l}$ is a convolutional layer in $f_{\phi}$. 
Importantly, the number of the task-specific adaptation parameters $\alpha$ are significantly smaller than the task-agnostic ones.
The adapters can be designed in different ways.

Next we propose two connection types for incorporating $r_\alpha$ to $f_{\phi_l}$: i) serial connection by subsequently applying it to the output of layer $f_{\phi_l}(\bh)$ as
\[
    f_{\{\phi_l,\alpha\}}(\bh) = r_{\alpha} \circ f_{\phi_l}(\bh)
\] which is illustrated in \cref{fig:framework}(b), and ii) parallel connection by a residual addition as in \cite{rebuffi2018efficient}
\[
    f_{\{\phi_l,\alpha\}}(\bh) = r_{\alpha}(\bh) + f_{\phi_l}(\bh)
\] which is illustrated in \cref{fig:framework}(c). 
In our experiments, we found the parallel setting performing the best when $\alpha$ is learned on a support set during meta-test (illustrated in \cref{fig:framework}(c)) which we discuss in \cref{sec:exp}. 

For the parameterization of $r_{\alpha}$, we consider two options.
Matrix multiplication (illustrated in \cref{fig:framework}(d)) with $\alpha\in\mathrm{R}^{C\times C}$:
\[
    r_{\alpha}(\bh) = \bh \ast \alpha,
\] where $\ast$ denotes a convolution, $\alpha\in \mathbb{R}^{C\times C}$ and the transformation is implemented as a convolutional operation with $1\times 1$ kernels in our code.
And channelwise scaling (illustrated in \cref{fig:framework}(e)):
\[
    r_{\alpha}(\bh) = \bh \odot \alpha,
\] where $\odot$ is a Hadamard product and $\alpha\in \mathbb{R}^C$.
Note that one can also use an additive bias weight in both settings, however, this has not resulted in any significant gains in our experiments. 
While the matrix multiplication is more powerful than the scaling operation, it also requires more parameters to be estimated or learned.
Note that, in a deep neural network, the number of input $C_{\text{in}}$ and output channels $C_{\text{out}}$ for a layer can be different. 
In that case, one can still use a non-square matrix: $\alpha\in\mathrm{R}^{C_{out}\times C_{in}}$, however, it is not possible to use a scaling operator in the parallel setting. 
In our experiments, we use ResNet architecture~\cite{he2016deep} where most input and output channels are the same.
$r_\alpha$ connected in parallel with matrix multiplication form, when its parameters $\alpha$ are learned on the support set, is known as residual adapter~\cite{rebuffi2018efficient} and $r_\alpha$ connected serial in channelwise is known as FiLM~\cite{perez2018film}.

\begin{table*}[t]
	\centering
    \resizebox{0.9\textwidth}{!}
    {
		\begin{tabular}{ccccccccccccc}

		    \toprule
		    Test Dataset & CNAPS~\cite{requeima2019fast} & Simple CNAPS~\cite{bateni2020improved} & TransductiveCNAPS~\cite{bateni2020enhancing} & SUR~\cite{dvornik2020selecting} & URT~\cite{liu2020universal} & FLUTE~\cite{triantafillou2021flute} & tri-M~\cite{liu2021multi} & URL~\cite{li2021universal} & Ours\\
		    \midrule
			ImageNet & $50.8 \pm 1.1$ & $58.4 \pm 1.1$ & $57.9 \pm 1.1$ & $56.2 \pm 1.0$ & $56.8 \pm 1.1$ & $58.6 \pm 1.0$ & $51.8 \pm 1.1$ & $58.8 \pm 1.1$ & ${\bf 59.5 \pm 1.0}$ \\
			Omniglot & $91.7 \pm 0.5$ & $91.6 \pm 0.6$ & $94.3 \pm 0.4$ & $94.1 \pm 0.4$ & $94.2 \pm 0.4$ & $92.0 \pm 0.6$ & $93.2 \pm 0.5$ & $94.5 \pm 0.4$ & ${\bf 94.9 \pm 0.4}$ \\
			Aircraft & $83.7 \pm 0.6$ & $82.0 \pm 0.7$ & $84.7 \pm 0.5$ & $85.5 \pm 0.5$ & $85.8 \pm 0.5$ & $82.8 \pm 0.7$ & $87.2 \pm 0.5$ & $89.4 \pm 0.4$ & ${\bf 89.9 \pm 0.4}$ \\
			Birds & $73.6 \pm 0.9$ & $74.8 \pm 0.9$ & $78.8 \pm 0.7$ & $71.0 \pm 1.0$ & $76.2 \pm 0.8$ & $75.3 \pm 0.8$ & $79.2 \pm 0.8$ & $80.7 \pm 0.8$ & ${\bf 81.1 \pm 0.8}$ \\
			Textures & $59.5 \pm 0.7$ & $68.8 \pm 0.9$ & $66.2 \pm 0.8$ & $71.0 \pm 0.8$ & $71.6 \pm 0.7$ & $71.2 \pm 0.8$ & $68.8 \pm 0.8$ & $77.2 \pm 0.7$ & ${\bf 77.5 \pm 0.7}$ \\
			Quick Draw & $74.7 \pm 0.8$ & $76.5 \pm 0.8$ & $77.9 \pm 0.6$ & $81.8 \pm 0.6$ & $82.4 \pm 0.6$ & $77.3 \pm 0.7$ & $79.5 \pm 0.7$ & ${\bf 82.5 \pm 0.6}$ & $81.7 \pm 0.6$ \\
			Fungi & $50.2 \pm 1.1$ & $46.6 \pm 1.0$ & $48.9 \pm 1.2$ & $64.3 \pm 0.9$ & $64.0 \pm 1.0$ & $48.5 \pm 1.0$ & $58.1 \pm 1.1$ & ${\bf 68.1 \pm 0.9}$ & $66.3 \pm 0.8$ \\
			VGG Flower & $88.9 \pm 0.5$ & $90.5 \pm 0.5$ & ${\bf 92.3 \pm 0.4}$ & $82.9 \pm 0.8$ & $87.9 \pm 0.6$ & $90.5 \pm 0.5$ & $91.6 \pm 0.6$ & $92.0 \pm 0.5$ & $92.2 \pm 0.5$ \\
			\midrule
			Traffic Sign & $56.5 \pm 1.1$ & $57.2 \pm 1.0$ & $59.7 \pm 1.1$ & $51.0 \pm 1.1$ & $48.2 \pm 1.1$ & $63.0 \pm 1.0$ & $58.4 \pm 1.1$ & $63.3 \pm 1.1$ & ${\bf 82.8 \pm 1.0}$ \\
			MSCOCO & $39.4 \pm 1.0$ & $48.9 \pm 1.1$ & $42.5 \pm 1.1$ & $52.0 \pm 1.1$ & $51.5 \pm 1.1$ & $52.8 \pm 1.1$ & $50.0 \pm 1.0$ & $57.3 \pm 1.0$ & ${\bf 57.6 \pm 1.0}$ \\
			MNIST & - & $94.6 \pm 0.4$ & $94.7 \pm 0.3$ & $94.3 \pm 0.4$ & $90.6 \pm 0.5$ & $96.2 \pm 0.3$ & $95.6 \pm 0.5$ & $94.7 \pm 0.4$ & ${\bf 96.7 \pm 0.4}$ \\
			CIFAR-10 & - & $74.9 \pm 0.7$ & $73.6 \pm 0.7$ & $66.5 \pm 0.9$ & $67.0 \pm 0.8$ & $75.4 \pm 0.8$ & $78.6 \pm 0.7$ & $74.2 \pm 0.8$ & ${\bf 82.9 \pm 0.7}$ \\
			CIFAR-100 & - & $61.3 \pm 1.1$ & $61.8 \pm 1.0$ & $56.9 \pm 1.1$ & $57.3 \pm 1.0$ & $62.0 \pm 1.0$ & $67.1 \pm 1.0$ & $63.5 \pm 1.0$ & ${\bf 70.4 \pm 0.9}$ \\
			\midrule
			Average Seen & $71.6$ & $73.7$ & $75.1$ & $75.9$ & $77.4$ & $74.5$ & $76.2$ & ${\bf 80.4}$ & ${\bf 80.4}$ \\
			Average Unseen & - & $67.4$ & $66.5$ & $64.1$ & $62.9$ & $69.9$ & $69.9$ & $70.6$ & ${\bf 78.1}$ \\
			Average All & - & $71.2$ & $71.8$ & $71.4$ & $71.8$ & $72.7$ & $73.8$ & $76.6$ & ${\bf 79.5}$ \\
			\midrule
			Average Rank & - & $6.1$ & $5.5$ & $5.6$ & $5.5$ & $4.8$ & $4.4$ & $2.5$ & ${\bf 1.6}$ \\
			\bottomrule
		\end{tabular}%
			}
		\vspace{-0.35cm}
		\caption{Comparison state-of-the-art methods on Meta-Dataset (using a multi-domain feature extractor of \cite{li2021universal}). Mean accuracy, 95\% confidence interval are reported. The first eight datasets are seen during training and the last five datasets are unseen and used for test only.}
		\label{tab:currmethod}
\end{table*}%

An alternative to reduce the dimensionality of $\alpha$ in case of matrix multiplication is matrix decomposition:
$\alpha=V\gamma^\top$, where $V\in\mathrm{R}^{C\times B}$ and $\gamma\in\mathrm{R}^{C\times B}$, $B\ll C$.
Using a bottleneck, \ie setting $B<C/2$, reduces the number of parameters in the multiplication.
In this work, we set $B=[C/N]$ and evaluate the performance for various $N$ in \cref{sec:exp}.

\paragraph{Classifier learning.} 
Finally, the adapted feature extractor $f_{(\phi,\alpha)}$ can be combined with a task-specific classifier $c_{\beta}$, parameterized by $\beta$ to obtain the final model, \ie $c \circ f_{(\phi,\alpha)}$.
Based on the recent works, we investigate use of various linear classifiers in~\cite{dhillon2019baseline,lee2019meta,chen2020new,requeima2019fast}, also nonparameteric ones including nearest centroid classifier (NCC)~\cite{mensink2013distance,snell2017prototypical} and their variants based on Mahalanobis distance (MD) \cite{bateni2020improved}.
Recently, it was shown in \cite{li2021universal} that nonparametric classifiers can be successfully combined with a pre-classifier transformation.
Concretely, the transformation in~\cite{li2021universal} that takes in the features computed from the network $f_{\{\phi, \alpha\}} \in \mathrm{R}^d$ and apply an affine transformation $A_{\beta}:\mathrm{R}^d\rightarrow \mathrm{R}^d$ parameterized by $\beta\in\mathrm{R}^{d\times d}$ to obtain the network embedding that is fed into the classifier, \ie $p_{\phi,\vartheta} =  c \circ A_{\beta} \circ f_{\{\phi, \alpha\}}$.
Note that in the case of non-parametric classifier, $c$ is not parameterized by $\beta$ and we use $\beta$ to denote the transformation parameters.

In our experiments, the best performing setting uses parallel adapters, whose parameters are in the matrix form, to adapt the feature extractor and followed by the pre-classifier transformation and NCC.

\section{Experiments}\label{sec:exp}

Here we start with experimental setup, and
then we compare our method to the state-of-the-art methods and rigorously evaluate various design decisions.
We finally provide further analysis.

\subsection{Experimental setup}
\paragraph{Dataset.} We use the Meta-Dataset~\cite{triantafillou2019meta} which is the standard benchmark for few-shot classification.
It contains images from 13 diverse datasets and we follow the standard protocol in \cite{triantafillou2019meta} (more details in the supplementary).

\paragraph{Implementation details.}
As in~\cite{dvornik2020selecting,bateni2020improved,li2021universal}, we build our method on ResNet-18~\cite{he2016deep} backbone, which is trained over eight training subdatasets by following~\cite{li2021universal}  with the same hyperparameters in our experiments, unless stated otherwise.
Once learned, we freeze its parameters and use them as the task-agnostic weights.
For learning task-specific weights ($\vartheta$), including the pre-classifier transformation $\beta$ and the adapter parameters, we directly attach them to the task-agnostic weights and learn them on the support samples in meta-test by using Adadelta optimizer~\cite{zeiler2012adadelta}. 

In the study of various task adaptation strategies in Section~\ref{sec:analysis}, we consider to only estimate the adapter parameters and learn the auxiliary network parameters by using Adam optimizer as in~\cite{requeima2019fast,bateni2020improved} in meta-train.
Note that estimation of pre-classifier and classifier weights via the auxiliary network leads to noisy and poor results and we do not report them.
Similarly, we found that the auxiliary network fails to estimate very high-dimensional weights.
Hence we only use it to estimate adapter weights that are parameterized with a vector for channelwise multiplication but not with a matrix.

\begin{table*}[ht]
	\centering
    \resizebox{0.9\textwidth}{!}
    {
		\begin{tabular}{cccccccc|ccc}

		   	& \multicolumn{7}{c}{ResNet-18} & \multicolumn{3}{c}{ResNet-34} \\
		   	\toprule
		    \multirow{2}{*}{Test Dataset} & Finetune & ProtoNet & fo-Proto- & ALFA+fo-Proto & BOHB & FLUTE & \multirow{2}{*}{Ours} & ProtoNet & CTX & \multirow{2}{*}{Ours}\\
		    & \cite{triantafillou2019meta} & \cite{triantafillou2019meta} & MAML~\cite{triantafillou2019meta} & -MAML~\cite{triantafillou2019meta} & \cite{saikia2020optimized} & \cite{triantafillou2021flute} &  & \cite{doersch2020crosstransformers} & \cite{doersch2020crosstransformers} & \\
		    \midrule
			ImageNet & $45.8 \pm 1.1$ & $50.5 \pm 1.1$ & $49.5 \pm 1.1$ & $52.8 \pm 1.1$ & $51.9 \pm 1.1$ & $46.9 \pm 1.1$ & ${\bf 59.5 \pm 1.1}$ & $53.7 \pm 1.1$ & $62.8 \pm 1.0$ & ${\bf 63.7 \pm 1.0}$ \\
			\midrule
			Omniglot & $60.9 \pm 1.6$ & $60.0 \pm 1.4$ & $63.4 \pm 1.3$ & $61.9 \pm 1.5$ & $67.6 \pm 1.2$ & $61.6 \pm 1.4$ & ${\bf 78.2 \pm 1.2}$ & $68.5 \pm 1.3$ & $82.2 \pm 1.0$ & ${\bf 82.6 \pm 1.1}$ \\
			Aircraft & $68.7 \pm 1.3$ & $53.1 \pm 1.0$ & $56.0 \pm 1.0$ & $63.4 \pm 1.1$ & $54.1 \pm 0.9$ & $48.5 \pm 1.0$ & ${\bf 72.2 \pm 1.0}$ & $58.0 \pm 1.0$ & $79.5 \pm 0.9$ & ${\bf 80.1 \pm 1.0}$ \\
			Birds & $57.3 \pm 1.3$ & $68.8 \pm 1.0$ & $68.7 \pm 1.0$ & $69.8 \pm 1.1$ & $70.7 \pm 0.9$ & $47.9 \pm 1.0$ & ${\bf 74.9 \pm 0.9}$ & $74.1 \pm 0.9$ & $80.6 \pm 0.9$ & ${\bf 83.4 \pm 0.8}$ \\
			Textures & $69.0 \pm 0.9$ & $66.6 \pm 0.8$ & $66.5 \pm 0.8$ & $70.8 \pm 0.9$ & $68.3 \pm 0.8$ & $63.8 \pm 0.8$ & ${\bf 77.3 \pm 0.7}$ & $68.8 \pm 0.8$ & $75.6 \pm 0.6$ & ${\bf 79.6 \pm 0.7}$ \\
			Quick Draw & $42.6 \pm 1.2$ & $49.0 \pm 1.1$ & $51.5 \pm 1.0$ & $59.2 \pm 1.2$ & $50.3 \pm 1.0$ & $57.5 \pm 1.0$ & ${\bf 67.6 \pm 0.9}$ & $53.3 \pm 1.1$ & ${\bf 72.7 \pm 0.8}$ & $71.0 \pm 0.8$ \\
			Fungi & $38.2 \pm 1.0$ & $39.7 \pm 1.1$ & $40.0 \pm 1.1$ & $41.5 \pm 1.2$ & $41.4 \pm 1.1$ & $31.8 \pm 1.0$ & ${\bf 44.7 \pm 1.0}$ & $40.7 \pm 1.1$ & ${\bf 51.6 \pm 1.1}$ & $51.4 \pm 1.2$ \\
			VGG Flower & $85.5 \pm 0.7$ & $85.3 \pm 0.8$ & $87.2 \pm 0.7$ & $86.0 \pm 0.8$ & $87.3 \pm 0.6$ & $80.1 \pm 0.9$ & ${\bf 90.9 \pm 0.6}$ & $87.0 \pm 0.7$ & ${\bf 95.3 \pm 0.4}$ & $94.0 \pm 0.5$ \\
			Traffic Sign & $66.8 \pm 1.3$ & $47.1 \pm 1.1$ & $48.8 \pm 1.1$ & $60.8 \pm 1.3$ & $51.8 \pm 1.0$ & $46.5 \pm 1.1$ & ${\bf 82.5 \pm 0.8}$ & $58.1 \pm 1.1$ & ${\bf 82.7 \pm 0.8}$ & $81.7 \pm 0.9$ \\
			MSCOCO & $34.9 \pm 1.0$ & $41.0 \pm 1.1$ & $43.7 \pm 1.1$ & $48.1 \pm 1.1$ & $48.0 \pm 1.0$ & $41.4 \pm 1.0$ & ${\bf 59.0 \pm 1.0}$ & $41.7 \pm 1.1$ & $59.9 \pm 1.0$ & ${\bf 61.7 \pm 0.9}$ \\
			MNIST & - & - & - & - & - & $80.8 \pm 0.8$ & ${\bf 93.9 \pm 0.6}$ & - & - & ${\bf 94.6 \pm 0.5}$ \\
			CIFAR-10 & - & - & - & - & - & $65.4 \pm 0.8$ & ${\bf 82.1 \pm 0.7}$ & - & - & ${\bf 86.0 \pm 0.6}$ \\
			CIFAR-100 & - & - & - & - & - & $52.7 \pm 1.1$ & ${\bf 70.7 \pm 0.9}$ & - & - & ${\bf 78.3 \pm 0.8}$ \\
			\midrule
			Average Seen & $45.8$ & $50.5$ & $49.5$ & $52.8$ & $51.9$ & $46.9$ & ${\bf 59.5}$ & $53.7$ & $62.8$ & ${\bf 63.7}$ \\
			Average Unseen & $58.2$ & $56.7$ & $58.4$ & $62.4$ & $60.0$ & $53.2$ & ${\bf 71.9}$ & $61.1$ & $75.6$ & ${\bf 76.2}$ \\
			Average All & $57.0$ & $56.1$ & $57.5$ & $61.4$ & $59.2$ & $52.6$ & ${\bf 70.7}$ & $60.4$ & $74.3$ & ${\bf 74.9}$ \\
			\midrule
			Average Rank & $7.9$ & $8.3$ & $7.0$ & $5.3$ & $6.0$ & $8.9$ & ${\bf 2.8}$ & $5.5$ & $1.8$ & ${\bf 1.5}$ \\
			\bottomrule
		\end{tabular}%
			}
		\vspace{-0.35cm}
		\caption{Comparison to state-of-the-art methods on Meta-Dataset (using a single-domain feature extractor which is trained only on ImageNet). Mean accuracy, 95\% confidence interval are reported. Only ImageNet is seen during training and the rest datasets are unseen for test only.}
		\label{tab:currmethodimagenet}
\end{table*}%

\subsection{Comparison to state-of-the-art methods}
We evaluate our method in two settings, with multi-domain or single-domain feature extractor and compare our method to existing state-of-the-art methods. We also evaluate our method incorporated with different feature extractors, \ie SDL, MDL, and URL in the supplementary.

\paragraph{Multi-domain feature extractor.}
Here we incorporate the proposed residual adapters in matrix form to the multi-domain feature extractor of \cite{li2021universal} and compare its performance with the the state-of-the-art methods  (CNAPS~\cite{requeima2019fast}, SUR~\cite{dvornik2020selecting}, URT~\cite{liu2020universal}, Simple CNAPS~\cite{bateni2020improved}, Transductive CNAPS~\cite{bateni2020enhancing}, FLUTE~\cite{triantafillou2021flute}, tri-M~\cite{liu2021multi}, and URL~\cite{li2021universal}) in \cref{tab:currmethod}. 
To better analyze the results, we divide the table into two blocks that show the few-shot classification accuracy in previously seen domains and unseen domains along with their average accuracy.
We also report average accuracy over all domains and the average rank as in~\cite{triantafillou2021flute,li2021universal}.\footnote{As mentioned in \url{https://github.com/google-research/meta-dataset/issues/54}, we further update the evaluation protocol and report the updated results of all methods in the supplementary.}
Simple CNAPS improves over CNAPS by adopting a simple Mahalanobis distance in stead of learning adapted linear classifier. Transductive CNAPS further improves by using unlabelled test images. SUR and URT fuse multi-domain features to get better performance. FLUTE improves URT by fusing FiLM parameters as initialization which is further finetuned on the support set in meta-test.
tri-M adopts the same strategy of learning modulation parameters as CNAPS, where the parameters are further divided into the domain-specific set and the domain-cooperative set to explore the intra-domain information and inter-domain correlations, respectively.
URL surpasses previous methods by learning a universal representation with distillation from multiple domains.

From the results, our method outperforms other methods on most domains (10 out of 13), especially obtaining significant improvement on 5 unseen datasets than the second best method, \ie Average Unseen (+7.5). More specifically, our method obtains significant better results than the second best approach on Traffic Sign (+19.5), CIFAR-10 (+8.7), and CIFAR-100 (+6.8). 
Achieving improvement on unseen domains is more challenging due to the large gap between seen and unseen domain and the scarcity of labeled samples for the unseen task. 
We address this problem by attaching light-weight adapters to the feature extractor residually and learn the attached adapters on support set from scratch. 
This allows the model to learn more accurate and effective task-specific parameters (adapters) from the support set to efficiently steer the task-agnostic features for the unseen task, compared with predicting task-specific parameters by an auxiliary network learned in meta-train, \eg Simple CNAPS, tri-M, or fusing representations from multiple feature extractors \eg SUR, URT.
Though FLUTE uses a hybrid approach which uses auxiliary networks learned from meta-train to initialize the FiLM parameters for further fine-tuning, their results are not better than URL, 
which achieves very competitive results as it learns a good universal representation that generalizes well to seen domains and can be further improved with the adaptation strategy proposed in this work, especially significant improvements on unseen domains.

\begin{table*}[ht!]
	\centering
    \resizebox{0.95\textwidth}{!}
    {
		\begin{tabular}{lcccccccccccccc|ccccc}

		    \toprule
		    \multirow{2}{*}{Test Dataset} & \multirow{2}{*}{classifier} & Aux-Net & serial or & M or & \multirow{2}{*}{$\beta$} & \multirow{2}{*}{\#params} & Image & Omni & Air- & \multirow{2}{*}{Birds} & Tex- & Quick & \multirow{2}{*}{Fungi} & VGG & Traffic & MS- & \multirow{2}{*}{MNIST} & CIFAR & CIFAR\\
		    & & or Ad & residual & CW & & & -Net & -glot & craft & & tures & Draw & & Flower & Sign & COCO & & -10 & -100 \\
		    \midrule
		    NCC & NCC & - & - & - & \XSolidBrush &  - & $57.0$ & $94.4$ & $88.0$ & $80.3$ & $74.6$ & $81.8$ & $66.2$ & $91.5$ & $49.8$ & $54.1$ & $91.1$ & $70.6$& $59.1$\\ 
		    MD & MD & - & - & - & \XSolidBrush & - & $53.9$ & $93.8$ & $87.6$ & $78.3$ & $73.7$ & $80.9$ & $57.7$ & $89.7$ & $62.2$ & $48.5$ & $95.1$ & $68.9$ & $60.0$ \\ 
		    LR & LR & - & - & - & \XSolidBrush & - & $56.0$ & $93.7$ & $88.3$  & $79.7$ & $74.7$ & $80.0$ & $62.1$ & $91.1$ & $59.7$ & $51.2$ & $93.5$  & $73.1$ & $60.1$ \\
		    SVM & SVM & - & - & - & \XSolidBrush & - & $54.5$ & $94.3$ & $87.7$ & $78.1$ & $73.8$ & $80.0$ & $58.5$ & $91.4$ & $65.7$ & $50.5$ & $95.4$ & $72.0$ & $60.5$ \\
		    \midrule
		    Finetune & NCC & - & - & - & \XSolidBrush & - & $55.9$ & $94.0$ & $87.3$ & $77.8$ & $76.8$ & $75.3$ & $57.6$ & $91.5$ & ${\bf 86.1}$ & $53.1$ & ${\bf 96.8}$ & $80.9$ & $65.9$ \\
		    \midrule
		    Aux-S-CW & NCC & Aux-Net & serial & CW & \XSolidBrush & 76.98\% & $54.6$ & $93.5$ &  $86.6$ & $78.6$ &  $71.5$ &   $79.3$ &  $66.0$ &   $87.6$ &  $43.3$ & $49.1$ & $87.9$ &  $62.8$ &  $51.5$  \\
		    Aux-R-CW & NCC & Aux-Net & residual & CW & \XSolidBrush & 76.98\% & $56.1$ &  $94.2$ & $88.4$ & $80.6$ & $74.9$ &  $82.0$ & $66.4$ & $91.6$ & $48.5$ & $53.5$ & $90.8$ &  $70.2$ & $59.7$ \\
		    Aux-S-CW & MD & Aux-Net & serial & CW & \XSolidBrush & 76.98\% & $55.1$ &$93.8$ &$86.8$ & $77.4$ & $73.2$ &$79.9$ &$57.4$ & $88.1$ & $58.4$ & $50.1$ & $92.7$ & $66.5$ & $55.7$ \\
		    Aux-R-CW & MD & Aux-Net & residual & CW & \XSolidBrush & 76.98\% & $54.8$ & $93.8$ & $87.4$ & $78.2$ & $73.4$ & $81.1$ & $58.8$ & $90.1$ & $63.6$ & $48.5$ & $94.8$ & $69.6$ & $60.6$ \\
		    \midrule
		    Ad-S-CW & NCC & Ad & serial & CW & \XSolidBrush & 0.06\% & $56.8$ & $94.8$ & $89.3$ & $80.7$ & $74.5$ & $81.6$ & $65.8$ & $91.3$ & $73.9$ & $53.6$ & $95.7$ & $78.4$ & $64.3$ \\
		    Ad-R-CW & NCC & Ad & residual & CW & \XSolidBrush & 1.57\%   & $57.6$ & $94.7$ & $89.0$ & $81.2$ & $75.2$ & $81.5$ & $65.4$ & $91.8$ & $79.2$ & $54.7$ & $96.4$ & $79.5$ & $67.4$ \\
		    Ad-S-M  & NCC & Ad & serial & M & \XSolidBrush & 12.50\% & $56.2$ & $94.4$ & $89.1$ & $80.6$ & $75.8$ & $81.6$ & ${\bf 67.1}$ & $92.1$ & $67.6$ & $54.8$ & $95.9$ & $78.9$ & $66.6$ \\

		    Ad-R-M & NCC & Ad & residual & M & \XSolidBrush & 10.93\%  & $57.3$ & $94.9$ & $88.9$ & $81.0$ & $76.7$ & $80.6$ & $65.4$ & $91.4$ & $82.6$ & $55.0$ & $96.6$ & $82.1$ & $66.4$ \\
			\midrule
		    Ad-R-CW-PA & NCC & Ad & residual & CW & \Checkmark & 3.91\% & $58.6$ & $94.5$ & ${\bf 90.0}$ & $80.5$ & ${\bf 77.6}$ & ${\bf 81.9}$ & $67.0$ & ${\bf 92.2}$ & $80.2$ & $57.2$ & $96.1$ & $81.5$ & ${\bf 71.4}$ \\
		    Ad-R-M-PA & NCC & Ad & residual & M & \Checkmark & 13.27\% & ${\bf 59.5}$ & ${\bf94.9}$ & $89.9$ & ${\bf 81.1}$ & $77.5$ & $81.7$ & $66.3$ & ${\bf 92.2}$ & $82.8$ & ${\bf 57.6}$ & $96.7$ & ${\bf 82.9}$ & $70.4$ \\
			\bottomrule
		\end{tabular}%
			}
		\vspace{-0.35cm}
		\caption{Comparisons to methods that learn classifiers and model adaptation methods during meta-test stage based on URL model. NCC, MD, LR, SVM denote nearest centroid classifier, Mahalanobis distance, logistic regression, support vector machines respectively. `Aux-Net or Ad' indicates using Auxiliary Network to predict $\alpha$ or attaching adapter $\alpha$ directly. `M or CW' means using matrix multiplication or channel-wise scaling adapters. `S' and `R' denote serial adapter and residual adapter, respectively. `$\beta$' indicates using the pre-classifier adaptation. The standard deviation results can be found in the supplementary. The first eight datasets are seen during training and the last five datasets are unseen and used for test only. 
		}
		\label{tab:testad}
\end{table*}%

\paragraph{Single-domain feature extractor.}
We also evaluate our method with a single-domain feature extractor trained on ImageNet only on ResNet-18 as in~\cite{triantafillou2019meta} or ResNet-34 as in~\cite{doersch2020crosstransformers}.
This setting is more challenging than the multi-domain one, as the model is trained only on one domain and tested on both test split of ImageNet but also of other domains.
We report the results of our method and state-of-the-art methods (BOHB~\cite{saikia2020optimized}, FLUTE~\cite{triantafillou2021flute}, Finetune~\cite{triantafillou2019meta}, ProtoNet~\cite{triantafillou2019meta}, fo-Proto-MAML~\cite{triantafillou2019meta}, and ALFA+fo-Proto-MAML~\cite{triantafillou2019meta}, CTX~\cite{doersch2020crosstransformers}) in \cref{tab:currmethodimagenet}.
ALFA+fo-Proto-MAML achieves the prior best performance by combining the complementary strengths of Prototypical Networks and MAML (fo-Proto-MAML), with extra meta-learning of per-step hyperparameters: learning rate and
weight decay coefficients. FLUTE fails to surpass it with one training source domain, probably due to the lack of FiLM parameters from multiple domains.
Our method, when using ResNet18 backbone, outperforms other methods on all domains, especially obtaining significant improvement, \ie Average Unseen (+9.5), on 12 unseen datasets than the second best method.
We compare our method to CTX and ProtoNet, which use ResNet-34 backbone.
\footnote{Note that CTX also uses augmentation strategies such as AutoAugment~\cite{cubuk2019autoaugment} and other ones from SimClr~\cite{chen2020simple}. We expect applying the same augmentation strategies to our method would yield further improvements, but we leave this for future work.}
CTX is very competitive by learning coarse spatial correspondence between the query and the support images with an attention mechanism. Ours is orthogonal to CTX and both CTX and our method can potentially be complementary, but we leave this as future work due to high computational cost of CTX.
Specifically,
we see that our method obtains the best average rank and outperforms CTX on most domains (6 out of 10) while our method being more efficient (We train our model on one single Nvidia GPU for around 33 hours while CTX requires 8 Nvidia V100 GPUs and 7 days for training. Please refer to the supplementary for more details).

\begin{figure*}[ht]
\RawFloats
\noindent\begin{minipage}[t]{0.32\textwidth}
\begin{center}
\includegraphics[width=0.9\linewidth]{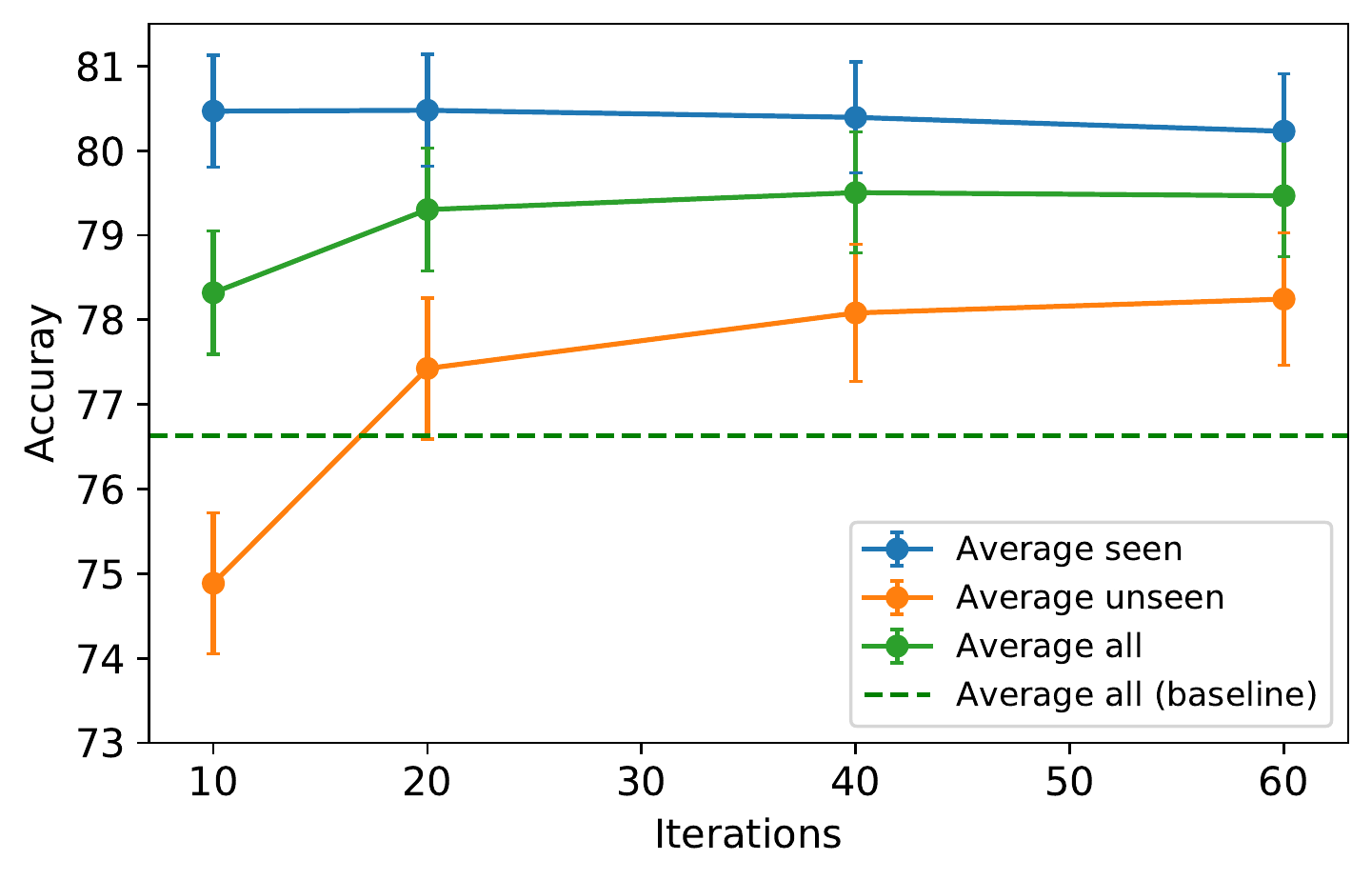}
\end{center}
\vspace{-0.3in}
\caption{Sensitivity of performance to number of iterations.}
\label{fig:urlstab}
\end{minipage} \hfill
\begin{minipage}[t]{0.32\textwidth}
\begin{center}
\includegraphics[width=0.96\linewidth]{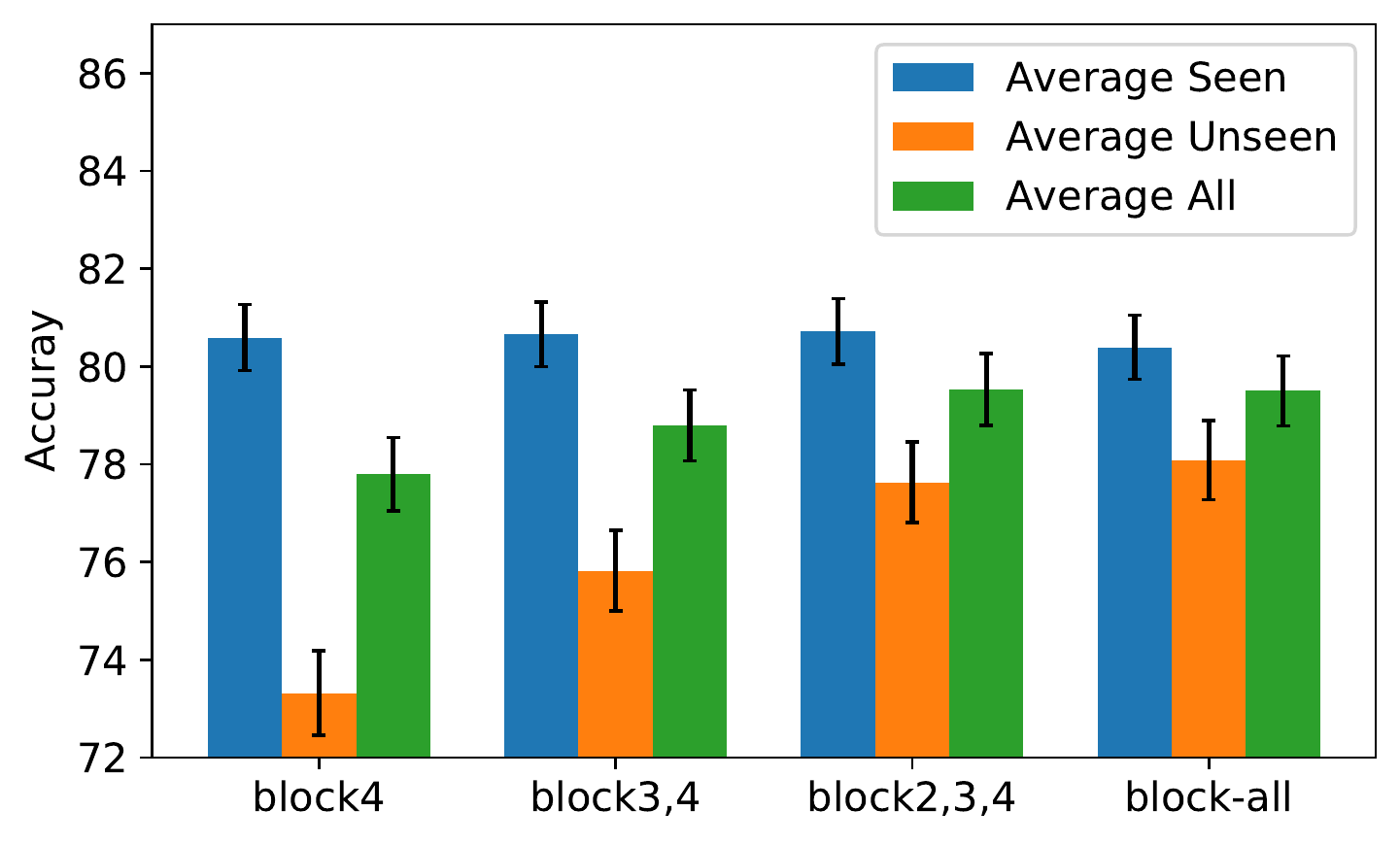}
\end{center}
\vspace{-0.3in}
\caption{Block (layer) analysis for adapters.}
\label{fig:urllayers}
\end{minipage} \hfill
\begin{minipage}[t]{0.32\textwidth}
\begin{center}
\includegraphics[width=0.9\linewidth]{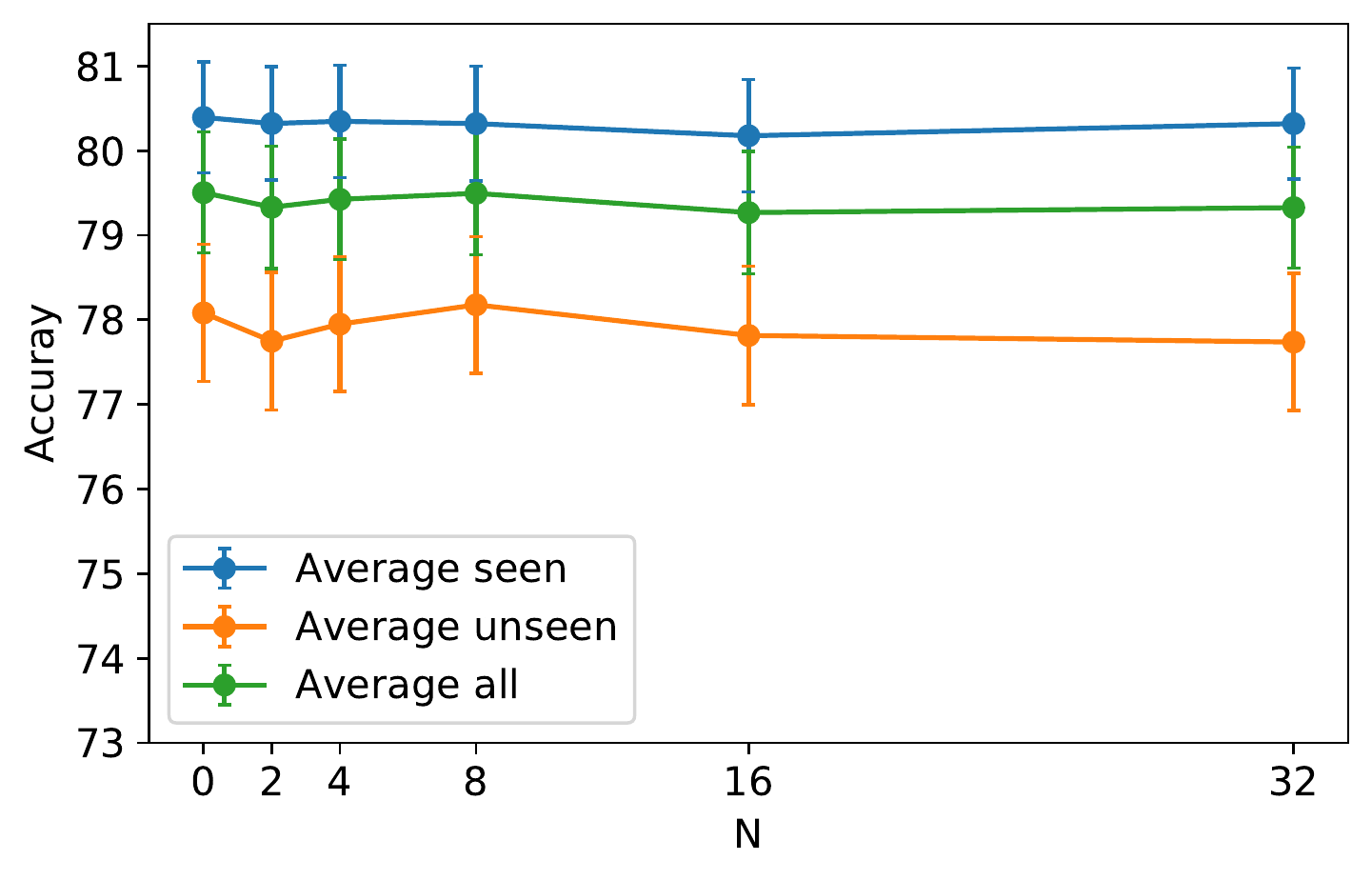}
\end{center}
\vspace{-0.3in}
\caption{Decomposed residual adapters on block-3,4.}
\label{fig:urldecoml34}
\end{minipage} \hfill
\vspace{-0.35cm}
\end{figure*}

\subsection{Analysis of task-specific parameterizations}
\label{sec:analysis}

\paragraph{Classifier learning.}
First we study the adaptation strategies for learning only a task-specific classifier on the pre-trained feature extractor of \cite{li2021universal}.
We evaluate non-parametric classifiers including nearest cetroid classifier (\emph{NCC}) and NCC Mahalanobis Distance (\emph{MD}) and parametric classifiers including logistic regression (\emph{LR}), support vector machine \emph{SVM} whose parameters are learned on support samples.
We also include another baseline with NCC that finetunes all the feature extractor parameters, and report the results in \cref{tab:testad}.
We observe that NCC obtains the best results for the seen domains and its performance is further improved by MD, while SVM achieves the best for the unseen domains among other classifiers. 
Finetuning baseline provides competitive results especially for the unseen domains. However, it performs poor in most seen domains.

\paragraph{Feature extractor adaptation.}
Next we analyze various design decisions for the feature extractor adaptation including connection types (serial, residual), \ie \cref{fig:framework}(b), (c), its parameterization including channelwise modulation (\emph{CW}) when they are estimated by an auxiliary network (\emph{Aux-Net}), which has around 77\% capacity of the feature extractor.
We use with each combination with two nonparameteric classifier, either NCC or MD.
\emph{While the adaptation strategies using residual connections performs better than the serial one in almost all cases, the gains are more substantial when generalizing to unseen domains.}
Learning adapter weights from few samples only can be very noisy. With residual addition, it is not necessary to change all connections for passing the information forward, which can improve the robustness of useful features and reduce learning burdens for new task, hence increase the generalization ability. While the serial connections may damage the previous learned structures.
We also observe that NCC and MD obtain comparable performances.
Note that Aux-S-CW with MD corresponds to our implementation of Simple CNAPS~\cite{bateni2020improved} with the more powerful feature extractor.
We show that replacing its serial connection with a residual one leads to a strong performance boost.

Next we look at the adaptation strategy that learns the task-specific weights directly on the support set as in~\cref{eq:learn-tsw}.
We evaluate serial and residual connection types with channelwise and matrix parameterizations by using NCC. 
We denote this setting as Ad in \cref{tab:testad}.
Note that we omit MD here, as it produces similar results to NCC.
First we observe that learning the weights on the support set outperforms the strategy of estimating them through an auxiliary network almost in all cases.
In addition, the learnable weights requires less number of parameters per task, while the capacity of auxiliary network is fixed.
We again observe that the residual connections are more effective, especially when used with the matrix parameterization (Ad-R-M).
However,the channelwise ones provide a good performance/computation tradeoff.
Finally using the pre-classifier alignment (Ad-R-CW-PA and Ad-R-M-PA) further boosts the performance of the best models and we use our best model Ad-R-M-PA to compare against the state-of-the-art.

\subsection{Further results}
\vspace{-0.15cm}
\paragraph{Varying-way Five-shot.}
After evaluating our method over a broad range of varying shots (\eg up to 100 shots), we follow \cite{doersch2020crosstransformers,li2021universal} to further analyze our method in 5-shot setting of varying number of categories.
In this setting, we sample a varying number of ways with a fixed number of shots to form balanced support and query sets. As shown in Table~\ref{tab:fixedshot}, overall performance for all methods decreases in most datasets compared to results in Table~\ref{tab:currmethod} indicating that this is a more challenging setting. It is due to that five-shot setting samples much less support images per class than the standard setting. 
The top-2 methods remain the same and ours still outperforms the state-of-the-art URL when the number of support images per class is fewer, especially on unseen domains (Average Unseen +6.2).

\paragraph{Five-way One-shot.}
The similar conclusion can be drawn from this challenging case. Note that there are extremely few samples available for training in this case. As we can see, Ours achieves similar results with URL on seen domains but much better performance on unseen domains due to the learning of attached residual adapters is less over-fitting.

\begin{table}[h]
	\centering
	    \resizebox{1.0\textwidth}{!}
    {
		\begin{tabular}{cccccc|ccccc}
			& \multicolumn{5}{c}{Varying-Way Five-Shot} & \multicolumn{5}{c}{Five-Way One-Shot} \\
		    \toprule
		    \multirow{2}{*}{Test Dataset} & Simple & SUR & URT & URL & \multirow{2}{*}{Ours}& Simple & SUR & URT & URL & \multirow{2}{*}{Ours}\\
		    &  CNAPS~\cite{bateni2020improved}  & \cite{dvornik2020selecting} & \cite{liu2020universal} & \cite{li2021universal} & & CNAPS~\cite{bateni2020improved} & \cite{dvornik2020selecting}& \cite{liu2020universal} & \cite{li2021universal} & \\
		    \midrule
		    Average Seen & $69.0$ & $71.2$ & $73.8$ & $76.6$ & ${\bf 76.7}$ & $65.0$ & $64.0$ & $70.6$ & $73.4$ & ${\bf 73.5}$ \\
		    Average Unseen & $62.6$ & $56.0$ & $59.6$ & $65.2$ & ${\bf 71.4}$ & $57.7$ & $49.6$ & $57.5$ & $62.4$ & ${\bf 63.4}$ \\
		    Average All & $66.5$ & $65.4$ & $68.3$ & $72.2$ & ${\bf 74.6}$ & $62.2$ & $58.5$ & $65.5$ & $69.2$ & ${\bf 69.6}$ \\
		    \midrule
		    Average Rank & $4.1$ & $3.9$ & $3.4$ & $2.1$ & ${\bf 1.5}$ & $3.8$ & $4.5$ & $3.3$ & ${\bf 1.7}$ & ${\bf 1.7}$ \\
			\bottomrule
		\end{tabular}%
			}
		\vspace{-0.25cm}
		\caption{Results of  Varying-Way Five-Shot and Five-Way One-Shot scenarios. Mean accuracies are reported and more detailed results can be found in the supplementary.}
		\label{tab:fixedshot}
\end{table}%
\vspace{-0.2cm}

\subsection{Further ablation study}
Here, we conduct ablation study for the sensitivity analysis for number of iterations, layer analysis for adapters, and decomposed residual adapters. We summarize results in figures and refer to supplementary for more detailed results.

\paragraph{Sensitivity analysis for number of iterations.}
In our method, we optimize the attached parameters ($\alpha,\beta$) with 40 iterations. \Cref{fig:urlstab} reports
the results with 10, 20, 40, 60 iterations and indicates that our method (solid green) converges to a stable solution after 20 iterations and achieves better average performance on all domains than the baseline URL (dash green).

\paragraph{Layer analysis for adapters.}
Here we investigate whether it is sufficient to attach the adapters only to the later layers.
We evaluate this on ResNet18 which is composed of four blocks and attach the adapters to only later blocks (block4, block3,4, block2,3,4 and block-all, see \cref{fig:framework}). 
\Cref{fig:urllayers} shows that applying our adapters to only the last block (block4) obtains around 78\% average accuracy on all domains which outperforms the URL. With attaching residual adapters to more layers, the performance on unseen domains is improved significantly while the one on seen domains remains stable.

\paragraph{Decomposing residual adapters.}
Here we investigate whether one can reduce the number of parameters in the adapters while retaining its performance by using matrix decomposition (see \cref{sec:method}).
As in deep neural network, the adapters in earlier layers are relatively small, we then decompose the adapters in the last two blocks only where the adapter dimensionality goes up to $512\times 512$. \Cref{fig:urldecoml34} shows that our method can achieve good performance with less parameters by decomposing large residual adapters, (\eg when $N=32$ where the number of additional parameters equal to around 4\% vs 13\%, the performance is still comparable to the original form of residual adapters, \ie N=0). We refer to supplementary for more details.

\section{Conclusion and Limitations}\label{sec:con}
In this work, we investigate various strategies for adapting deep networks to few-shot classification tasks and show that light-weight adapters connected to a deep network with residual connections achieves strong adaptation to new tasks and domains only from few samples and obtains state-of-the-art performance while being efficient in the challenging Meta-Dataset benchmark. 
We demonstrate that the proposed solution can be incorporated to various feature extractors with a negligible increase in number of parameters.

Our method has limitations too. 
We build our method on existing backbones such ResNet-18 and ResNet-34, employ fixed adapter parameterizations and connection types which may not be optimal for every layer and task in multi-domain few-shot learning. 
Thus it would be desirable to have more flexible adapter structures that can be altered and tuned based on the target task.

\paragraph{Acknowledgments.} HB is supported by the EPSRC programme grant Visual AI EP/T028572/1.

\bibliographystyle{ieee_fullname}
\bibliography{ref}

\clearpage
\appendix

\section{Dataset}
Meta-Dataset~\cite{triantafillou2019meta} is a few-shot classification benchmark that initially consists of ten datasets: ILSVRC\_2012~\cite{russakovsky2015imagenet} (ImageNet),  Omniglot~\cite{Lake1332}, FGVC-Aircraft~\cite{maji2013fine} (Aircraft), CUB-200-2011~\cite{wah2011caltech} (Birds), Describable Textures~\cite{cimpoi2014describing} (DTD), QuickDraw~\cite{jongejan2016quick}, FGVCx Fungi~\cite{brigit2018fungi} (Fungi), VGG Flower~\cite{nilsback2008automated} (Flower), Traffic Signs~\cite{houben2013detection} and MSCOCO~\cite{lin2014microsoft} then further expands with MNIST~\cite{lecun1998gradient}, CIFAR-10~\cite{krizhevsky2009learning} and CIFAR-100~\cite{krizhevsky2009learning}. We follow the standard procedure in~\cite{triantafillou2019meta} and consider both the `Training on all datasets' (multi-domain learning) and `Training on ImageNet only' (single-domain learning) settings.
In `Training on all datasets' setting, we follow the standard procedure and use the first eight datasets for meta-training, in which each dataset is further divided into train, validation and test set with disjoint classes. 
While the evaluation within these datasets is used to measure the generalization ability in the seen domains, the remaining five datasets are reserved as unseen domains in meta-test for measuring the cross-domain generalization ability.
In `Training on ImageNet only' setting, we follow the standard procedure and only use train split of ImageNet for meta-training. The evaluation of models is in the test split of ImageNet and the rest 12 datasets which are reserved as unseen domains in meta-test.
As in~\cite{triantafillou2019meta}, we evaluate our method on 600 randomly sampled tasks for each dataset with varying number of ways and shots, and report average accuracy and 95\% confidence score in all experiments.

\section{Implementation details}

In this section, we explain the details of task-agnostic (feature extractor) learning and then task-specific (adapter) learning.

\subsection{Task-agnostic learning}

Here we consider learning the parameters of the feature extractor from either multiple or single domains.

\paragraph{Multi-domain learning.}
When we learn the feature extractor from multiple domains, we consider two cases.
In the first case, which we call vanilla multiple domain learning (or MDL), we design a deep network where we share all the layers across all domains and have domain-specific classifiers.
This setting corresponds to Eq~(1) in the main text.
Second we consider a variant of MDL, URL~\cite{li2021universal} which also involves learning a single network with shared and domain-specific layers as such, however, it is learned by distilling information from multiple domain-specific networks as described \cite{li2021universal}.
In these two settings, as in~\cite{dvornik2020selecting,bateni2020improved,li2021universal}, we build  MDL and URL on the ResNet-18~\cite{he2016deep} backbone and use $84\times 84$ image size.

For optimization of both MDL and URL, we follow the same protocol in~\cite{li2021universal}, use SGD optimizer and cosine annealing with a weight decay of $7\times 10^{-4}$  for learning 240,000 iterations. The learning rate is 0.03 and the annealing frequency is 48,000. As in~\cite{li2021universal}, the batch size for ImageNet is $64\times 7$ and is $64$ for the other 7 datasets. We refer readers to \cite{li2021universal} for more details.

\paragraph{Single domain learning (SDL).}

We also evaluate our method on a feature extractor that is learned on single domain which we call SDL.
Here we evaluate our method on two backbones, ResNet-18 (SDL-ResNet-18) and ResNet34 (SDL-ResNet-34).

\begin{table}[h!]
	\centering
    \resizebox{1.0\textwidth}{!}
    {
		\begin{tabular}{ccccc}
		    \toprule
		    Backbone & learning rate & batch size & annealing freq. & max. iter. \\
		    \midrule
		    SDL-ResNet-18 & $3\times 10^{-2}$ & 64 & 48,000 & 480,000\\
		    SDL-ResNet-34 & $3\times 10^{-2}$ & 128 & 48,000 & 480,000\\
			\bottomrule
		\end{tabular}%
			}
		\vspace{-0.25cm}
		\caption{Training hyper-parameters of single domain learning.}
		\label{supptab:hyperparams}
\end{table}%

\begin{table*}[ht]
	\centering
    \resizebox{1.0\textwidth}{!}
    {
		\begin{tabular}{lcccccccc|ccccc}

		    \toprule
		    Test Dataset & ImageNet & Omniglot & Aircraft & Birds & Textures & Quick Draw & Fungi & VGG Flower & Traffic Sign & MSCOCO & MNIST & CIFAR-10 & CIFAR-100\\
		    \midrule
		    MDL & $53.4 \pm 1.1$ & $93.8 \pm 0.4$ & $86.6 \pm 0.5$ & $78.6 \pm 0.8$ & $71.4 \pm 0.7$ & $81.5 \pm 0.6$ & $61.9 \pm 1.0$ & $88.7 \pm 0.6$ & $51.0 \pm 1.0$ & $49.7 \pm 1.1$ & $94.4 \pm 0.3$ & $66.7 \pm 0.8$ & $53.6 \pm 1.0$ \\
		    Ours (MDL) & ${\bf 55.6 \pm 1.0}$ & ${\bf 94.3 \pm 0.4}$ & ${\bf 86.7 \pm 0.5}$ & ${\bf 79.4 \pm 0.8}$ & ${\bf 73.2 \pm 0.8}$ & ${\bf 81.7 \pm 0.6}$ & ${\bf 64.0 \pm 0.9}$ & ${\bf 90.9 \pm 0.5}$ & ${\bf 81.1 \pm 0.9}$ & ${\bf 51.4 \pm 1.1}$ & ${\bf 96.9 \pm 0.3}$ & ${\bf 78.5 \pm 0.8}$ & ${\bf 64.3 \pm 1.1}$ \\
		    \midrule
		    URL~\cite{li2021universal} & $58.8 \pm 1.1$ & $94.5 \pm 0.4$ & $89.4 \pm 0.4$ & $80.7 \pm 0.8$ & $77.2 \pm 0.7$ & ${\bf 82.5 \pm 0.6}$ & ${\bf 68.1 \pm 0.9}$ & $92.0 \pm 0.5$ & $63.3 \pm 1.2$ & $57.3 \pm 1.0$ & $94.7 \pm 0.4$ & $74.2 \pm 0.8$ & $63.6 \pm 1.0$ \\
		    Ours (URL) & ${\bf 59.5 \pm 1.0}$ & ${\bf94.9 \pm 0.4}$ & ${\bf89.9 \pm 0.4}$ & ${\bf 81.1 \pm 0.8}$ & ${\bf77.5 \pm 0.7}$ & $81.7 \pm 0.6$ & $66.3 \pm 0.9$ & ${\bf 92.2 \pm 0.5}$ & ${\bf 82.8 \pm 1.0}$ & ${\bf 57.6 \pm 1.0}$ & ${\bf 96.7 \pm 0.4}$ & ${\bf 82.9 \pm 0.7}$ & ${\bf 70.4 \pm 1.0}$ \\
		    \midrule
		    SDL-ResNet-18 & $55.8 \pm 1.0$ & $67.4 \pm 1.2$ & $49.5 \pm 0.9$ & $71.2 \pm 0.9$ & $73.0 \pm 0.6$ & $53.9 \pm 1.0$ & $41.6 \pm 1.0$ & $87.0 \pm 0.6$ & $47.4 \pm 1.1$ & $53.5 \pm 1.0$ & $78.1 \pm 0.7$ & $67.3 \pm 0.8$ & $56.6 \pm 0.9$ \\
		    Ours (SDL-ResNet-18) & ${\bf 59.5 \pm 1.1}$ & ${\bf 78.2 \pm 1.2}$ & ${\bf 72.2 \pm 1.0}$ & ${\bf 74.9 \pm 0.9}$ & ${\bf 77.3 \pm 0.7}$ & ${\bf 67.6 \pm 0.9}$ & ${\bf 44.7 \pm 1.0}$ & ${\bf 90.9 \pm 0.6}$ & ${\bf 82.5 \pm 0.8}$ & ${\bf 59.0 \pm 1.0}$ & ${\bf 93.9 \pm 0.6}$ & ${\bf 82.1 \pm 0.7}$ & ${\bf 70.7 \pm 0.9}$ \\
		    \midrule
		    SDL-ResNet-34 & $62.2 \pm 1.1$ & $72.8 \pm 1.1$ & $62.9 \pm 0.9$ & $79.6 \pm 0.8$ & $75.6 \pm 0.6$ & $64.5 \pm 0.8$ & $47.4 \pm 1.1$ & $90.4 \pm 0.6$ & $54.8 \pm 1.0$ & $56.1 \pm 1.0$ & $79.3 \pm 0.6$ & $83.0 \pm 0.6$ & $74.8 \pm 0.8$  \\
		    Ours (SDL-ResNet-34) & ${\bf 63.7 \pm 1.0}$ & ${\bf 82.6 \pm 1.1}$ & ${\bf 80.1 \pm 1.0}$ & ${\bf 83.4 \pm 0.8}$ & ${\bf 79.6 \pm 0.7}$ & ${\bf 71.0 \pm 0.8}$ & ${\bf 51.4 \pm 1.2}$ & ${\bf 94.0 \pm 0.5}$ & ${\bf 81.7 \pm 0.9}$ & ${\bf 61.7 \pm 0.9}$ & ${\bf 94.6 \pm 0.5}$ & ${\bf 86.0 \pm 0.6}$ & ${\bf 78.3 \pm 0.8}$ \\
			\bottomrule
		\end{tabular}%
			}
		\vspace{-0.35cm}
		\caption{Results of attaching residual adapters to different baselines. `SDL-ResNet-18' is the single domain model with ResNet-18 backbone pretrained on ImageNet. `SDL-ResNet-34' is the single domain model with ResNet-34 backbone pretrained on ImageNet. `MDL' is a vanilla Multi-Domain Learning (MDL) model trained on eight seen datasets jointly.}
		\label{supptab:baselines}
\end{table*}%

\begin{table*}[t]
	\centering
    \resizebox{1.0\textwidth}{!}
    {
		\begin{tabular}{lcccccccccccccc|ccccc}

		    \toprule
		    \multirow{2}{*}{Test Dataset} & \multirow{2}{*}{classifier} & Aux-Net & serial or & M or & \multirow{2}{*}{$\beta$} & \multirow{2}{*}{\#params} & \multirow{2}{*}{ImageNet} & \multirow{2}{*}{Omniglot} & \multirow{2}{*}{Aircraft} & \multirow{2}{*}{Birds} & \multirow{2}{*}{Textures} & \multirow{2}{*}{Quick Draw} & \multirow{2}{*}{Fungi} & \multirow{2}{*}{VGG Flower} & \multirow{2}{*}{Traffic Sign} & \multirow{2}{*}{MSCOCO} & \multirow{2}{*}{MNIST} & \multirow{2}{*}{CIFAR-10} & \multirow{2}{*}{CIFAR-100}\\
		    & & or Ad & parallel & CW & & & & & & & & & & & & & & & \\
		    \midrule
		    NCC & NCC & - & - & - & \XSolidBrush &  - & $57.0\pm1.1$ & $94.4\pm0.4$ & $88.0\pm0.5$ & $80.3\pm0.7$ & $74.6\pm0.7$ & $81.8\pm0.6$ & $66.2\pm0.9$ & $91.5\pm0.5$ & $49.8\pm1.1$ & $54.1\pm1.0$ & $91.1\pm0.4$ & $70.6\pm0.7$& $59.1\pm1.0$\\ 
		    MD & MD & - & - & - & \XSolidBrush & - & $53.9\pm1.0$ & $93.8\pm0.5$ & $87.6\pm0.5$ & $78.3\pm0.7$ & $73.7\pm0.7$ & $80.9\pm0.7$ & $57.7\pm0.9$ & $89.7\pm0.6$ & $62.2\pm1.1$ & $48.5\pm1.0$ & $95.1\pm0.4$ & $68.9\pm0.8$ & $60.0\pm0.9$ \\ 
		    LR & LR & - & - & - & \XSolidBrush & - & $56.0\pm1.1$ & $93.7\pm0.5$ & $88.3\pm0.6$  & $79.7\pm0.8$ & $74.7\pm0.7$ & $80.0\pm0.7$ & $62.1\pm0.8$ & $91.1\pm0.5$ & $59.7\pm1.1$ & $51.2\pm1.1$ & $93.5\pm0.5$  & $73.1\pm0.8$ & $60.1\pm1.1$ \\
		    SVM & SVM & - & - & - & \XSolidBrush & - & $54.5\pm1.1$ & $94.3\pm0.5$ & $87.7\pm0.5$ & $78.1\pm0.8$ & $73.8\pm0.8$ & $80.0\pm0.6$ & $58.5\pm0.9$ & $91.4\pm0.6$ & $65.7\pm1.2$ & $50.5\pm1.0$ & $95.4\pm0.4$ & $72.0\pm0.8$ & $60.5\pm1.1$ \\
		    Softmax & Softmax & - & - & - & \XSolidBrush & - & $42.2 \pm 1.0$ & $85.3 \pm 0.7$ & $71.9 \pm 0.8$ & $59.6 \pm 1.0$ & $62.0 \pm 0.8$ & $61.2 \pm 1.0$ & $37.3 \pm 0.9$ & $66.7 \pm 1.0$ & $51.4 \pm 1.1$ & $48.2 \pm 1.1$ & $93.5 \pm 0.5$ & $70.4 \pm 0.8$ & $59.3 \pm 1.0$ \\
		    KNN & KNN & - & - & - & \XSolidBrush & - & $48.1 \pm 1.1$ & $94.1 \pm 0.4$ & $84.5 \pm 0.6$ & $70.7 \pm 0.8$ & $65.9 \pm 0.8$ & $74.8 \pm 0.7$ & $53.5 \pm 0.9$ & $86.0 \pm 0.6$ & $56.9 \pm 1.2$ & $44.7 \pm 1.1$ & $91.4 \pm 0.5$ & $60.3 \pm 0.8$ & $49.4 \pm 1.0$ \\
		    \midrule
		    PA & NCC & - & - & - & \Checkmark & - & $58.8 \pm 1.1$ & $94.5 \pm 0.4$ & $89.4 \pm 0.4$ & $80.7 \pm 0.8$ & $77.2 \pm 0.7$ & ${\bf 82.5 \pm 0.6}$ & ${\bf 68.1 \pm 0.9}$ & $92.0 \pm 0.5$ & $63.3 \pm 1.1$ & $57.3 \pm 1.0$ & $94.7 \pm 0.4$ & $74.2 \pm 0.8$ & $63.5 \pm 1.0$ \\
		    PA & Softmax & - & - & - & \Checkmark & - & $53.4 \pm 1.2$ & $92.7 \pm 0.5$ & $85.7 \pm 0.6$ & $76.1 \pm 0.9$ & $73.9 \pm 0.8$ & $76.5 \pm 0.8$ & $51.1 \pm 0.9$ & $86.9 \pm 0.7$ & $52.5 \pm 1.1$ & $48.2 \pm 1.1$ & $94.3 \pm 0.4$ & $69.7 \pm 0.8$ & $60.4 \pm 1.0$ \\
		    \midrule
		    Finetune & NCC & - & - & - & \XSolidBrush & - & $55.9 \pm 1.2$ & $94.0 \pm 0.5$ & $87.3 \pm 0.6$ & $77.8 \pm 0.9$ & $76.8 \pm 0.8$ & $75.3 \pm 0.9$ & $57.6 \pm 1.1$ & $91.5 \pm 0.6$ & ${\bf 86.1 \pm 0.9}$ & $53.1 \pm 1.2$ & ${\bf 96.8 \pm 0.4}$ & $80.9 \pm 0.8$ & $65.9 \pm 1.1$ \\
		    Finetune & Softmax & - & - & - & \XSolidBrush & - & $48.4 \pm 1.2$ & $92.2 \pm 0.6$ & $81.6 \pm 0.9$ & $70.3 \pm 1.3$ & $72.0 \pm 0.9$ & $73.5 \pm 1.0$ & $44.2 \pm 1.1$ & $90.3 \pm 0.7$ & $65.5 \pm 1.4$ & $41.0 \pm 1.3$ & $96.3 \pm 0.4$ & $71.6 \pm 1.0$ & $53.8 \pm 1.4$ \\
		    \midrule
		    Aux-S-CW & NCC & Aux-Net & serial & CW & \XSolidBrush & - & $54.6\pm1.1$ & $93.5\pm0.5$ &  $86.6\pm0.5$ & $78.6\pm0.8$ &  $71.5\pm0.7$ &   $79.3\pm0.6$ &  $66.0\pm0.9$ &   $87.6\pm0.6$ &  $43.3\pm0.9$ & $49.1\pm1.0$ & $87.9\pm0.5$ &  $62.8\pm0.8$ &  $51.5\pm1.0$  \\
		    Aux-R-CW & NCC & Aux-Net & residual & CW & \XSolidBrush & - & $56.1\pm1.1$ &  $94.2\pm0.4$ & $88.4\pm0.5$ & $80.6\pm0.7$ & $74.9\pm0.6$ &  $82.0\pm0.6$ & $66.4\pm0.9$ & $91.6\pm0.5$ & $48.5\pm1.0$ & $53.5\pm1.0$ & $90.8\pm0.5$ &  $70.2\pm0.8$ & $59.7\pm1.0$ \\
		    Aux-S-CW & MD & Aux-Net & serial & CW & \XSolidBrush & - & $55.1\pm1.1$ &$93.8\pm0.5$ &$86.8\pm0.5$ & $77.4\pm0.8$ & $73.2\pm0.8$ &$79.9\pm0.7$ &$57.4\pm0.9$ & $88.1\pm0.7$ & $58.4\pm1.1$ & $50.1\pm1.1$ & $92.7\pm0.5$ & $66.5\pm0.8$ & $55.7\pm1.1$ \\
		    Aux-R-CW & MD & Aux-Net & residual & CW & \XSolidBrush & - & $54.8\pm1.1$ & $93.8\pm0.5$ & $87.4\pm0.5$ & $78.2\pm0.7$ & $73.4\pm0.7$ & $81.1\pm0.7$ & $58.8\pm0.9$ & $90.1\pm0.5$ & $63.6\pm1.2$ & $48.5\pm1.1$ & $94.8\pm0.4$ & $69.6\pm0.8$ & $60.6\pm0.9$ \\
		    \midrule
		    Ad-S-CW & NCC & Ad & serial & CW & \XSolidBrush & 0.06\% & $56.8 \pm 1.1$ & $94.8 \pm 0.4$ & $89.3 \pm 0.5$ & $80.7 \pm 0.7$ & $74.5 \pm 0.7$ & $81.6 \pm 0.6$ & $65.8 \pm 0.9$ & $91.3 \pm 0.5$ & $73.9 \pm 1.1$ & $53.6 \pm 1.1$ & $95.7 \pm 0.4$ & $78.4 \pm 0.7$ & $64.3 \pm 1.0$ \\
		    Ad-R-CW & NCC & Ad & residual & CW & \XSolidBrush & 1.57\%   & $57.6 \pm 1.1$ & $94.7 \pm 0.4$ & $89.0 \pm 0.4$ & $81.2 \pm 0.8$ & $75.2 \pm 0.7$ & $81.5 \pm 0.6$ & $65.4 \pm 0.8$ & $91.8 \pm 0.5$ & $79.2 \pm 1.1$ & $54.7 \pm 1.1$ & $96.4 \pm 0.4$ & $79.5 \pm 0.8$ & $67.4 \pm 1.0$ \\
		    Ad-S-M & NCC & Ad & serial & M & \XSolidBrush & 12.50\% & $56.2 \pm 1.1$ & $94.4 \pm 0.4$ & $89.1 \pm 0.5$ & $80.6 \pm 0.7$ & $75.8 \pm 0.7$ & $81.6 \pm 0.6$ & $67.1 \pm 0.9$ & $92.1 \pm 0.4$ & $67.6 \pm 1.2$ & $54.8 \pm 1.1$ & $95.9 \pm 0.4$ & $78.9 \pm 0.7$ & $66.6 \pm 1.1$ \\
		    Ad-R-M & NCC & Ad & residual & M & \XSolidBrush & 10.93\%  & $57.3 \pm 1.1$ & $94.9 \pm 0.4$ & $88.9 \pm 0.5$ & $81.0 \pm 0.7$ & $76.7 \pm 0.7$ & $80.6 \pm 0.6$ & $65.4 \pm 0.9$ & $91.4 \pm 0.5$ & $82.6 \pm 1.0$ & $55.0 \pm 1.1$ & $96.6 \pm 0.4$ & $82.1 \pm 0.7$ & $66.4 \pm 1.1$ \\
		    \midrule
		    Ad-R-CW-PA & NCC & Ad & residual & CW & \Checkmark & 3.91\% & $58.6 \pm 1.1$ & $94.5 \pm 0.4$ & ${\bf 90.0 \pm 0.4}$ & $80.5 \pm 0.8$ & ${\bf 77.6 \pm 0.7}$ & $81.9 \pm 0.6$ & $67.0 \pm 0.9$ & $92.2 \pm 0.5$ & $80.2 \pm 0.9$ & $57.2 \pm 1.0$ & $96.1 \pm 0.4$ & $81.5 \pm 0.8$ & ${\bf 71.4 \pm 0.9}$ \\
		    Ad-R-M-PA & NCC & Ad & residual & M & \Checkmark & 13.27\% & ${\bf 59.5 \pm 1.0}$ & ${\bf94.9 \pm 0.4}$ & $89.9 \pm 0.4$ & ${\bf 81.1 \pm 0.8}$ & $77.5 \pm 0.7$ & $81.7 \pm 0.6$ & $66.3 \pm 0.9$ & ${\bf 92.2 \pm 0.5}$ & $82.8 \pm 1.0$ & ${\bf 57.6 \pm 1.0}$ & $96.7 \pm 0.4$ & ${\bf 82.9 \pm 0.7}$ & $70.4 \pm 1.0$ \\
			\bottomrule
		\end{tabular}%
			}
		\vspace{-0.25cm}
		\caption{Comparisons to methods that learn classifiers and model adaptation methods during meta-test stage based on URL model. NCC, MD, LR, SVM, Softmax, KNN denote nearest centroid classifier, Mahalanobis distance, logistic regression, support vector machines, softmax classifier and k-nearest neighbors classifier respectively. PA indicates pre-classifier alignment. `Aux-Net or Ad' indicates using Auxiliary Network to predict $\alpha$ or attaching adapter $\alpha$ directly. `M or CW' means using matrix multiplication or channel-wise scaling adapters. 'S' and 'R' denote serial adapter and residual adapter, respectively. `$\beta$' indicates using the pre-classifier adaptation. Mean accuracy, 95\% confidence interval are reported. The first eight datasets are seen during training and the last five datasets are unseen and used for test only.}
		\label{supptab:testad}
\end{table*}%

\begin{table*}[t]
	\centering
	    \resizebox{0.95\textwidth}{!}
    {
		\begin{tabular}{cccccc|ccccc}
			& \multicolumn{4}{c}{Varying-Way Five-Shot} & \multicolumn{4}{c}{Five-Way One-Shot} \\
		    \toprule
		    \multirow{2}{*}{Test Dataset} & Simple & SUR & URT & URL & \multirow{2}{*}{Ours}& Simple & SUR & URT & URL & \multirow{2}{*}{Ours}\\
		    &  CNAPS~\cite{bateni2020improved}  & \cite{dvornik2020selecting} & \cite{liu2020universal} & \cite{li2021universal} & & CNAPS~\cite{bateni2020improved} & \cite{dvornik2020selecting}& \cite{liu2020universal} & \cite{li2021universal} & \\
		    \midrule
		    ImageNet & $47.2\pm1.0$& $46.7\pm1.0$& $48.6\pm1.0$& ${\bf 49.4 \pm 1.0}$ & $48.3 \pm 1.0$ & $42.6\pm0.9$& $40.7\pm1.0$& $47.4\pm1.0$& ${\bf 49.6 \pm 1.1}$ & $48.0 \pm 1.0$ \\
		    Omniglot & $95.1\pm0.3$& $95.8\pm0.3$& $96.0\pm0.3$& $96.0 \pm 0.3$ & ${\bf 96.8 \pm 0.3}$ & $93.1\pm0.5$& $93.0\pm0.7$& $95.6\pm0.5$& $95.8 \pm 0.5$ & ${\bf 96.3 \pm 0.4}$ \\
		    Aircraft & $74.6\pm0.6$& $82.1\pm0.6$& $81.2\pm0.6$& $84.8 \pm 0.5$ & ${\bf 85.5 \pm 0.5}$  & $65.8\pm0.9$&  $67.1\pm1.4$& $77.9\pm0.9$& ${\bf 79.6 \pm 0.9}$ & ${\bf 79.6 \pm 0.9}$ \\
		    Birds & $69.6\pm0.7$& $62.8\pm0.9$& $71.2\pm0.7$& $76.0 \pm 0.6$ & ${\bf 76.6 \pm 0.6}$  &  $67.9\pm0.9$& $59.2\pm1.0$& $70.9\pm0.9$& ${\bf 74.9 \pm 0.9}$ & $74.5 \pm 0.9$ \\
		    Textures & $57.5\pm0.7$& $60.2\pm0.7$& $65.2\pm0.7$& ${\bf 69.1 \pm 0.6}$ & $68.3 \pm 0.7$  &$42.2\pm0.8$& $42.5\pm0.8$& $49.4\pm0.9$& $53.6 \pm 0.9$ & ${\bf 54.5 \pm 0.9}$ \\
		    Quick Draw & $70.9\pm0.6$& $79.0\pm0.5$& ${\bf 79.2\pm0.5}$& $78.2 \pm 0.5$ & $77.9 \pm 0.6$  &$70.5\pm0.9$& ${\bf 79.8\pm0.9}$& $79.6\pm0.9$& $79.0 \pm 0.8$ & $79.3 \pm 0.9$ \\
		    Fungi & $50.3\pm1.0$& $66.5\pm0.8$& $66.9\pm0.9$& $70.0 \pm 0.8$ & ${\bf 70.4 \pm 0.8}$  & $58.3\pm1.1$& $64.8\pm1.1$& $71.0\pm1.0$& $75.2 \pm 1.0$ & ${\bf 75.3 \pm 1.0}$ \\
		    VGG Flower & $86.5\pm0.4$& $76.9\pm0.6$& $82.4\pm0.5$& $89.3 \pm 0.4$ & ${\bf 89.5 \pm 0.4}$  & $79.9\pm0.7$& $65.0\pm1.0$& $72.7\pm0.0$& $79.9 \pm 0.8$ & ${\bf 80.3 \pm 0.8}$ \\
		    \midrule
		    Traffic Sign & $55.2\pm0.8$& $44.9\pm0.9$& $45.1\pm0.9$& $57.5 \pm 0.8$ & ${\bf 72.3 \pm 0.6}$ & $55.3\pm0.9$& $44.6\pm0.9$& $52.7\pm0.9$& ${\bf 57.9 \pm 0.9}$ & $57.2 \pm 1.0$ \\
		    MSCOCO & $49.2\pm0.8$& $48.1\pm0.9$& $52.3\pm0.9$& ${\bf 56.1 \pm 0.8}$ & $56.0 \pm 0.8$ & $48.8\pm0.9$& $47.8\pm1.1$& $56.9\pm1.1$& $59.2 \pm 1.0$ & ${\bf 59.9 \pm 1.0}$ \\
		    MNIST & $88.9\pm0.4$& $90.1\pm0.4$& $86.5\pm0.5$& $89.7 \pm 0.4$ & ${\bf 92.5 \pm 0.4}$  & ${\bf 80.1\pm0.9}$& $77.1\pm0.9$& $75.6\pm0.9$& $78.7 \pm 0.9$ & ${\bf 80.1 \pm 0.9}$ \\
		    CIFAR-10 & $66.1\pm0.7$& $50.3\pm1.0$& $61.4\pm0.7$& $66.0 \pm 0.7$ & ${\bf 72.0 \pm 0.7}$  & $50.3\pm0.9$& $35.8\pm0.8$& $47.3\pm0.9$& $54.7 \pm 0.9$ & ${\bf 55.8 \pm 0.9}$ \\
		    CIFAR-100 & $53.8\pm0.9$& $46.4\pm0.9$& $52.5\pm0.9$& $57.0 \pm 0.9$ & ${\bf 64.1 \pm 0.8}$  & $53.8\pm0.9$& $42.9\pm1.0$& $54.9\pm1.1$& $61.8 \pm 1.0$ & ${\bf 63.7 \pm 1.0}$ \\
		    \midrule
		    Average Seen & $69.0$ & $71.2$ & $73.8$ & $76.6$ & ${\bf 76.7}$ & $65.0$ & $64.0$ & $70.6$ & $73.4$ & ${\bf 73.5}$ \\
		    Average Unseen & $62.6$ & $56.0$ & $59.6$ & $65.2$ & ${\bf 71.4}$ & $57.7$ & $49.6$ & $57.5$ & $62.4$ & ${\bf 63.4}$ \\
		    Average All & $66.5$ & $65.4$ & $68.3$ & $72.2$ & ${\bf 74.6}$ & $62.2$ & $58.5$ & $65.5$ & $69.2$ & ${\bf 69.6}$ \\
		    \midrule
		    Average Rank & $4.1$ & $3.9$ & $3.4$ & $2.1$ & ${\bf 1.5}$ & $3.8$ & $4.5$ & $3.3$ & ${\bf 1.7}$ & ${\bf 1.7}$ \\
			\bottomrule
		\end{tabular}%
			}
		\vspace{-0.25cm}
		\caption{Results of Varying-Way Five-Shot and Five-Way One-Shot scenarios. Mean accuracy, 95\% confidence interval are reported.}
		\label{supptab:fixedshot}
\end{table*}%

\begin{table*}[h!]
	\centering
    \resizebox{0.95\textwidth}{!}
    {
		\begin{tabular}{ccccccccccccc}

		    \toprule
		    Test Dataset & CNAPS~\cite{requeima2019fast} & Simple CNAPS~\cite{bateni2020improved} & TransductiveCNAPS~\cite{bateni2020enhancing} & SUR~\cite{dvornik2020selecting} & URT~\cite{liu2020universal} & FLUTE~\cite{triantafillou2021flute} & tri-M~\cite{liu2021multi} & URL~\cite{li2021universal} & Ours\\
		    \midrule
			ImageNet & $50.8 \pm 1.1$ & $56.5 \pm 1.1$ & $57.9 \pm 1.1$ & $54.5 \pm 1.1$ & $55.0 \pm 1.1$ & $51.8 \pm 1.1$ & ${\bf 58.6 \pm 1.0}$ & $57.5 \pm 1.1$ & $57.4 \pm 1.1$ \\
			Omniglot & $91.7 \pm 0.5$ & $91.9 \pm 0.6$ & $94.3 \pm 0.4$ & $93.0 \pm 0.5$ & $93.3 \pm 0.5$ & $93.2 \pm 0.5$ & $92.0 \pm 0.6$ & $94.5 \pm 0.4$ & ${\bf 95.0 \pm 0.4}$ \\
			Aircraft & $83.7 \pm 0.6$ & $83.8 \pm 0.6$ & $84.7 \pm 0.5$ & $84.3 \pm 0.5$ & $84.5 \pm 0.6$ & $87.2 \pm 0.5$ & $82.8 \pm 0.7$ & $88.6 \pm 0.5$ & ${\bf 89.3 \pm 0.4}$ \\
			Birds & $73.6 \pm 0.9$ & $76.1 \pm 0.9$ & $78.8 \pm 0.7$ & $70.4 \pm 1.1$ & $75.8 \pm 0.8$ & $79.2 \pm 0.8$ & $75.3 \pm 0.8$ & $80.5 \pm 0.7$ & ${\bf 81.4 \pm 0.7}$ \\
			Textures & $59.5 \pm 0.7$ & $70.0 \pm 0.8$ & $66.2 \pm 0.8$ & $70.5 \pm 0.7$ & $70.6 \pm 0.7$ & $68.8 \pm 0.8$ & $71.2 \pm 0.8$ & $76.2 \pm 0.7$ & ${\bf 76.7 \pm 0.7}$ \\
			Quick Draw & $74.7 \pm 0.8$ & $78.3 \pm 0.7$ & $77.9 \pm 0.6$ & $81.6 \pm 0.6$ & ${\bf 82.1 \pm 0.6}$ & $79.5 \pm 0.7$ & $77.3 \pm 0.7$ & $81.9 \pm 0.6$ & $82.0 \pm 0.6$ \\
			Fungi & $50.2 \pm 1.1$ & $49.1 \pm 1.2$ & $48.9 \pm 1.2$ & $65.0 \pm 1.0$ & $63.7 \pm 1.0$ & $58.1 \pm 1.1$ & $48.5 \pm 1.0$ & ${\bf 68.8 \pm 0.9}$ & $67.4 \pm 1.0$ \\
			VGG Flower & $88.9 \pm 0.5$ & $91.3 \pm 0.6$ & ${\bf 92.3 \pm 0.4}$ & $82.2 \pm 0.8$ & $88.3 \pm 0.6$ & $91.6 \pm 0.6$ & $90.5 \pm 0.5$ & $92.1 \pm 0.5$ & $92.2 \pm 0.5$ \\
			\midrule
			Traffic Sign & $56.5 \pm 1.1$ & $59.2 \pm 1.0$ & $59.7 \pm 1.1$ & $49.8 \pm 1.1$ & $50.1 \pm 1.1$ & $58.4 \pm 1.1$ & $63.0 \pm 1.0$ & $63.3 \pm 1.2$ & ${\bf 83.5 \pm 0.9}$ \\
			MSCOCO & $39.4 \pm 1.0$ & $42.4 \pm 1.1$ & $42.5 \pm 1.1$ & $49.4 \pm 1.1$ & $48.9 \pm 1.1$ & $50.0 \pm 1.0$ & $52.8 \pm 1.1$ & $54.0 \pm 1.0$ & ${\bf 55.8 \pm 1.1}$ \\
			MNIST & - & $94.3 \pm 0.4$ & $94.7 \pm 0.3$ & $94.9 \pm 0.4$ & $90.5 \pm 0.4$ & $95.6 \pm 0.5$ & $96.2 \pm 0.3$ & $94.5 \pm 0.5$ & ${\bf 96.7 \pm 0.4}$ \\
			CIFAR-10 & - & $72.0 \pm 0.8$ & $73.6 \pm 0.7$ & $64.2 \pm 0.9$ & $65.1 \pm 0.8$ & $78.6 \pm 0.7$ & $75.4 \pm 0.8$ & $71.9 \pm 0.7$ & ${\bf 80.6 \pm 0.8}$ \\
			CIFAR-100 & - & $60.9 \pm 1.1$ & $61.8 \pm 1.0$ & $57.1 \pm 1.1$ & $57.2 \pm 1.0$ & $67.1 \pm 1.0$ & $62.0 \pm 1.0$ & $62.6 \pm 1.0$ & ${\bf 69.6 \pm 1.0}$ \\
			\midrule
			Average Seen & $71.6$ & $74.6$ & $75.1$ & $75.2$ & $76.7$ & $76.2$ & $74.5$ & $80.0$ & $80.2$ \\
			Average Unseen & - & $65.8$ & $66.5$ & $63.1$ & $62.4$ & $69.9$ & $69.9$ & $69.3$ & $77.2$ \\
			Average All & - & $71.2$ & $71.8$ & $70.5$ & $71.2$ & $73.8$ & $72.7$ & $75.9$ & $79.0$ \\
			\midrule
			Average Rank & - & $6.3$ & $4.9$ & $5.8$ & $5.7$ & $4.3$ & $4.8$ & $2.7$ & $1.5$ \\
			\bottomrule
		\end{tabular}%
			}
		\vspace{-0.35cm}
		\caption{Comparison state-of-the-art methods on Meta-Dataset (using a multi-domain feature extractor of \cite{li2021universal}). Mean accuracy, 95\% confidence interval are reported. The first eight datasets are seen during training and the last five datasets are unseen and used for test only.}
		\label{supptab:currmethod}
\end{table*}%

\begin{table*}[t]
	\centering
    \resizebox{1.0\textwidth}{!}
    {
		\begin{tabular}{lcccccccc|ccccc}

		    \toprule
		    Test Dataset & ImageNet & Omniglot & Aircraft & Birds & Textures & Quick Draw & Fungi & VGG Flower & Traffic Sign & MSCOCO & MNIST & CIFAR-10 & CIFAR-100\\
		    \midrule
		    10 iterations & $55.5 \pm 1.1$ & $93.9 \pm 0.5$ & $86.4 \pm 0.5$ & $78.6 \pm 0.7$ & $73.3 \pm 0.7$ & $81.9 \pm 0.6$ & $63.1 \pm 0.9$ & $90.3 \pm 0.5$ & $77.6 \pm 1.0$ & $50.6 \pm 1.1$ & $96.9 \pm 0.3$ & $77.0 \pm 0.8$ & $62.6 \pm 1.1$ \\
		    20 iterations & $56.2 \pm 1.1$ & $94.7 \pm 0.4$ & $86.3 \pm 0.5$ & $78.3 \pm 0.8$ & $73.9 \pm 0.7$ & $81.6 \pm 0.6$ & $63.4 \pm 0.9$ & $90.1 \pm 0.6$ & $79.4 \pm 1.0$ & $52.8 \pm 1.1$ & $97.2 \pm 0.3$ & $78.6 \pm 0.8$ & $65.9 \pm 1.1$ \\
		    40 iterations & $55.6 \pm 1.0$ & $94.3 \pm 0.4$ & $86.7 \pm 0.5$ & $79.4 \pm 0.8$ & $73.2 \pm 0.8$ & $81.7 \pm 0.6$ & $64.0 \pm 0.9$ & $90.9 \pm 0.5$ & $81.1 \pm 0.9$ & $51.4 \pm 1.1$ & $96.9 \pm 0.3$ & $78.5 \pm 0.8$ & $64.3 \pm 1.1$ \\
		    60 iterations & $55.9 \pm 1.1$ & $95.1 \pm 0.4$ & $85.9 \pm 0.6$ & $77.5 \pm 0.8$ & $74.7 \pm 0.7$ & $80.9 \pm 0.6$ & $62.1 \pm 0.9$ & $90.7 \pm 0.6$ & $82.2 \pm 0.9$ & $52.2 \pm 1.1$ & $97.0 \pm 0.4$ & $78.4 \pm 0.8$ & $64.4 \pm 1.1$ \\
			\bottomrule
		\end{tabular}%
			}
		\vspace{-0.25cm}
		\caption{Sensitivity of performance to number of iterations based on MDL model.}
		\label{supptab:mtliterations}
\end{table*}%

\paragraph{SDL-ResNet-18.}
Following~\cite{triantafillou2019meta,dvornik2020selecting,li2021universal}, we train a ResNet-18 on the train split of ImageNet and use $84\times 84$ image size, which is denoted as SDL-ResNet-18.
For optimization, we follow the training protocol in \cite{dvornik2020selecting,li2021universal}. Specifically, we use SGD optimizer and cosine annealing for all experiments with a momentum of 0.9 and a weight decay of $7\times 10^{-4}$. 
Some other hyperparameters are shown in \cref{supptab:hyperparams} as in~\cite{dvornik2020selecting,li2021universal}. To regularize training, we also use the exact same data augmentations as in \cite{dvornik2020selecting,li2021universal}, \eg random crops and random color augmentations. 

\paragraph{SDL-ResNet-34.}
We also apply our method to the single domain learning model with ResNet-34 backbone learned on ImageNet only as in~\cite{doersch2020crosstransformers}. We follow \cite{doersch2020crosstransformers} and use higher-resolution ($224\times 224$) images for meta-training and meta-testing. For optimization, we follow the training protocol as in~\cite{dvornik2020selecting,li2021universal}. Specifically, we use SGD optimizer and cosine annealing with a momentum of 0.9, a weight decay of $1\times 10^{-4}$ with a batch size of 128. Other hyperparameters are the same as in SDL-ResNet-18 and are shown in \cref{supptab:hyperparams}.
To regularize training, we also use the exact same data augmentations as in \cite{dvornik2020selecting,li2021universal}, \eg random crops and random color augmentations with an additional stage that randomly downsamples and upsamples images as in~\cite{doersch2020crosstransformers}.

\subsection{Task-specific learning}

\paragraph{Attaching and learning adapters.}
For the optimization of the adaptation parameters $\alpha$ which is attached directly and learned on support set and the pre-classifier adaptation $\beta$, we follow the optimization strategy in~\cite{li2021universal}, initialize $\beta$ as an identity matrix and optimize both $\alpha$ and $\beta$ for 40 iterations using Adadelta~\cite{zeiler2012adadelta} as optimizer. The learning rate of $\beta$ is 0.1 for first eight datasets and 1 for the last five datasets as in~\cite{li2021universal} and we set the learning rate of $\alpha$ as half of the learning rate of $\beta$, \ie 0.05 for the first eight datasets and 0.5 for the last five datasets. Note that, we learn $\alpha$ and $\beta$ on a per-task basis using the task's support set during meta-test. That is, $\alpha$ and $\beta$ are not re-used across the test tasks drawn from $\mathcal{D}_t$.

\paragraph{Predicting $r_{\alpha}$.}
In case of modulating $\alpha$ with the auxiliary network, we follow the auxiliary training protocols in~\cite{bateni2020improved}. We train for 10K episodes to optimize the task encoder using Adam with a learning rate of $1\times 10^{-5}$ on eight training domains in meta-train. We validate every 5K iterations to save the best model for test.

\begin{table*}[h!]
	\centering
    \resizebox{1.0\textwidth}{!}
    {
		\begin{tabular}{lcccccccc|ccccc}

		    \toprule
		    Test Dataset & ImageNet & Omniglot & Aircraft & Birds & Textures & Quick Draw & Fungi & VGG Flower & Traffic Sign & MSCOCO & MNIST & CIFAR-10 & CIFAR-100\\
		    \midrule
		    10 iterations & $58.4 \pm 1.1$ & $94.8 \pm 0.4$ & $89.9 \pm 0.4$ & $81.3 \pm 0.7$ & $76.6 \pm 0.7$ & $81.8 \pm 0.6$ & $68.4 \pm 0.9$ & $92.5 \pm 0.5$ & $76.5 \pm 1.1$ & $55.6 \pm 1.1$ & $96.4 \pm 0.4$ & $79.0 \pm 0.7$ & $66.9 \pm 1.0$ \\
		    20 iterations & $58.2 \pm 1.1$ & $94.8 \pm 0.4$ & $89.9 \pm 0.4$ & $81.1 \pm 0.7$ & $77.5 \pm 0.8$ & $81.9 \pm 0.6$ & $68.0 \pm 0.9$ & $92.4 \pm 0.5$ & $81.8 \pm 1.0$ & $57.8 \pm 1.1$ & $96.7 \pm 0.4$ & $81.7 \pm 0.8$ & $69.1 \pm 0.9$ \\
		    40 iterations & $59.5 \pm 1.0$ & $94.9 \pm 0.4$ & $89.9 \pm 0.4$ & $81.1 \pm 0.8$ & $77.5 \pm 0.7$ & $81.7 \pm 0.6$ & $66.3 \pm 0.9$ & $92.2 \pm 0.5$ & $82.8 \pm 1.0$ & $57.6 \pm 1.0$ & $96.7 \pm 0.4$ & $82.9 \pm 0.7$ & $70.4 \pm 1.0$ \\
		    60 iterations & $58.7 \pm 1.1$ & $94.9 \pm 0.4$ & $89.5 \pm 0.5$ & $80.8 \pm 0.7$ & $77.4 \pm 0.8$ & $81.8 \pm 0.6$ & $66.2 \pm 0.9$ & $92.5 \pm 0.5$ & $83.7 \pm 0.9$ & $56.9 \pm 1.0$ & $96.6 \pm 0.3$ & $82.0 \pm 0.8$ & $72.0 \pm 0.9$ \\
			\bottomrule
		\end{tabular}%
			}
		\vspace{-0.25cm}
		\caption{Sensitivity of performance to number of iterations based on URL model.}
		\label{supptab:urliterations}
\end{table*}%

\section{More results}

\subsection{Our method with different feature extractors} 
\cref{supptab:baselines} shows the results of our method (the proposed residual adapters in matrix form) when incorporated to different feature extractors, single domain model with ResNet-18 backbone (SDL-ResNet-18) pre-trained on ImageNet, single domain model with ResNet-34 (SDL-ResNet-34) pre-trained on ImageNet, vanilla multi-domain learning (MDL) and URL~\cite{li2021universal}. We see that attaching and learning residual adapters can significantly improve the performance on all domains over SDL-ResNet-18, SDL-ResNet-34 and MDL and obtain better performance on most domains over URL (11 out of 13 domains). This strongly indicates that our method can efficiently adapt the model for unseen categories and domains with few support samples while being agnostic to the feature extractor with different backbone and resolution of images.

\subsection{Task-specific parameterizations}

In \cref{supptab:testad}, we report additional 95\% confidence interval of each dataset to the main paper for the comparison of different $r_{\alpha}$ choices based on the URL model. The first eight datasets are seen during training and the last five datasets are unseen and used for test only. We can see that the confidence intervals for different methods have marginal differences.

\subsection{Varying-way 5-shot and 5-way-1-shot}

In the main paper, we only report the average accuracy of Varying-Way Five-Shot and Five-Way One-Shot scenarios due to limited space, and detailed results are depicted in \cref{supptab:fixedshot}. In the table, we report the Mean accuracy, 95\% confidence interval of each dataset. The first eight datasets are seen during training and the last five datasets are unseen and used for test only. URT and URL are two strong baselines surpassing both Simple CNAPS and SUR, while Ours outperforms them on most datasets, especially on unseen domains.

\subsection{Results evaluated with updated evaluation protocol.}
As the code from Meta-dataset has been updated, we evaluate all methods with the updated evaluation protocol from the Meta-dataset~\footnote{As mentioned in \url{https://github.com/google-research/meta-dataset/issues/54}, we also set the shuffle\_buffer\_size as 1000 to evaluate all methods and report the results in \cref{supptab:currmethod}. This change does not affect much on the results as the datasets we used were shuffled using the latest data convert code from \href{https://github.com/google-research/meta-dataset}{Meta-Dataset}.} and report the results~\footnote{The results of Simple CNAPS~\cite{bateni2020improved} and Transductive CNAPS~\cite{bateni2020enhancing} are reproduced by the authors and reported at \url{https://github.com/peymanbateni/simple-cnaps}. Results of FLUTE~\cite{triantafillou2021flute} and tri-M~\cite{liu2021multi} are from their papers. We reproduce the results of SUR~\cite{dvornik2020selecting} and URT~\cite{liu2020universal} with the updated evaluation protocol for fair comparison.} in \cref{supptab:currmethod}. As shown in \cref{supptab:currmethod}, the update does not affect much on the results and our method rank 1.5 in average and the state-of-the-art method URL rank 2.7. Our method outperforms other methods on most domains (9 out of 13), especially obtaining significant improvement on 5 unseen datasets than the second best method, \ie Average Unseen (+7.9). More specifically, our method obtains significant better results than the second best approach (URL) on Traffic Sign (+20.2), CIFAR-10 (+8.7), and CIFAR-100 (+7.0).

\subsection{Ablation study}
\begin{figure}[h]
\begin{center}
\includegraphics[width=0.9\linewidth]{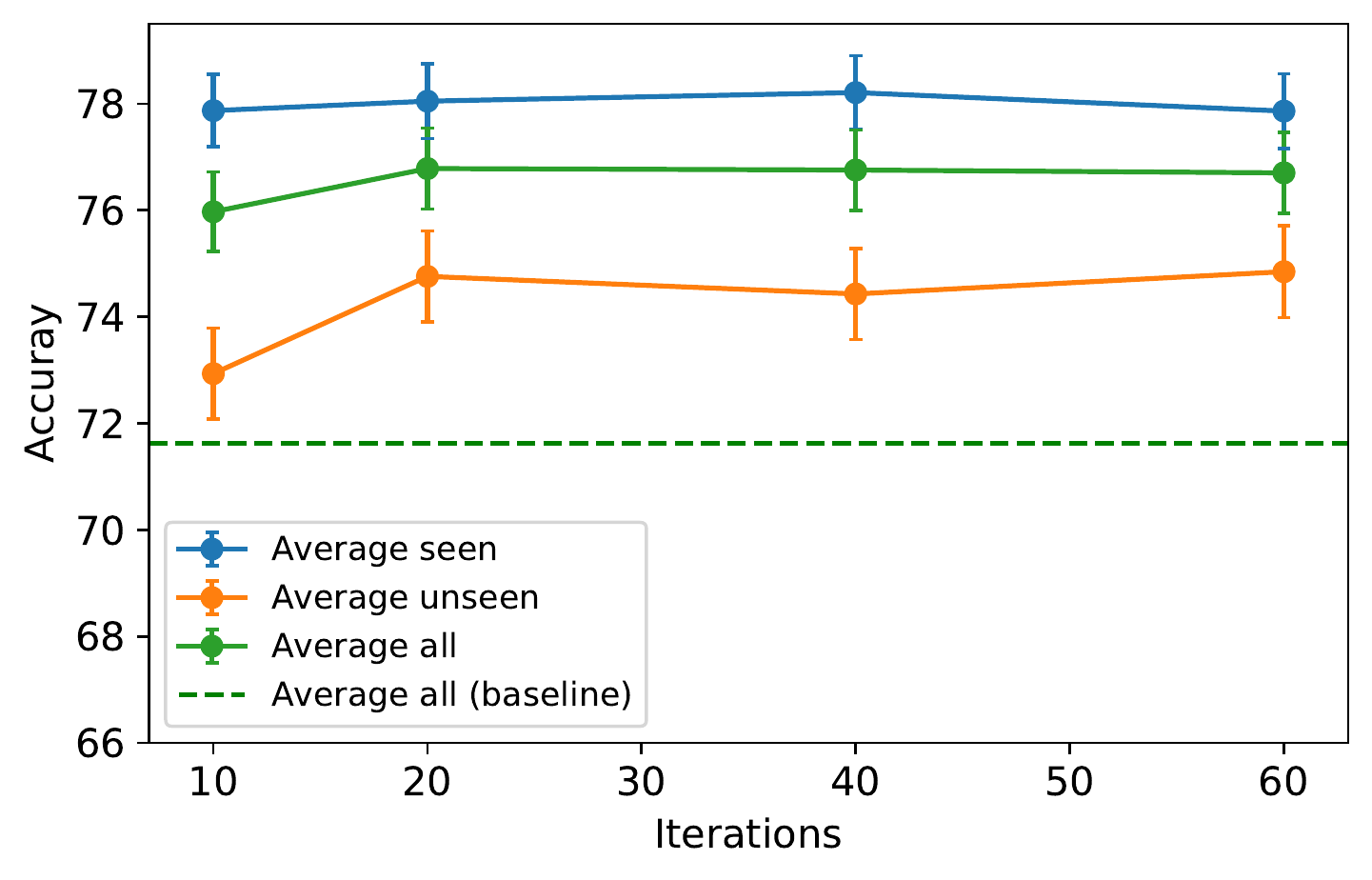}
\end{center}
\vspace{-0.3in}
\caption{Sensitivity of performance to number of iterations based on MDL model.}
\label{suppfig:mdlstab}
\end{figure}

\begin{figure}[h]
\begin{center}

\includegraphics[width=0.9\linewidth]{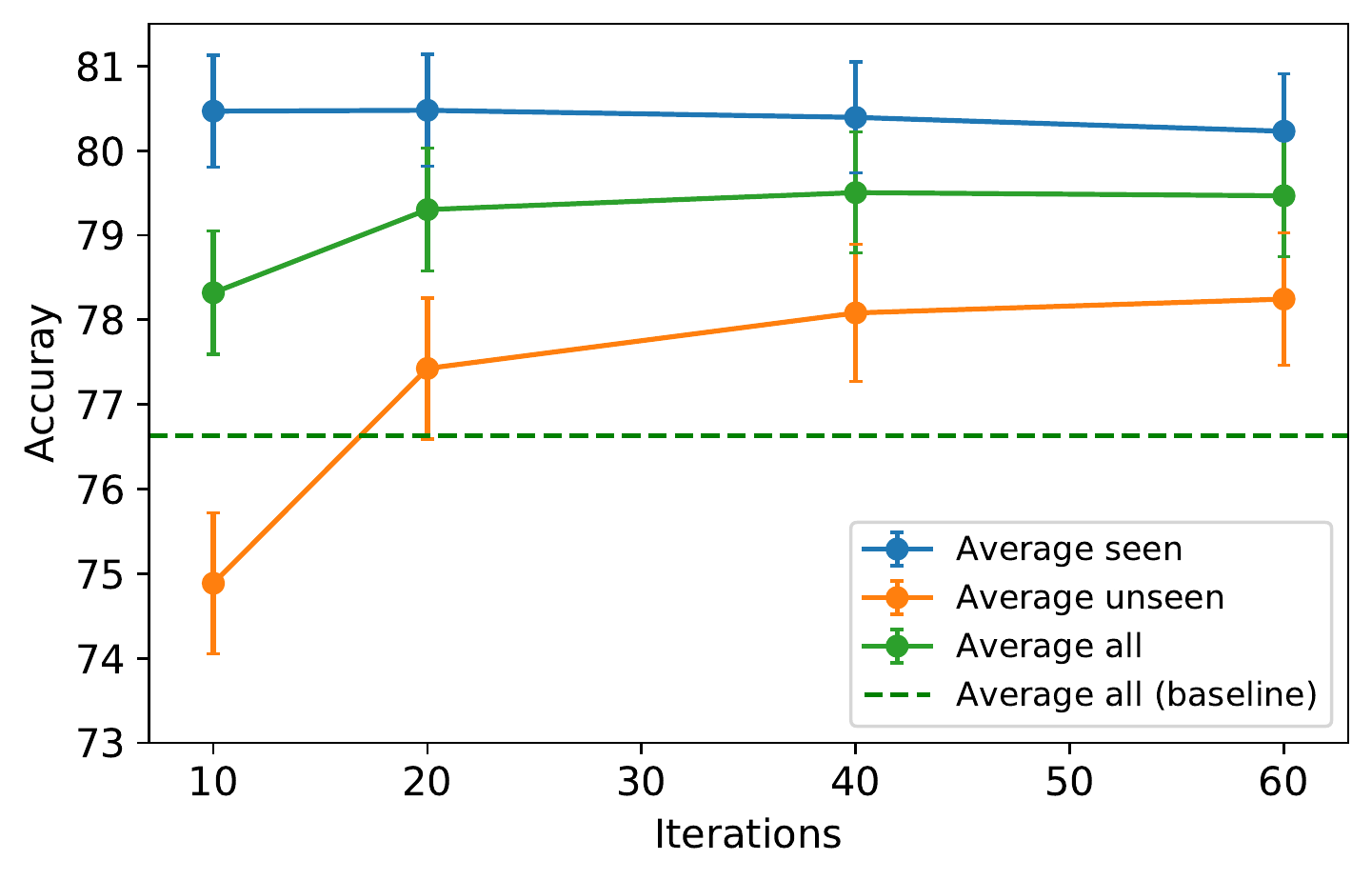}
\end{center}
\vspace{-0.3in}
\caption{Sensitivity of performance to number of iterations based on URL model.}
\label{suppfig:urlstab}
\end{figure}

Here, we conduct ablation study of our method with the URL model, unless stated otherwise.
\paragraph{Sensitivity analysis for number of iterations.}
In our method, we optimize the attached parameters ($\alpha,\beta$) with 40 iterations. \Cref{suppfig:mdlstab} and \Cref{suppfig:urlstab} report
the results with 10, 20, 40, 60 iterations and indicates that our method (solid green) converges to a stable solution after 20 iterations and achieves better average performance on all domains than the baseline URL (dash green). The mean accuracy with 95\% confidence interval are reported in \cref{supptab:mtliterations,supptab:urliterations}

\paragraph{Influence of $\alpha$ and $\beta$.} We evaluate different components of our method and report the results in \cref{supptab:com}. The results show that both residual adapters $\alpha$ and the linear transformation $\beta$ help adapt features to unseen classes while residual adapters significantly improve the performance on unseen domains. The best results are achieved by using both $\alpha$ and $\beta$.

\begin{table*}[t]
	\centering
    \resizebox{1.0\textwidth}{!}
    {
		\begin{tabular}{lcccccccc|ccccc}

		    \toprule
		    Test Dataset & ImageNet & Omniglot & Aircraft & Birds & Textures & Quick Draw & Fungi & VGG Flower & Traffic Sign & MSCOCO & MNIST & CIFAR-10 & CIFAR-100\\
		    \midrule
		    Ours w/o $\alpha$ \& $\beta$ & $57.0 \pm 1.1$ & $94.4 \pm 0.4$ & $88.0 \pm 0.5$ & $80.3 \pm 0.7$ & $74.6 \pm 0.7$ & $81.8 \pm 0.6$ & $66.2 \pm 0.9$ & $91.5 \pm 0.5$ & $49.8 \pm 1.1$ & $54.1 \pm 1.0$ & $91.1 \pm 0.4$ & $70.6 \pm 0.7$ & $59.1 \pm 1.0$ \\
		    Ours w/o $\beta$ & $57.3 \pm 1.1$ & ${\bf 94.9 \pm 0.4}$ & $88.9 \pm 0.5$ & $81.0 \pm 0.7$ & $76.7 \pm 0.7$ & $80.6 \pm 0.6$ & $65.4 \pm 0.9$ & $91.4 \pm 0.5$ & $82.6 \pm 1.0$ & $55.0 \pm 1.1$ & $96.6 \pm 0.4$ & $82.1 \pm 0.7$ & $66.4 \pm 1.1$ \\
		    Ours w/o $\alpha$ & $58.8 \pm 1.1$ & $94.5 \pm 0.4$ & $89.4 \pm 0.4$ & $80.7 \pm 0.8$ & $77.2 \pm 0.7$ & ${\bf 82.5 \pm 0.6}$ & ${\bf 68.1 \pm 0.9}$ & $92.0 \pm 0.5$ & $63.3 \pm 1.2$ & $57.3 \pm 1.0$ & $94.7 \pm 0.4$ & $74.2 \pm 0.8$ & $63.6 \pm 1.0$ \\
		    Ours & ${\bf 59.5 \pm 1.0}$ & ${\bf 94.9 \pm 0.4}$ & ${\bf 89.9 \pm 0.4}$ & ${\bf 81.1 \pm 0.8}$ & ${\bf 77.5 \pm 0.7}$ & $81.7 \pm 0.6$ & $66.3 \pm 0.9$ & ${\bf 92.2 \pm 0.5}$ & ${\bf 82.8 \pm 1.0}$ & ${\bf 57.6 \pm 1.0}$ & ${\bf 96.7 \pm 0.4}$ & ${\bf 82.9 \pm 0.7}$ & ${\bf 70.4 \pm 1.0}$ \\
			\bottomrule
		\end{tabular}%
			}
		\vspace{-0.35cm}
		\caption{Effect of each component. We build our method on the URL model and `Ours w/o $\alpha$ \& $\beta$' means we remove both residual adapters $\alpha$ and the pre-classifier adaptation layer $\beta$ in our method.}
		\label{supptab:com}
\end{table*}%

\begin{table*}[h!]
	\centering
    \resizebox{1.0\textwidth}{!}
    {
		\begin{tabular}{lcccccccc|ccccc}

		    \toprule
		    Test Dataset & ImageNet & Omniglot & Aircraft & Birds & Textures & Quick Draw & Fungi & VGG Flower & Traffic Sign & MSCOCO & MNIST & CIFAR-10 & CIFAR-100\\
		    \midrule
		    Ours(SDL-ResNet-18)-I & $59.5 \pm 1.1$& $78.2 \pm 1.2$& $72.2 \pm 1.0$& $74.9 \pm 0.9$& $77.3 \pm 0.7$& $67.6 \pm 0.9$& $44.7 \pm 1.0$& $90.9 \pm 0.6$& $82.5 \pm 0.8$& $59.0 \pm 1.0$& $93.9 \pm 0.6$& $82.1 \pm 0.7$ & $70.7 \pm 0.9$ \\
		    Ours(SDL-ResNet-18)-R & $58.2 \pm 1.0$& $78.4 \pm 1.2$& $71.1 \pm 1.1$& $74.4 \pm 1.0$& $77.1 \pm 0.7$& $67.2 \pm 1.0$& $45.9 \pm 1.0$& $90.7 \pm 0.6$& $81.9 \pm 1.0$& $57.7 \pm 1.1$& $94.1 \pm 0.5$& $81.9 \pm 0.7$& $70.5 \pm 0.9$ \\
		    \midrule
		    Ours(MDL)-I & $55.6 \pm 1.0$ & $94.3 \pm 0.4$ & $86.7 \pm 0.5$ & $79.4 \pm 0.8$ & $73.2 \pm 0.8$ & $81.7 \pm 0.6$ & $64.0 \pm 0.9$ & $90.9 \pm 0.5$ & $81.1 \pm 0.9$ & $51.4 \pm 1.1$ & $96.9 \pm 0.3$ & $78.5 \pm 0.8$ & $64.3 \pm 1.1$ \\
		    Ours(MDL)-R & $56.0 \pm 1.1$ & $94.1 \pm 0.4$ & $87.1 \pm 0.5$ & $79.7 \pm 0.8$ & $74.0 \pm 0.7$ & $82.0 \pm 0.6$ & $62.6 \pm 0.9$ & $90.6 \pm 0.6$ & $80.9 \pm 0.9$ & $51.7 \pm 1.1$ & $96.9 \pm 0.4$ & $77.7 \pm 0.9$ & $65.8 \pm 1.1$ \\
		    \midrule
		    Ours(URL)-I & $59.5 \pm 1.0$ & $94.9 \pm 0.4$ & $89.9 \pm 0.4$ & $81.1 \pm 0.8$ & $77.5 \pm 0.7$ & $81.7 \pm 0.6$ & $66.3 \pm 0.9$ & $92.2 \pm 0.5$ & $82.8 \pm 1.0$ & $57.6 \pm 1.0$ & $96.7 \pm 0.4$ & $82.9 \pm 0.7$ & $70.4 \pm 1.0$ \\
		    Ours(URL)-R & $58.8 \pm 1.1$ & $94.9 \pm 0.4$ & $90.5 \pm 0.4$ & $81.8 \pm 0.6$ & $77.7 \pm 0.7$ & $82.3 \pm 0.6$ & $66.8 \pm 0.9$ & $92.6 \pm 0.5$ & $83.7 \pm 0.8$ & $57.7 \pm 1.1$ & $96.9 \pm 0.4$ & $82.5 \pm 0.7$ & $72.0 \pm 0.9$ \\
			\bottomrule
		\end{tabular}%
			}
		\vspace{-0.25cm}
		\caption{Initialization analysis of adapters. `Ours(URL)-I' indicates our method using URL as the pretrained model and initializing residual adapters as identity matrix (scaled by $\delta = 0.0001$) while `Ours(URL)-R' means our method initialize residual adapters randomly.}
		\label{supptab:initialization}
\end{table*}%

\begin{table*}[h!]
	\centering
    \resizebox{1.0\textwidth}{!}
    {
		\begin{tabular}{lcccccccc|ccccc}

		    \toprule
		    Test Dataset & ImageNet & Omniglot & Aircraft & Birds & Textures & Quick Draw & Fungi & VGG Flower & Traffic Sign & MSCOCO & MNIST & CIFAR-10 & CIFAR-100\\
		    \midrule
		    Ours (block4) & $59.0 \pm 1.1$ & $95.0 \pm 0.4$ & $90.0 \pm 0.4$ & $80.6 \pm 0.8$ & $77.8 \pm 0.7$ & $82.3 \pm 0.6$ & $68.2 \pm 0.9$ & $91.8 \pm 0.6$ & $70.6 \pm 1.1$ & $57.1 \pm 1.1$ & $95.9 \pm 0.4$ & $77.2 \pm 0.8$ & $65.9 \pm 1.0$ \\
		    Ours (block3,4) & $60.4 \pm 1.1$ & $94.7 \pm 0.4$ & $90.0 \pm 0.5$ & $80.4 \pm 0.7$ & $77.8 \pm 0.7$ & $82.2 \pm 0.6$ & $67.2 \pm 0.8$ & $92.5 \pm 0.5$ & $77.2 \pm 1.0$ & $57.9 \pm 1.0$ & $96.7 \pm 0.3$ & $78.8 \pm 0.9$ & $68.6 \pm 0.9$ \\
		    Ours (block2,3,4) & $59.6 \pm 1.1$ & $94.9 \pm 0.4$ & $89.9 \pm 0.5$ & $81.0 \pm 0.8$ & $78.2 \pm 0.7$ & $82.4 \pm 0.6$ & $67.6 \pm 0.9$ & $92.3 \pm 0.5$ & $81.5 \pm 1.0$ & $57.9 \pm 1.0$ & $96.6 \pm 0.4$ & $81.5 \pm 0.8$ & $70.6 \pm 1.0$ \\
		    Ours (block-all) & $59.5 \pm 1.0$ & $94.9 \pm 0.4$ & $89.9 \pm 0.4$ & $81.1 \pm 0.8$ & $77.5 \pm 0.7$ & $81.7 \pm 0.6$ & $66.3 \pm 0.9$ & $92.2 \pm 0.5$ & $82.8 \pm 1.0$ & $57.6 \pm 1.0$ & $96.7 \pm 0.4$ & $82.9 \pm 0.7$ & $70.4 \pm 1.0$ \\
			\bottomrule
		\end{tabular}%
			}
		\vspace{-0.25cm}
		\caption{Block (layer) analysis for adapters based on URL model.}
		\label{supptab:layer}
\end{table*}%

\begin{figure}[h!]

\begin{center}

\includegraphics[width=0.9\linewidth]{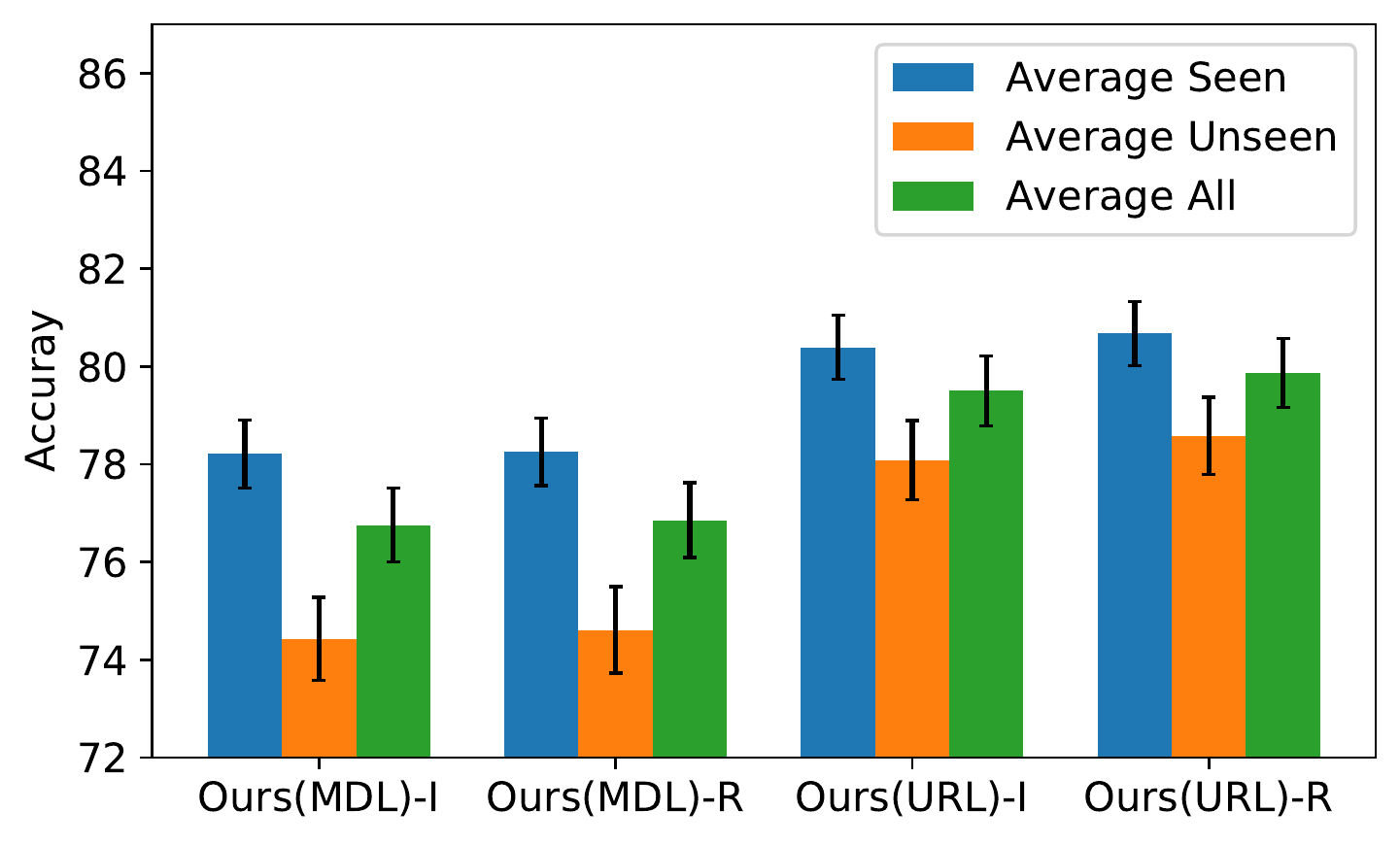}
\end{center}
\vspace{-0.3in}
\caption{Initialization analysis for adapters. '-I' indicates identity initialization and `-R' is randomly initialization.}
\label{suppfig:init}
\end{figure}

\paragraph{Initialization analysis for adapters.}
Here, we investigate using different initialization strategies for adapters: i) Identity initialization: in this work we initialize each residual adapter as an identity matrix scaled by a scalar $\delta$ and we set $\delta=1e-4$; ii) randomly initialization: alternatively, we can randomly initialize each residual adapter. The results of different initialization are summarized in \cref{suppfig:init}. We can see that our methods with different initialization strategies obtain similar results, which indicates that our method works also with randomly initialization and again verifies the stability of our method. Detailed results of each datasets are shown in \cref{supptab:initialization}.

\begin{figure}[h]
\begin{center}
\includegraphics[width=0.9\linewidth]{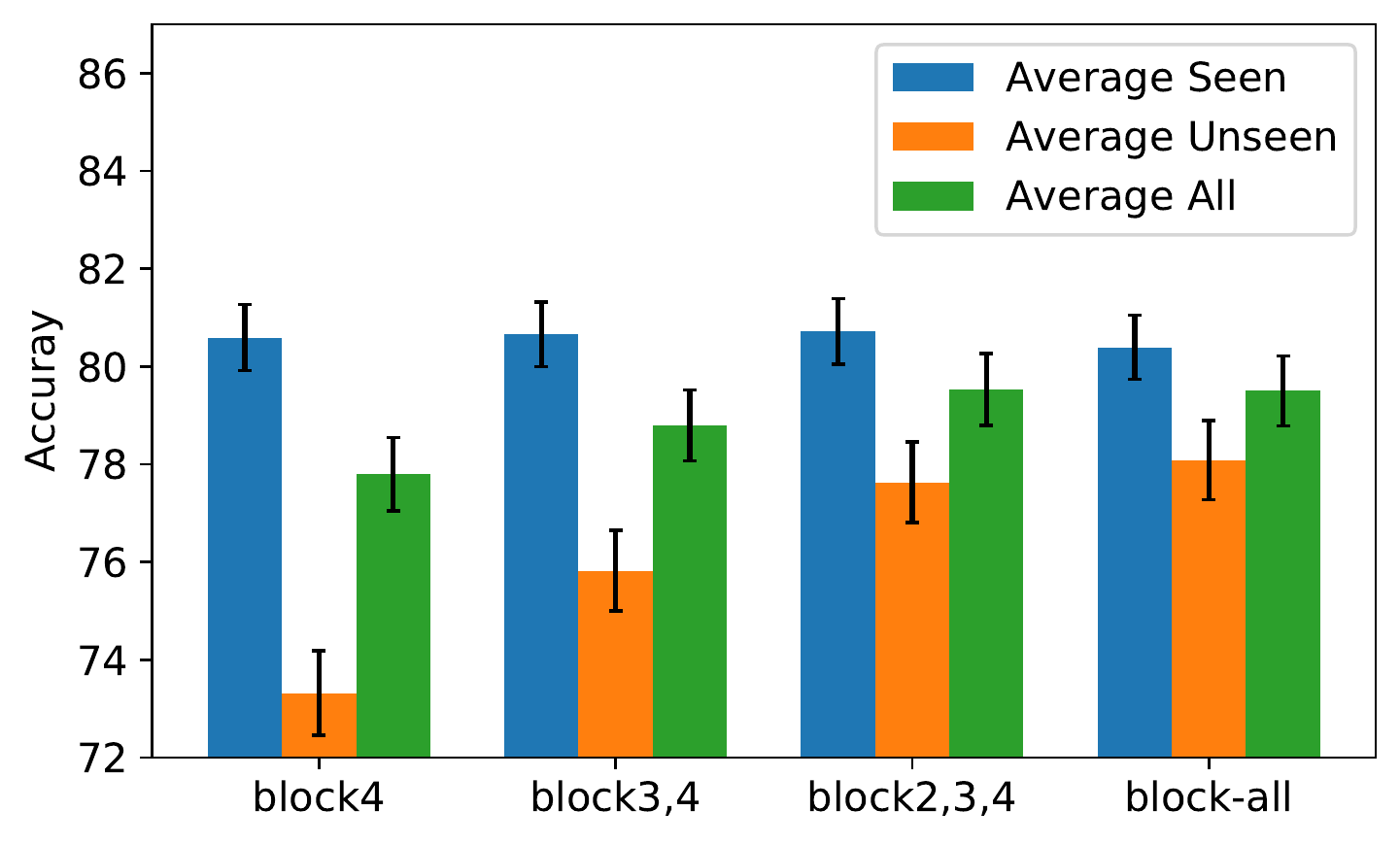}
\end{center}
\vspace{-0.3in}
\caption{Block (layer) analysis for adapters.}
\label{suppfig:urllayers}

\end{figure}

\begin{table*}[t]
	\centering
    \resizebox{1.0\textwidth}{!}
    {
		\begin{tabular}{lcccccccc|ccccc}

		    \toprule
		    Test Dataset & ImageNet & Omniglot & Aircraft & Birds & Textures & Quick Draw & Fungi & VGG Flower & Traffic Sign & MSCOCO & MNIST & CIFAR-10 & CIFAR-100\\
		    \midrule
		    Ours & $59.5 \pm 1.0$ & $94.9 \pm 0.4$ & $89.9 \pm 0.4$ & $81.1 \pm 0.8$ & $77.5 \pm 0.7$ & $81.7 \pm 0.6$ & $66.3 \pm 0.9$ & $92.2 \pm 0.5$ & $82.8 \pm 1.0$ & $57.6 \pm 1.0$ & $96.7 \pm 0.4$ & $82.9 \pm 0.7$ & $70.4 \pm 1.0$ \\
		    Ours(N=2) & $58.9 \pm 1.1$ & $95.2 \pm 0.4$ & $89.7 \pm 0.5$ & $80.9 \pm 0.7$ & $76.7 \pm 0.7$ & $81.4 \pm 0.6$ & $67.7 \pm 0.9$ & $92.2 \pm 0.5$ & $82.4 \pm 1.0$ & $57.1 \pm 1.0$ & $96.5 \pm 0.4$ & $82.4 \pm 0.7$ & $70.3 \pm 1.0$ \\
		    Ours(N=4) & $58.7 \pm 1.1$ & $94.9 \pm 0.4$ & $89.7 \pm 0.5$ & $80.3 \pm 0.7$ & $77.0 \pm 0.7$ & $82.5 \pm 0.6$ & $67.2 \pm 0.9$ & $92.5 \pm 0.5$ & $82.6 \pm 1.0$ & $57.5 \pm 1.1$ & $96.5 \pm 0.4$ & $82.5 \pm 0.7$ & $70.8 \pm 0.9$ \\
		    Ours(N=8) & $59.1 \pm 1.1$ & $95.0 \pm 0.4$ & $89.8 \pm 0.5$ & $80.2 \pm 0.8$ & $77.2 \pm 0.7$ & $82.1 \pm 0.6$ & $67.0 \pm 0.9$ & $92.2 \pm 0.5$ & $82.5 \pm 1.0$ & $57.2 \pm 1.1$ & $96.8 \pm 0.4$ & $82.6 \pm 0.7$ & $71.8 \pm 0.9$ \\
		    Ours(N=16) & $58.2 \pm 1.1$ & $94.7 \pm 0.4$ & $90.1 \pm 0.4$ & $80.3 \pm 0.8$ & $76.9 \pm 0.7$ & $81.7 \pm 0.6$ & $67.6 \pm 0.9$ & $92.0 \pm 0.5$ & $81.8 \pm 1.0$ & $58.1 \pm 1.1$ & $96.4 \pm 0.4$ & $81.8 \pm 0.7$ & $71.1 \pm 0.9$ \\
		    Ours(N=32) & $59.2 \pm 1.1$ & $94.8 \pm 0.4$ & $89.6 \pm 0.5$ & $80.0 \pm 0.8$ & $77.3 \pm 0.6$ & $82.4 \pm 0.6$ & $67.2 \pm 0.9$ & $92.1 \pm 0.5$ & $82.1 \pm 1.0$ & $57.1 \pm 1.0$ & $96.7 \pm 0.3$ & $81.6 \pm 0.8$ & $71.1 \pm 0.9$ \\
			\bottomrule
		\end{tabular}%
			}
		\vspace{-0.15cm}
		\caption{Results of using decomposed RA on layer3,4.}
		\label{supptab:decoml34}
\end{table*}%

\begin{table*}[h!]
	\centering
    \resizebox{1.0\textwidth}{!}
    {
		\begin{tabular}{lcccccccc|ccccc}

		    \toprule
		    Test Dataset & ImageNet & Omniglot & Aircraft & Birds & Textures & Quick Draw & Fungi & VGG Flower & Traffic Sign & MSCOCO & MNIST & CIFAR-10 & CIFAR-100\\
		    \midrule
		    Ours & $59.5 \pm 1.0$ & $94.9 \pm 0.4$ & $89.9 \pm 0.4$ & $81.1 \pm 0.8$ & $77.5 \pm 0.7$ & $81.7 \pm 0.6$ & $66.3 \pm 0.9$ & $92.2 \pm 0.5$ & $82.8 \pm 1.0$ & $57.6 \pm 1.0$ & $96.7 \pm 0.4$ & $82.9 \pm 0.7$ & $70.4 \pm 1.0$ \\
		    Ours(N=2) & $58.1 \pm 1.1$ & $94.8 \pm 0.4$ & $89.7 \pm 0.5$ & $80.2 \pm 0.8$ & $76.9 \pm 0.7$ & $82.1 \pm 0.6$ & $67.8 \pm 0.9$ & $92.0 \pm 0.6$ & $82.5 \pm 0.9$ & $56.9 \pm 1.1$ & $96.7 \pm 0.3$ & $82.0 \pm 0.8$ & $70.3 \pm 1.0$ \\
		    Ours(N=4) & $59.6 \pm 1.1$ & $94.8 \pm 0.4$ & $89.9 \pm 0.5$ & $80.3 \pm 0.8$ & $77.4 \pm 0.7$ & $82.6 \pm 0.6$ & $66.6 \pm 0.9$ & $92.9 \pm 0.5$ & $79.7 \pm 1.1$ & $57.6 \pm 1.1$ & $96.5 \pm 0.4$ & $80.9 \pm 0.8$ & $70.6 \pm 1.0$ \\
		    Ours(N=8) & $58.2 \pm 1.1$ & $94.6 \pm 0.4$ & $89.6 \pm 0.5$ & $81.2 \pm 0.8$ & $76.6 \pm 0.7$ & $82.7 \pm 0.6$ & $66.5 \pm 0.9$ & $92.3 \pm 0.5$ & $78.1 \pm 1.1$ & $57.3 \pm 1.0$ & $96.3 \pm 0.3$ & $81.0 \pm 0.8$ & $70.9 \pm 0.9$ \\
		    Ours(N=16) & $58.9 \pm 1.1$ & $94.6 \pm 0.4$ & $89.7 \pm 0.5$ & $80.1 \pm 0.7$ & $77.0 \pm 0.7$ & $82.1 \pm 0.6$ & $68.4 \pm 0.9$ & $91.9 \pm 0.5$ & $78.3 \pm 1.0$ & $57.8 \pm 1.1$ & $96.0 \pm 0.4$ & $82.0 \pm 0.7$ & $70.3 \pm 1.0$ \\
			\bottomrule
		\end{tabular}%
			}
		\vspace{-0.15cm}
		\caption{Results of using decomposed RA on all layers.}
		\label{supptab:decom}
\end{table*}%

\paragraph{Layer analysis for adapters.}
Here we investigate whether it is sufficient to attach the adapters only to the later layers.
We evaluate this on ResNet18 which is composed of four blocks and attach the adapters to only later blocks (block4, block3,4, block2,3,4 and block-all. 
\Cref{suppfig:urllayers} shows that applying our adapters to only the last block (block4) obtains around 78\% average accuracy on all domains which outperforms the URL. With attaching residual adapters to more layers, the performance on unseen domains is improved significantly while the one on seen domains remains stable.
The mean accuracy with 95\% confidence interval for layer analysis are shown in \cref{supptab:layer}.

\begin{figure}[h]

\begin{center}

\includegraphics[width=0.9\linewidth]{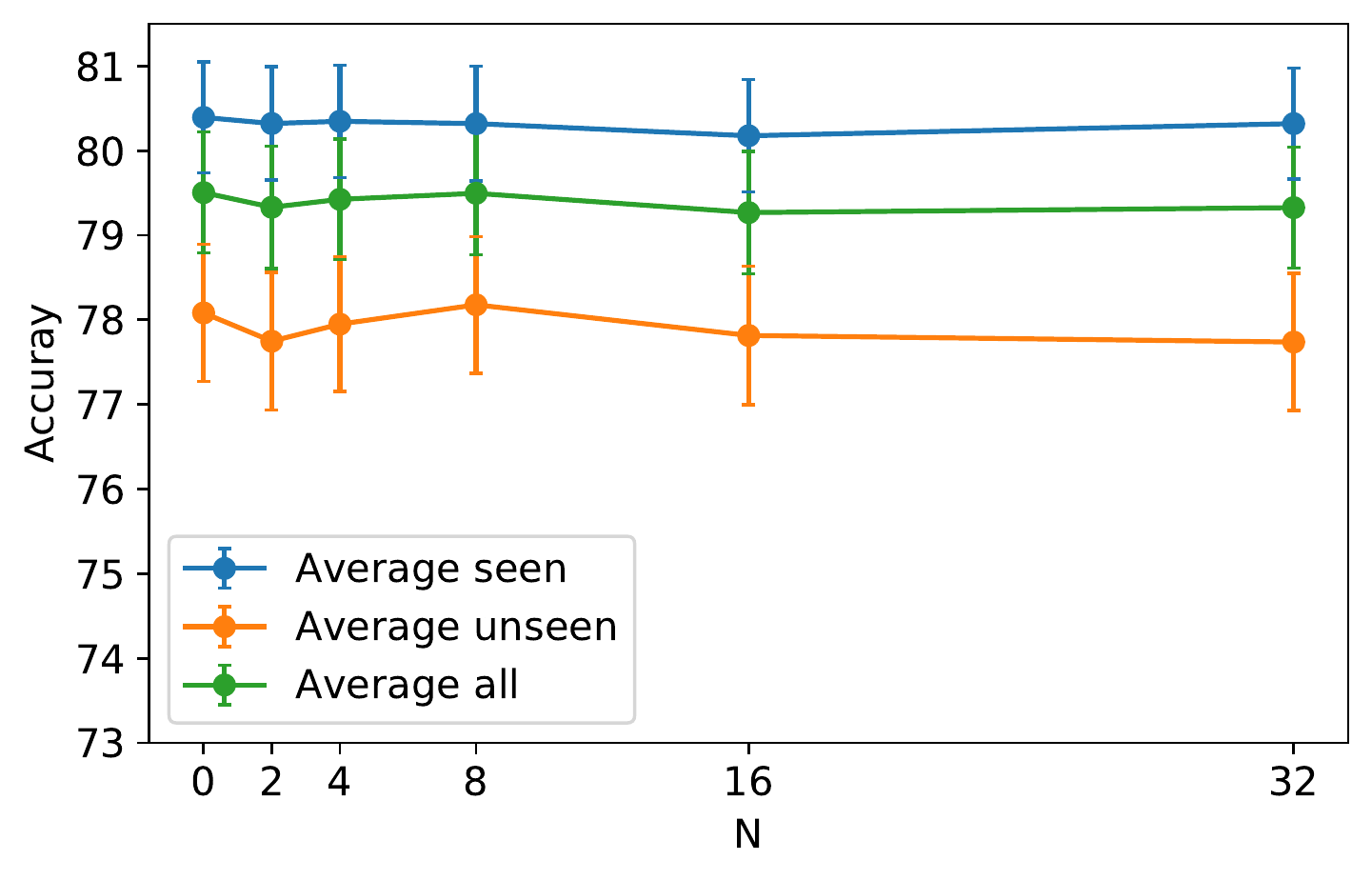}
\end{center}
\vspace{-0.3in}
\caption{Decomposed residual adapters on block-3,4.}
\label{suppfig:urldecoml34}
\end{figure}

\paragraph{Decomposing residual adapters.}
Here we investigate whether one can reduce the number of parameters in the adapters while retaining its performance by using matrix decomposition.
As in deep neural network, the adapters in earlier layers are relatively small, we then decompose the adapters in the last two blocks only where the adapter dimensionality goes up to $512\times 512$. \Cref{suppfig:urldecoml34} shows that our method can achieve good performance with less parameters by decomposing large residual adapters, (\eg when $N=32$ where the number of additional parameters equal to around 4\% vs 13\%, the performance is still comparable to the original form of residual adapters, \ie N=0). Results of each datasets in \cref{supptab:decoml34}, also show that, by decomposing large residual adapters, the performance of our method is still comparable to the original form of residual adapters (\ie Ours) with less parameters.

\begin{figure}[h]
\begin{center}
\includegraphics[width=0.9\linewidth]{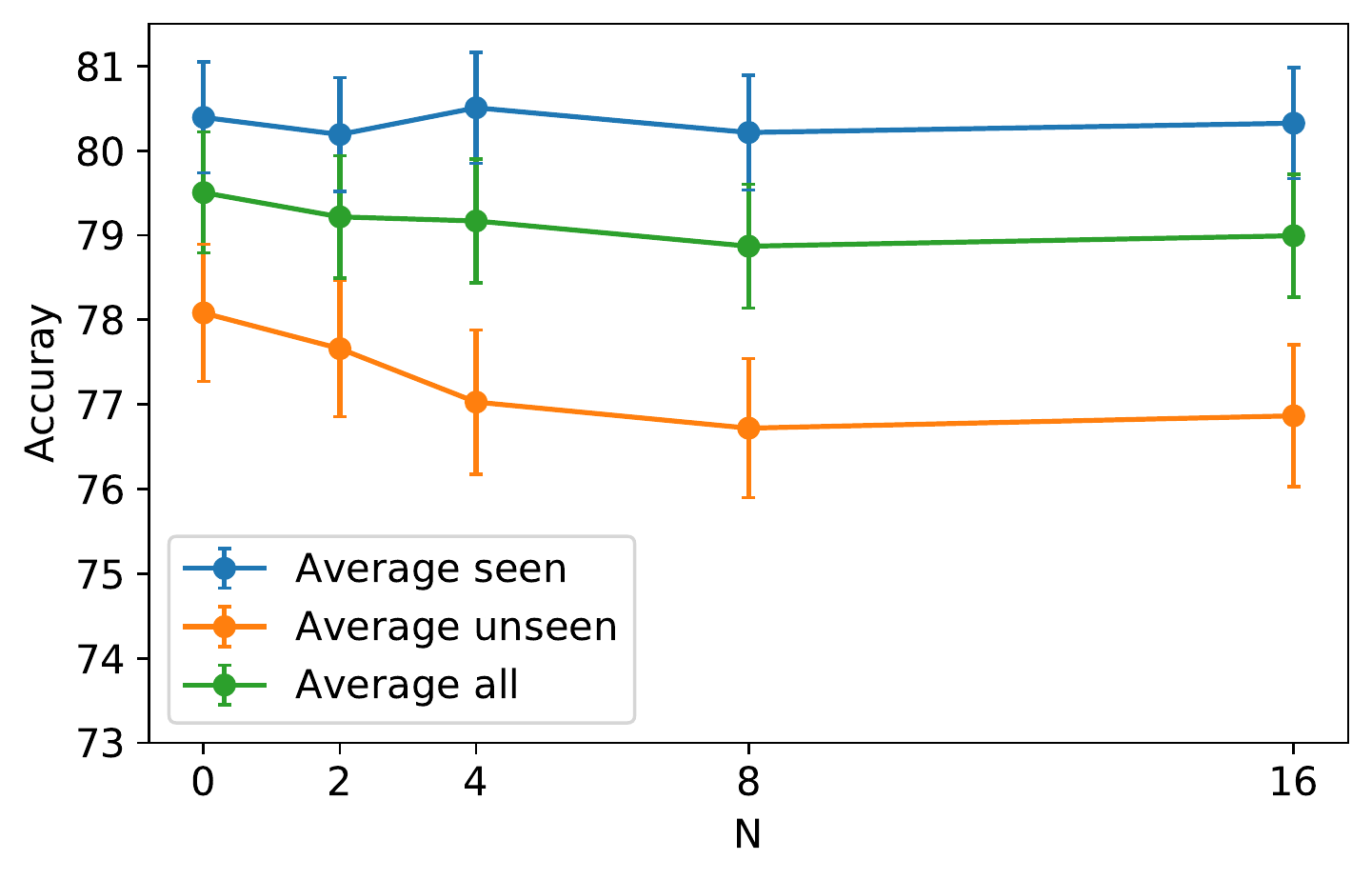}
\end{center}
\vspace{-0.3in}
\caption{Decomposed residual adapters on all layers.}
\label{suppfig:urldecom}
\end{figure}

The similar conclusion can be drawn from results (shown in \cref{suppfig:urldecom}) of our method using decomposed residual adapters in all layers. When N increases, \ie, smaller residual adapters, the average accuracy on all domains is still comparable to the original form of residual adapters (\ie N=0) with less parameters though the average accuracy on unseen domains drops slightly. From the results depicted in \cref{supptab:decom}, we can see that when $N$ increases, the performance of most domains are still comparable to the original form of residual adapters (\ie Ours) while the performance on Traffic Sign drops slightly as the adapters in earlier layers are small and when N is larger the decomposed residual adapters might be too small to tranform the features. In overall, our method can achieve good performance with less parameters by decomposing large residual adapters.

\paragraph{Training time.}
The training time (meta-train) of our method is equal to the one of URL (hence no additional cost), \ie 48 hours in multi-domain setting, 6 hours for Resnet-18 and 33 hours for Resnet-34 in single-domain learning in one Nvidia V100 GPU. 
Whereas CTX meta-training requires 8 Nvidia V100 GPUs for 7 days and approximately 40 times more expensive than ours. 
During the meta-test stage, the model parameters are further trained using support set of each episode. 
Meta-test training cost is depicted in \cref{supptab:testtime} for Meta-Dataset tasks. 
URL baseline only finetunes parameters of PA $\beta$.
Finetune+NCC updates the entire backbone parameters.
Ours learn RA and PA parameters. 
While URL is the fastest baseline, as it does not require backpropagating the error to early layers, ours is more efficient than finetuning all the backbone parameters.

\setcounter{table}{11}
\begin{table}[ht!]
	\centering
	
    \resizebox{1.0\textwidth}{!}
    {
		\begin{tabular}{lccccccccccccccc}

		    \toprule
		    \multirow{2}{*}{Test Dataset} & Image & Omni & Air- & \multirow{2}{*}{Birds} & Tex- & Quick & \multirow{2}{*}{Fungi} & VGG & Traffic & MS- & \multirow{2}{*}{MNIST} & CIFAR & CIFAR\\
		    & -Net & -glot & craft & & tures & Draw & & Flower & Sign & COCO & & -10 & -100 \\
		    \midrule
		    URL & $0.7$ & $0.7$ & $0.4$ & $0.7$ & $0.4$ & $1.0$ & $1.0$ & $0.5$ & $0.9$ & $0.9$ & $0.4$ & $0.4$ & $1.0$ \\
		    Finetune+NCC & $7.7$ & $2.5$ & $7.4$ & $7.0$ & $5.8$ & $9.3$ & $8.7$ & $6.6$ & $9.1$ & $9.0$ & $6.5$ & $6.7$ & $9.3$ \\
		    Ours (URL+RA+PA) & $7.2$ & $2.4$ & $6.1$ & $6.8$ & $4.8$ & $8.9$ & $7.4$ & $5.2$ & $8.8$ & $8.3$ & $6.0$ & $6.2$ & $8.6$ \\
			\bottomrule
		\end{tabular}%
			}
		\vspace{-0.35cm}
		\caption{\footnotesize Computation cost (\# second per task) during meta-test.}
		\label{supptab:testtime}
\end{table}%

\subsection{Qualitative results}
\begin{figure}[h!]
\begin{center}
\includegraphics[width=0.9\linewidth]{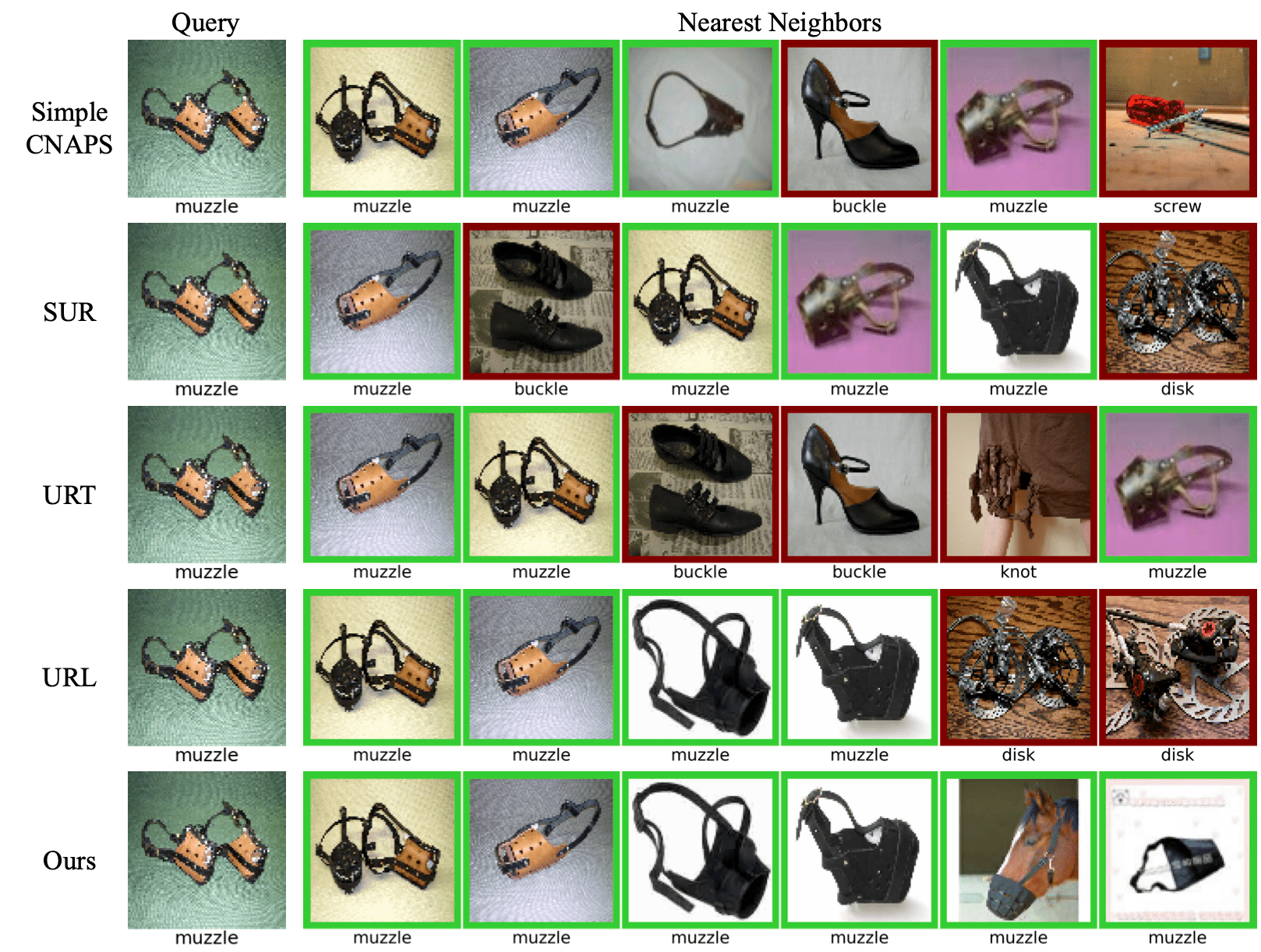}
\end{center}
\vspace{-0.3in}
\caption{Qualitative comparison to Simple CNAPS~\cite{bateni2020improved}, SUR~\cite{dvornik2020selecting}, URT~\cite{liu2020universal}, and URL~\cite{li2021universal} in ImageNet. Green and red colors indicate correct and false predictions respectively.}
\label{suppfig:imagenet}
\end{figure}

\begin{figure}[h!]
\begin{center}
\includegraphics[width=0.9\linewidth]{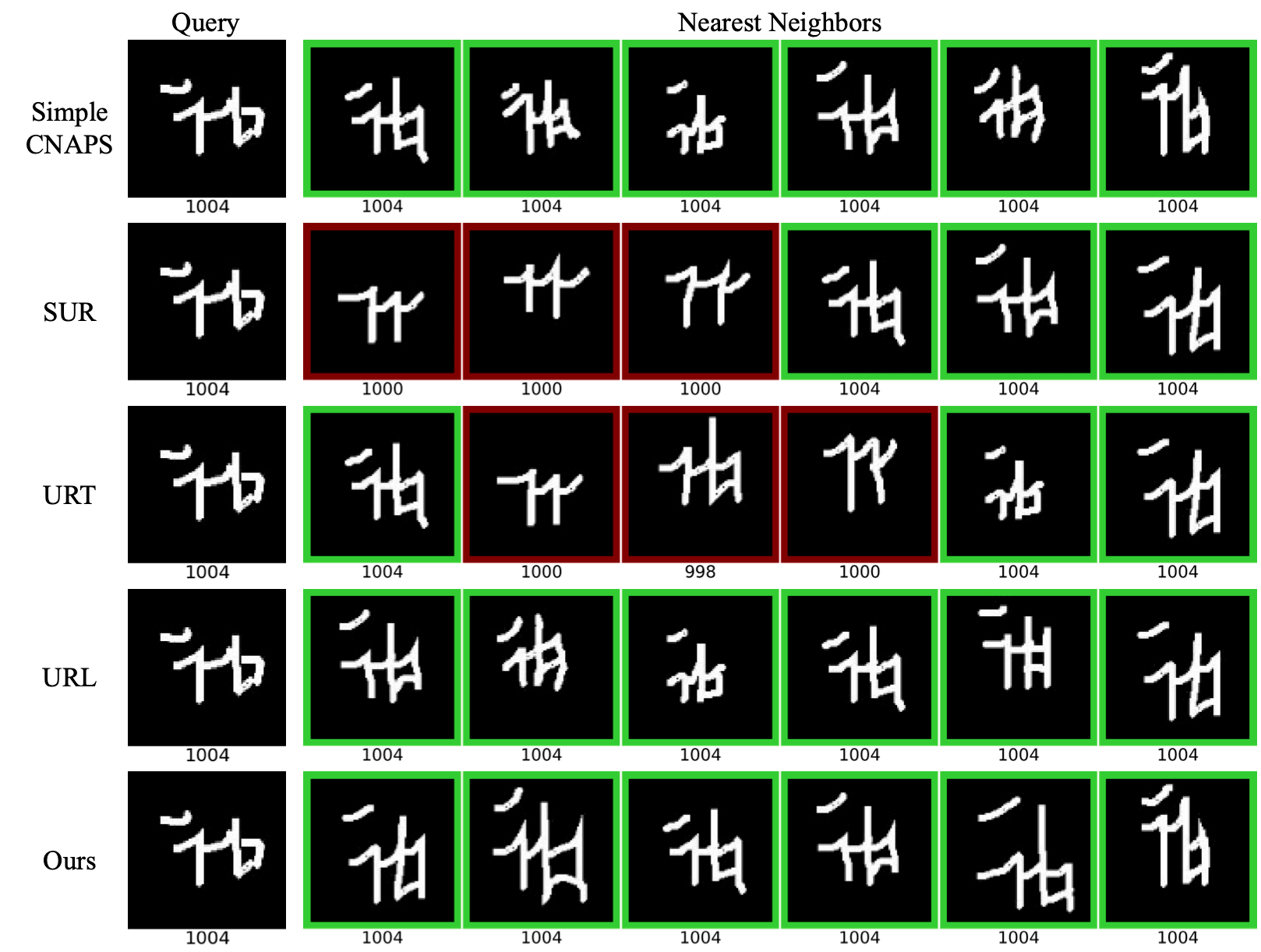}
\end{center}
\vspace{-0.3in}
\caption{Qualitative comparison to Simple CNAPS~\cite{bateni2020improved}, SUR~\cite{dvornik2020selecting}, URT~\cite{liu2020universal}, and URL~\cite{li2021universal} in Omniglot. Green and red colors indicate correct and false predictions respectively.}
\label{suppfig:omniglot}
\end{figure}

\begin{figure}[h!]
\begin{center}
\includegraphics[width=0.9\linewidth]{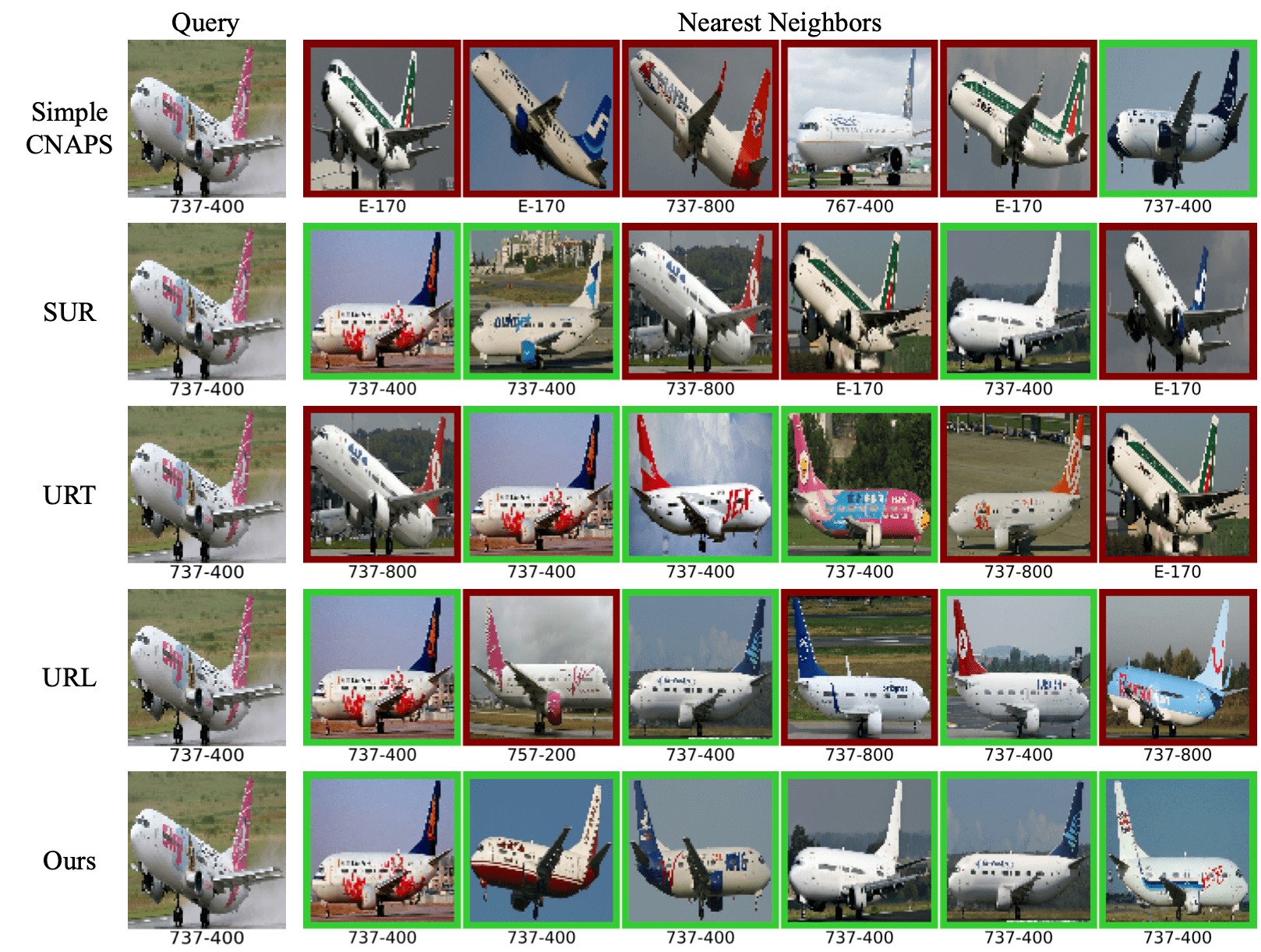}
\end{center}
\vspace{-0.3in}
\caption{Qualitative comparison to Simple CNAPS~\cite{bateni2020improved}, SUR~\cite{dvornik2020selecting}, URT~\cite{liu2020universal}, and URL~\cite{li2021universal} in Aircraft. Green and red colors indicate correct and false predictions respectively.}
\label{suppfig:aircraft}
\end{figure}

\begin{figure}[h!]
\begin{center}
\includegraphics[width=0.9\linewidth]{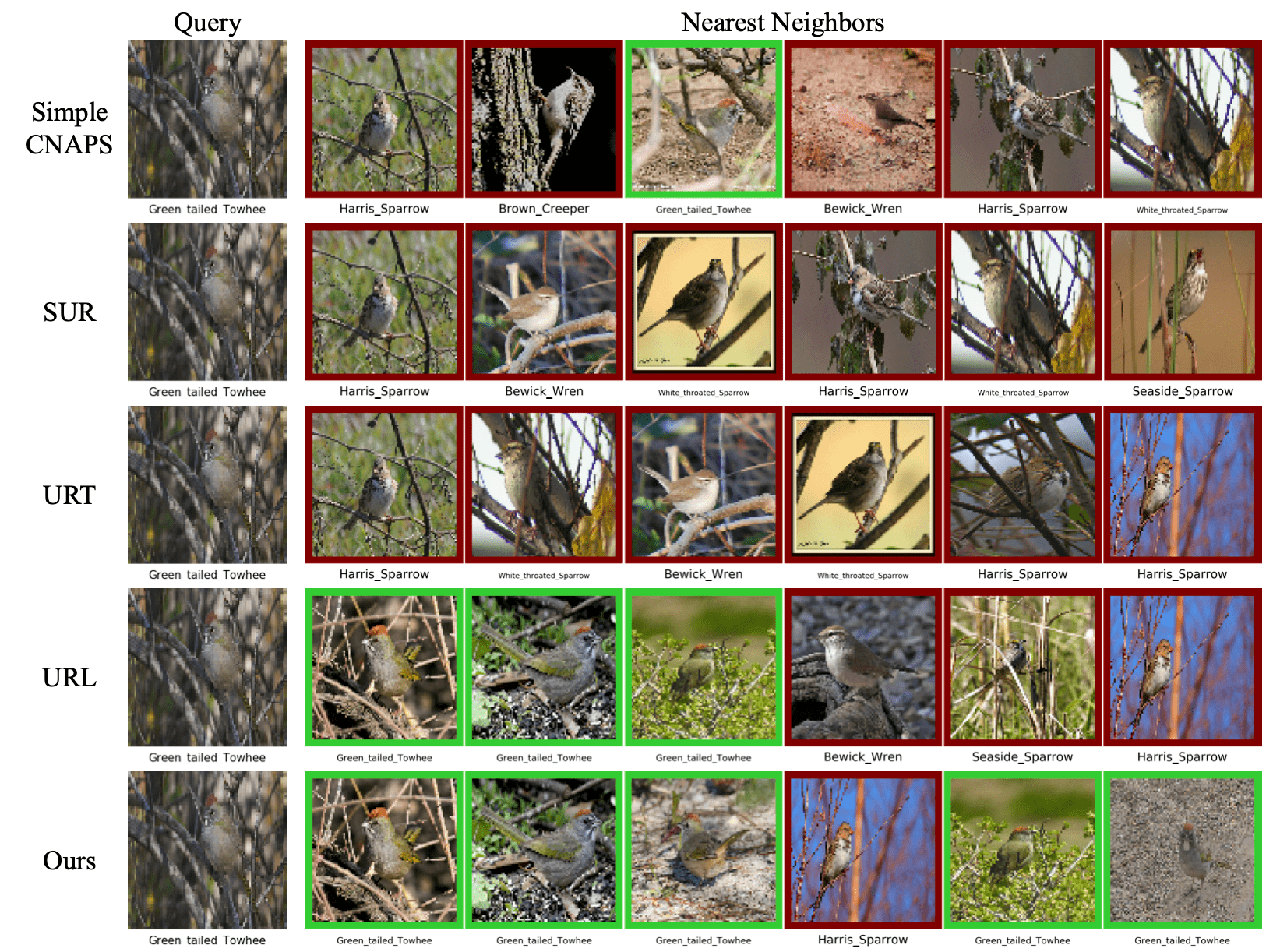}
\end{center}
\vspace{-0.3in}
\caption{Qualitative comparison to Simple CNAPS~\cite{bateni2020improved}, SUR~\cite{dvornik2020selecting}, URT~\cite{liu2020universal}, and URL~\cite{li2021universal} in Birds. Green and red colors indicate correct and false predictions respectively.}
\label{suppfig:birds}
\end{figure}

\begin{figure}[h!]
\begin{center}
\includegraphics[width=0.9\linewidth]{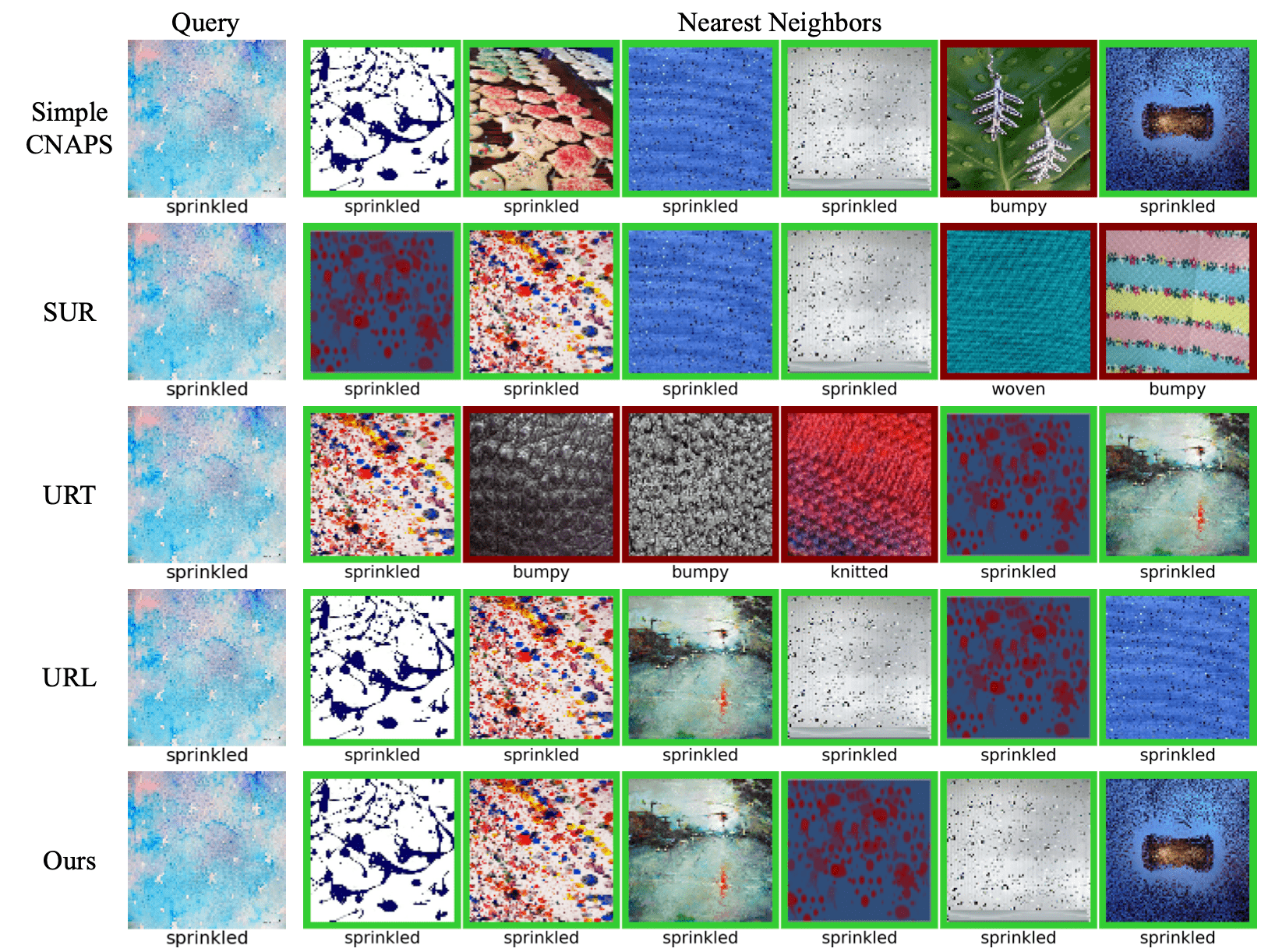}
\end{center}
\vspace{-0.3in}
\caption{Qualitative comparison to Simple CNAPS~\cite{bateni2020improved}, SUR~\cite{dvornik2020selecting}, URT~\cite{liu2020universal}, and URL~\cite{li2021universal} in Textures. Green and red colors indicate correct and false predictions respectively.}
\label{suppfig:texture}
\end{figure}

\begin{figure}[h!]
\begin{center}
\includegraphics[width=0.9\linewidth]{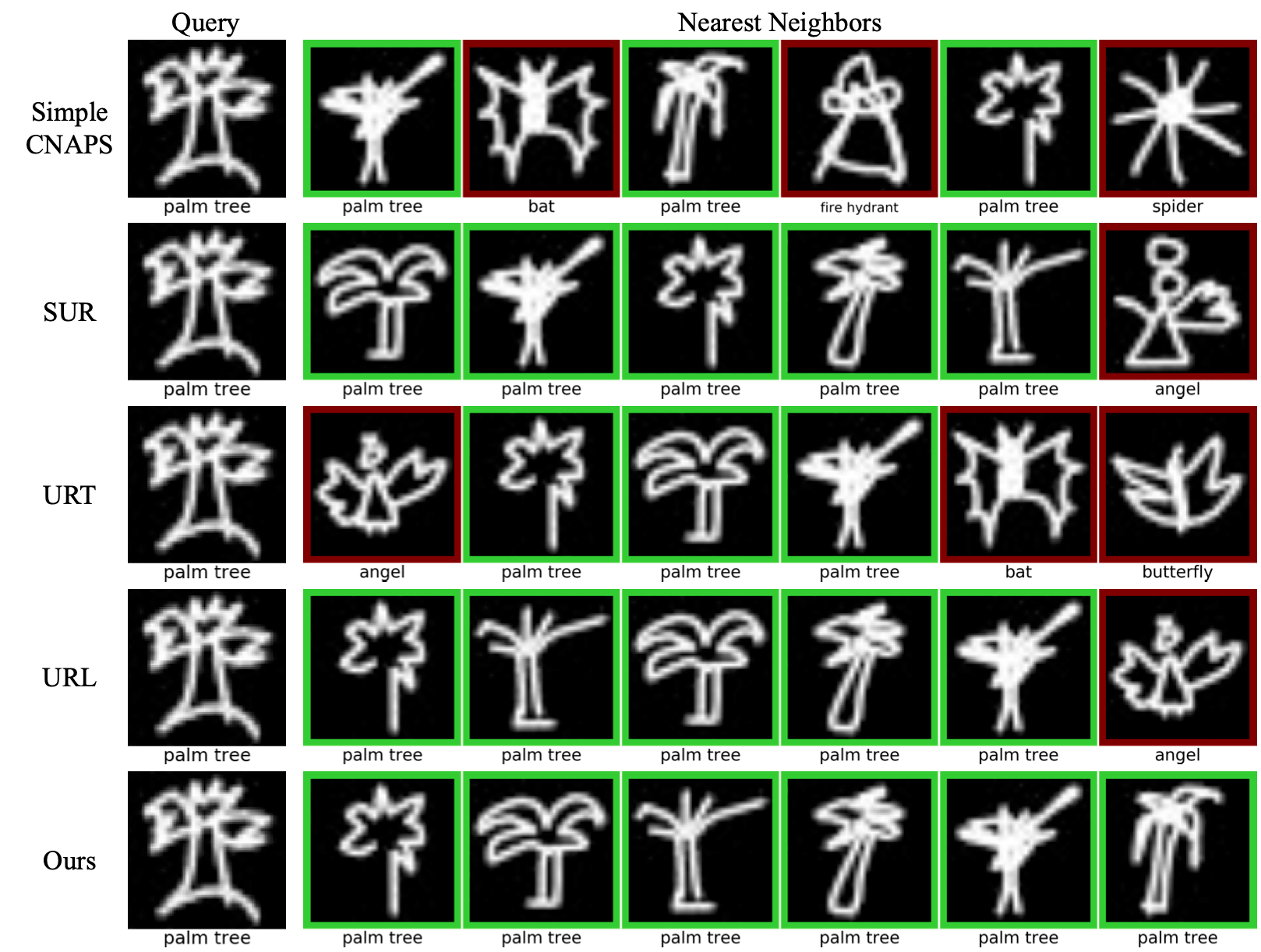}
\end{center}
\vspace{-0.3in}
\caption{Qualitative comparison to Simple CNAPS~\cite{bateni2020improved}, SUR~\cite{dvornik2020selecting}, URT~\cite{liu2020universal}, and URL~\cite{li2021universal} in Quick Draw. Green and red colors indicate correct and false predictions respectively.}
\label{suppfig:quickdraw}
\end{figure}

\begin{figure}[h!]
\begin{center}
\includegraphics[width=0.9\linewidth]{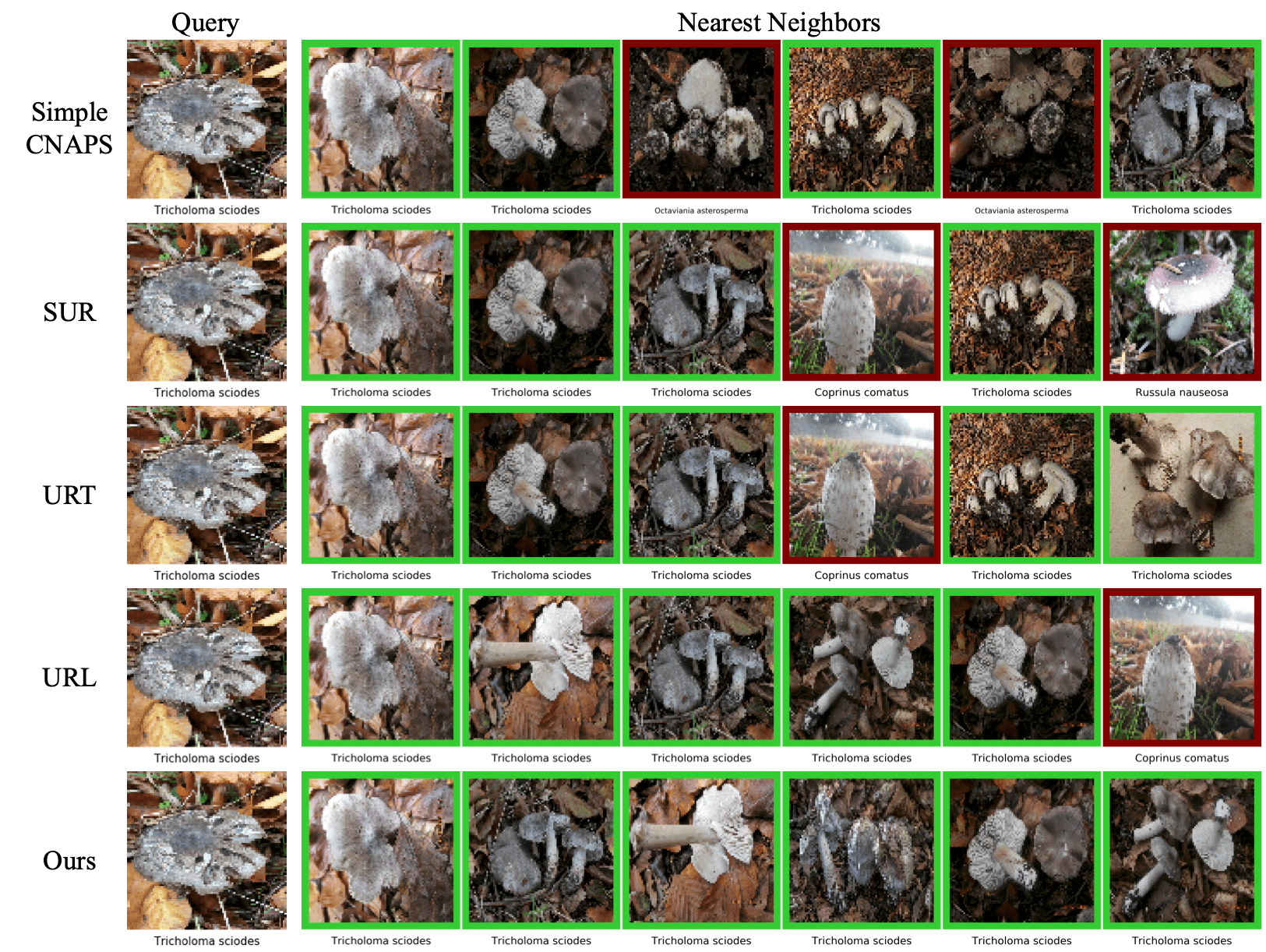}
\end{center}
\vspace{-0.3in}
\caption{Qualitative comparison to Simple CNAPS~\cite{bateni2020improved}, SUR~\cite{dvornik2020selecting}, URT~\cite{liu2020universal}, and URL~\cite{li2021universal} in Fungi. Green and red colors indicate correct and false predictions respectively.}
\label{suppfig:fungi}
\end{figure}

\begin{figure}[h!]
\begin{center}
\includegraphics[width=0.9\linewidth]{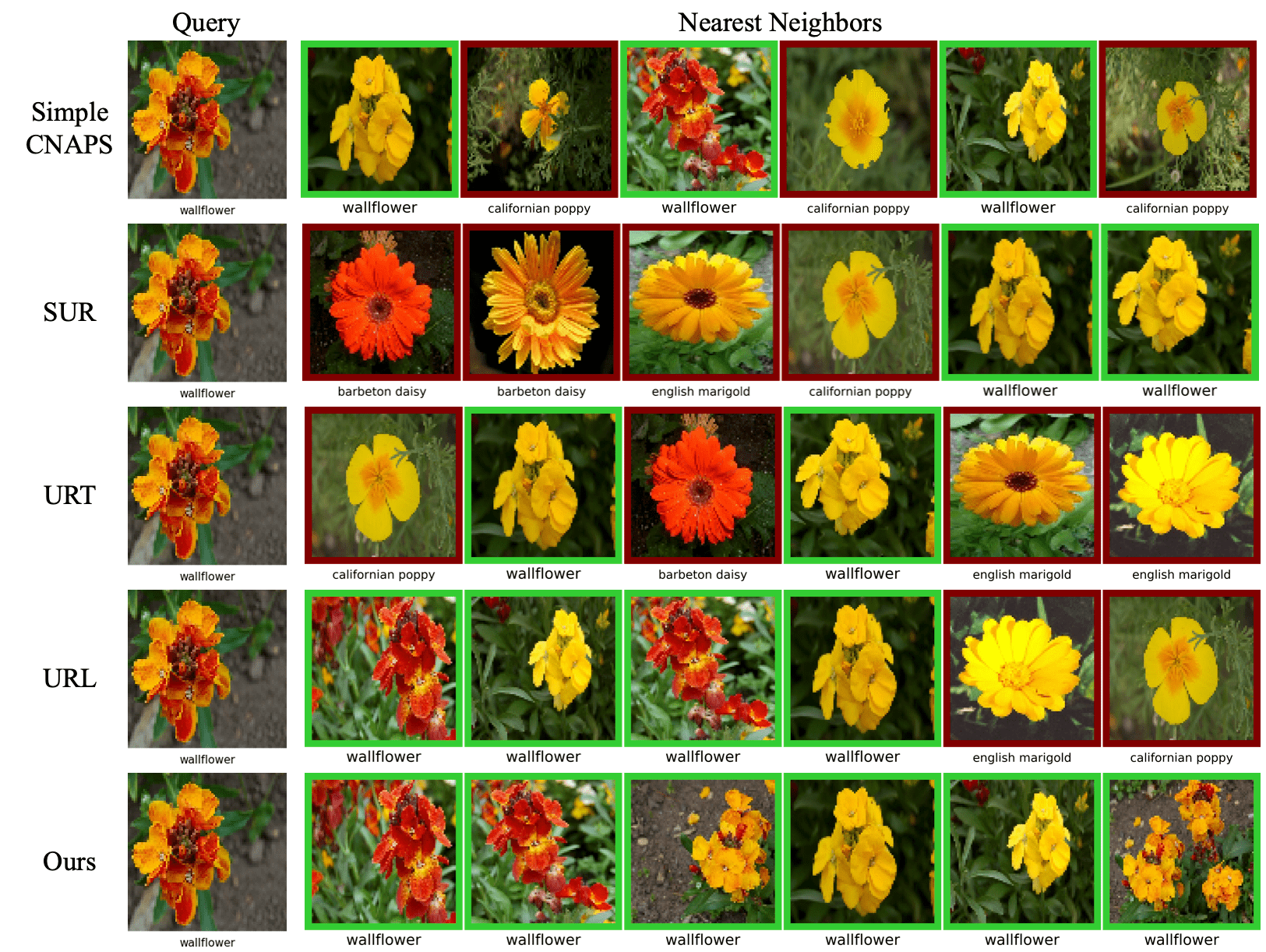}
\end{center}
\vspace{-0.3in}
\caption{Qualitative comparison to Simple CNAPS~\cite{bateni2020improved}, SUR~\cite{dvornik2020selecting}, URT~\cite{liu2020universal}, and URL~\cite{li2021universal} in VGG Flower. Green and red colors indicate correct and false predictions respectively.}
\label{suppfig:flower}
\end{figure}

\begin{figure}[h!]
\begin{center}
\includegraphics[width=0.9\linewidth]{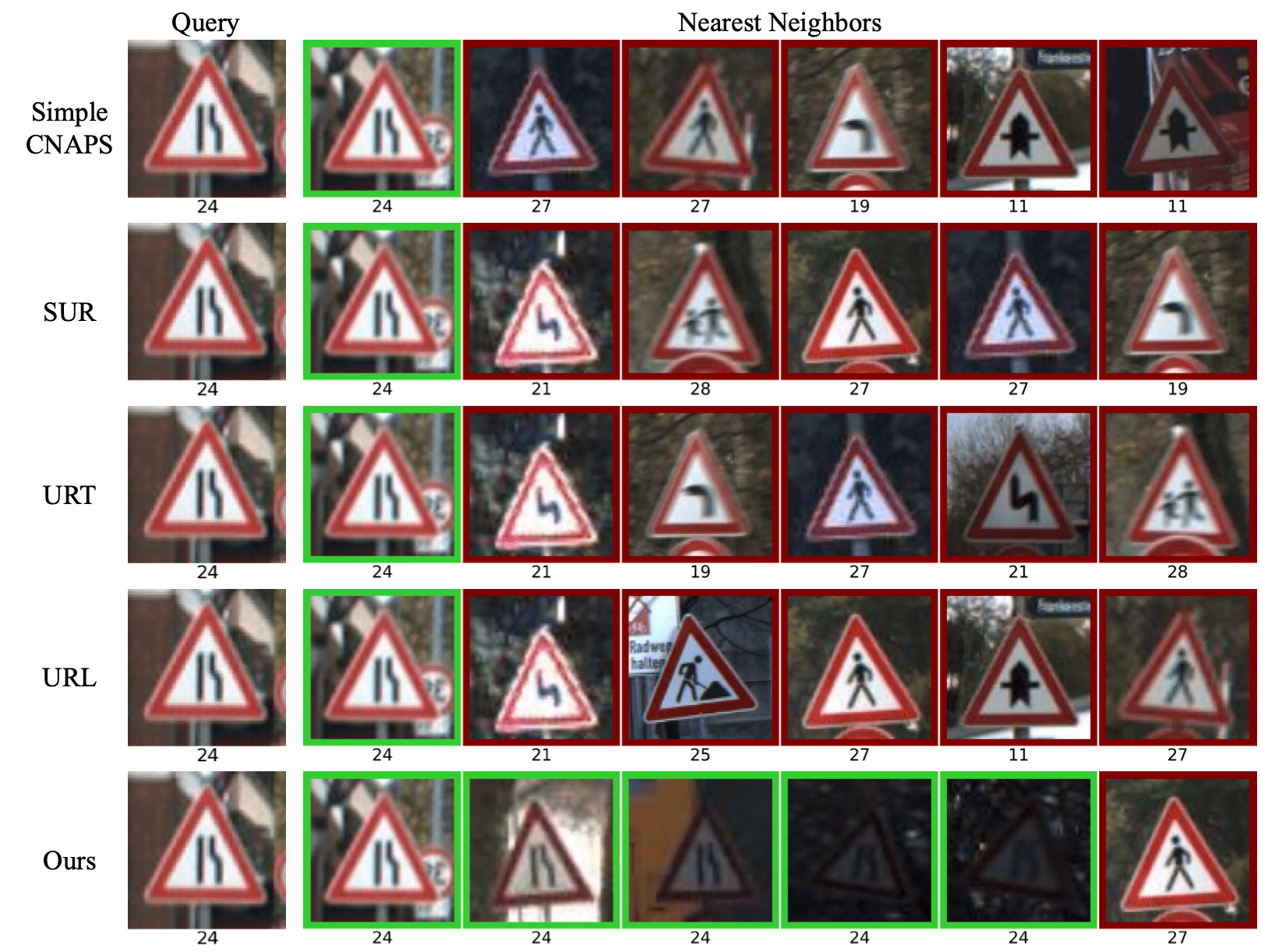}
\end{center}
\vspace{-0.3in}
\caption{Qualitative comparison to Simple CNAPS~\cite{bateni2020improved}, SUR~\cite{dvornik2020selecting}, URT~\cite{liu2020universal}, and URL~\cite{li2021universal} in Traffic Sign. Green and red colors indicate correct and false predictions respectively.}
\label{suppfig:traffic}
\end{figure}

\begin{figure}[h!]
\begin{center}
\includegraphics[width=0.9\linewidth]{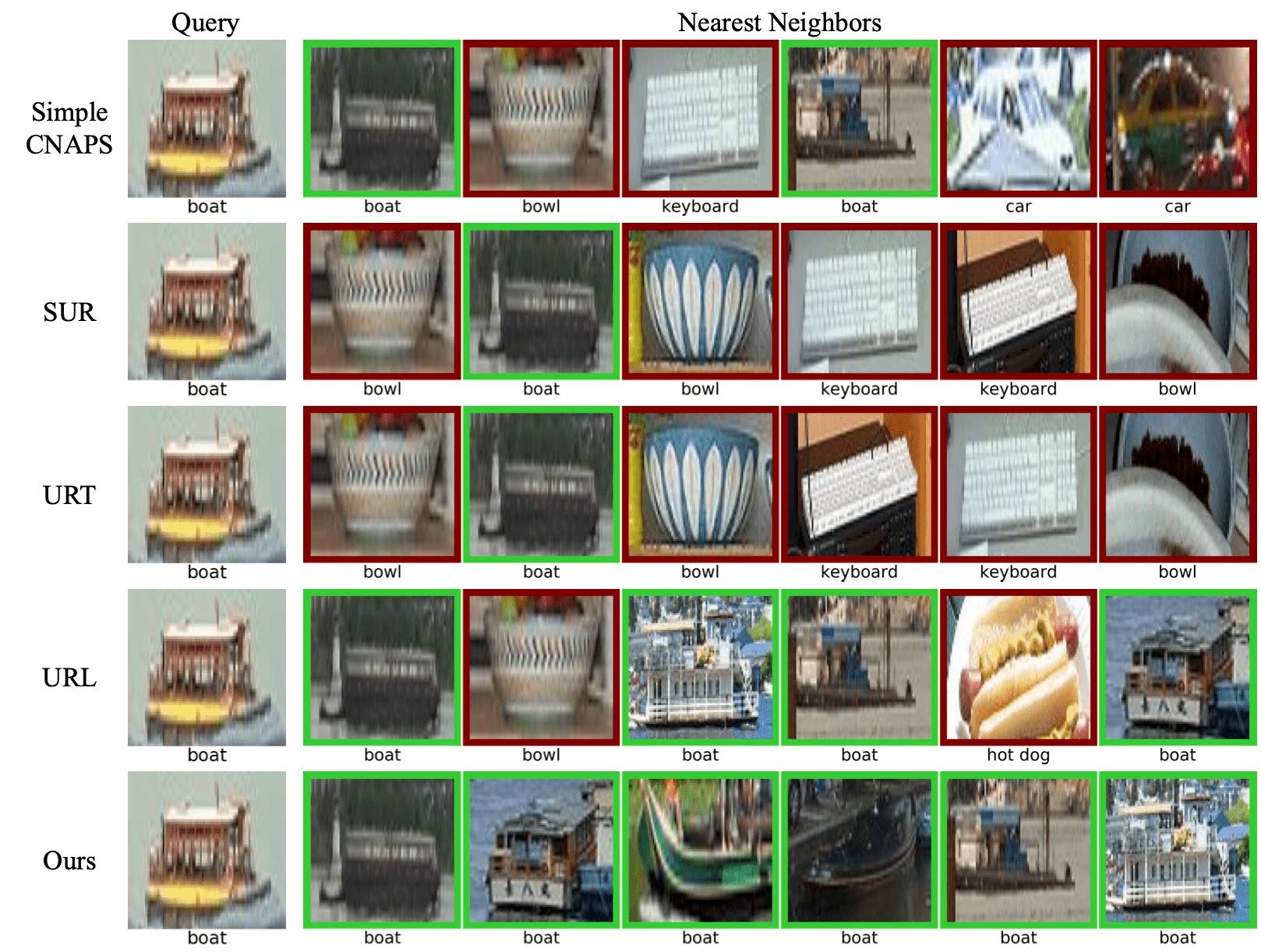}
\end{center}
\vspace{-0.3in}
\caption{Qualitative comparison to Simple CNAPS~\cite{bateni2020improved}, SUR~\cite{dvornik2020selecting}, URT~\cite{liu2020universal}, and URL~\cite{li2021universal} in MSCOCO. Green and red colors indicate correct and false predictions respectively.}
\label{suppfig:mscoco}
\end{figure}

\begin{figure}[h!]
\begin{center}
\includegraphics[width=0.9\linewidth]{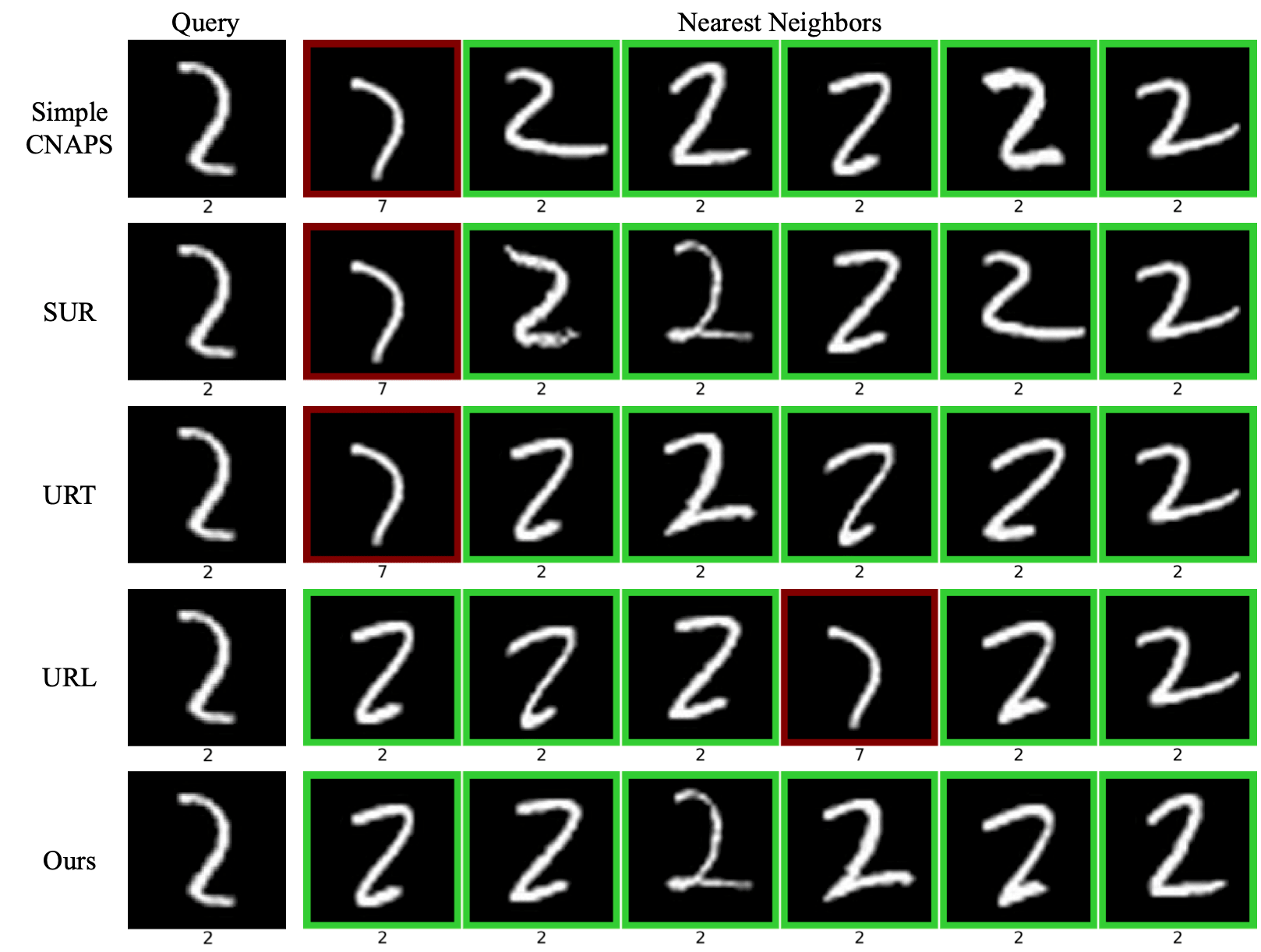}
\end{center}
\vspace{-0.3in}
\caption{Qualitative comparison to Simple CNAPS~\cite{bateni2020improved}, SUR~\cite{dvornik2020selecting}, URT~\cite{liu2020universal}, and URL~\cite{li2021universal} in MNIST. Green and red colors indicate correct and false predictions respectively.}
\label{suppfig:mnist}
\end{figure}

\begin{figure}[h!]
\begin{center}
\includegraphics[width=0.9\linewidth]{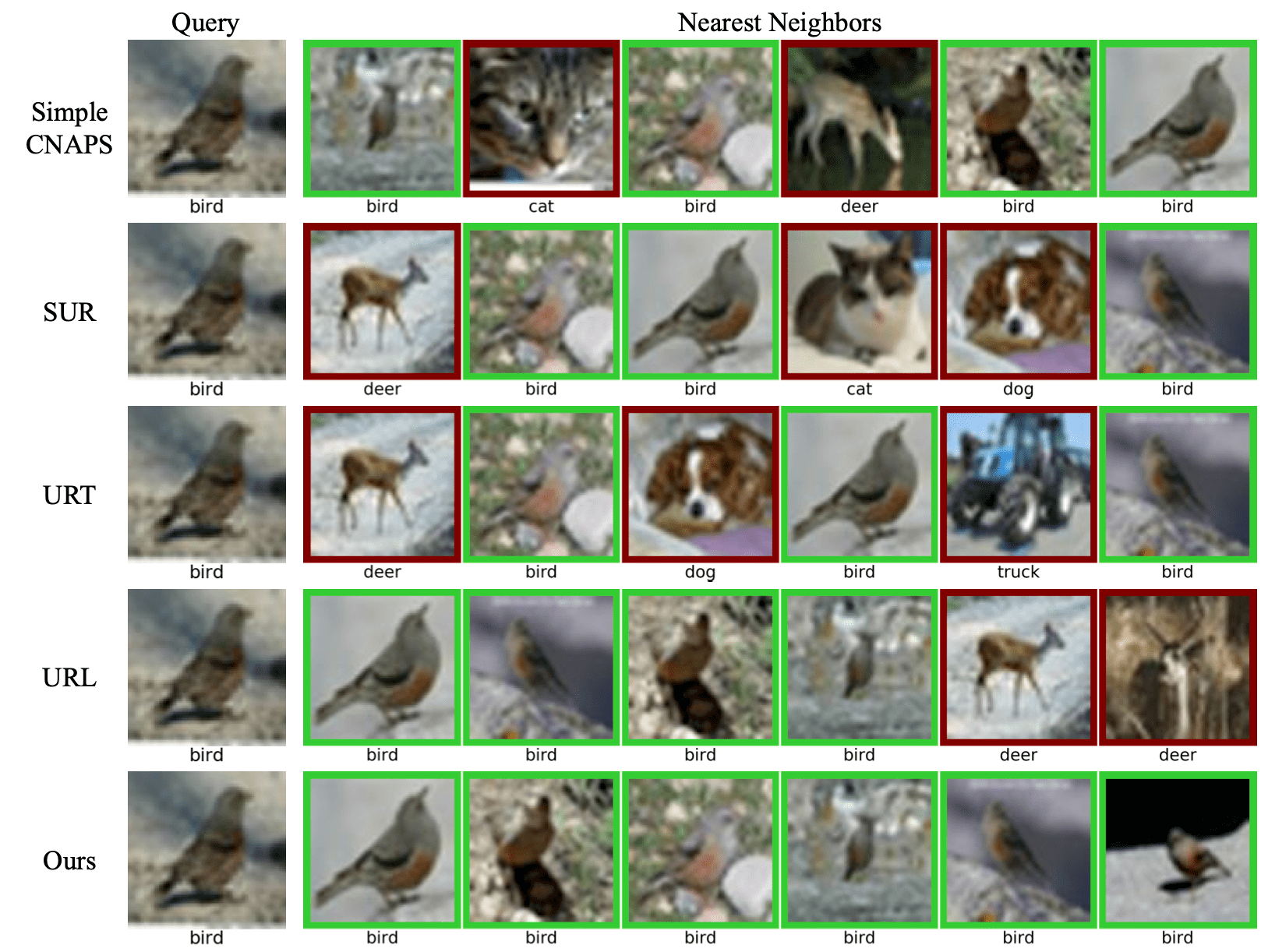}
\end{center}
\vspace{-0.3in}
\caption{Qualitative comparison to Simple CNAPS~\cite{bateni2020improved}, SUR~\cite{dvornik2020selecting}, URT~\cite{liu2020universal}, and URL~\cite{li2021universal} in CIFAR-10. Green and red colors indicate correct and false predictions respectively.}
\label{suppfig:cifar10}
\end{figure}

\begin{figure}[h!]
\begin{center}
\includegraphics[width=0.9\linewidth]{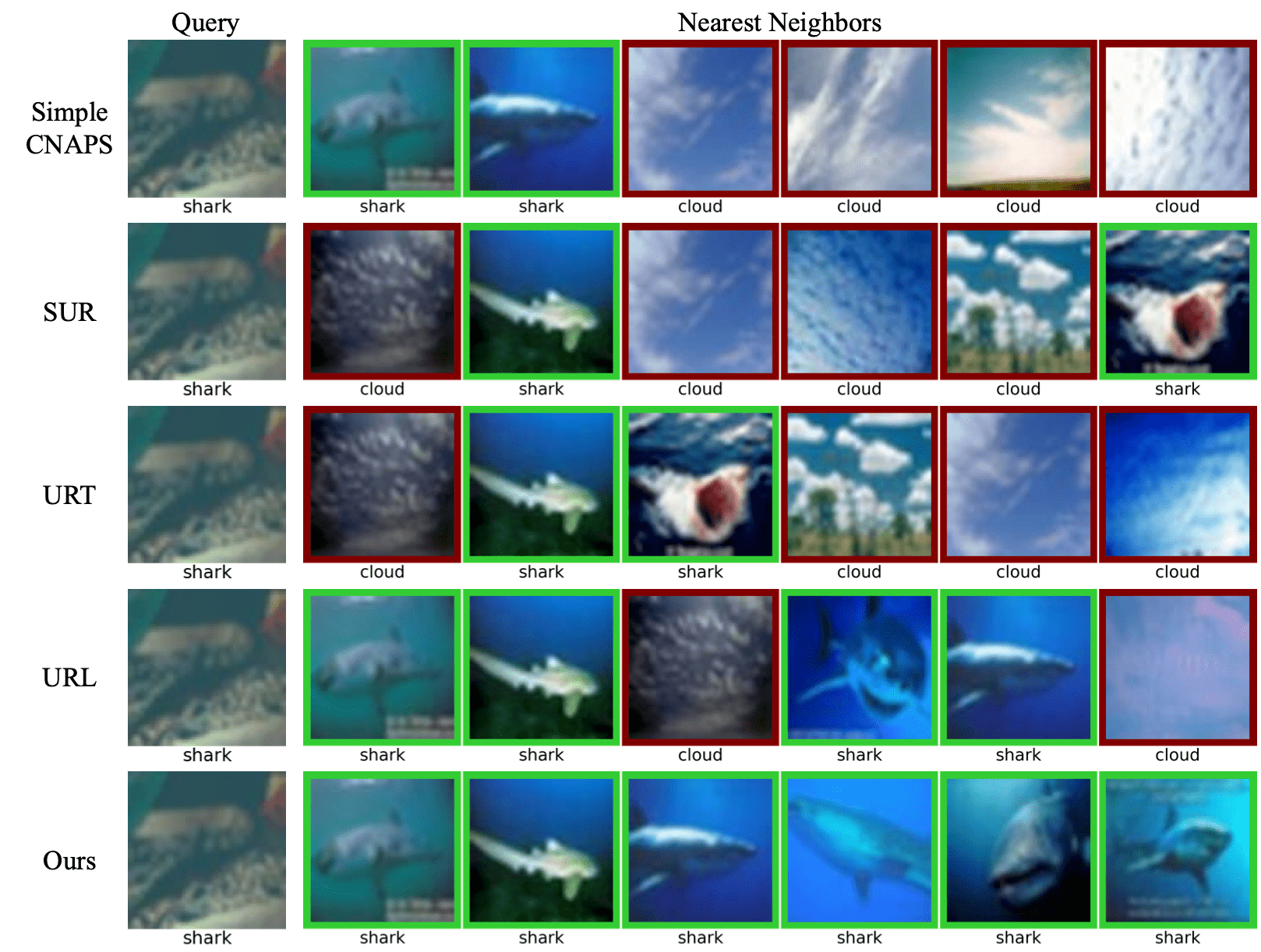}
\end{center}
\vspace{-0.3in}
\caption{Qualitative comparison to Simple CNAPS~\cite{bateni2020improved}, SUR~\cite{dvornik2020selecting}, URT~\cite{liu2020universal}, and URL~\cite{li2021universal} in CIFAR-100. Green and red colors indicate correct and false predictions respectively.}
\label{suppfig:cifar100}
\end{figure}

We qualitatively analyze our method and compare it to Simple CNAPS~\cite{bateni2020improved}, SUR~\cite{dvornik2020selecting}, URT~\cite{liu2020universal}, and URL~\cite{li2021universal} in \cref{suppfig:imagenet,suppfig:omniglot,suppfig:aircraft,suppfig:birds,suppfig:texture,suppfig:quickdraw,suppfig:fungi,suppfig:flower,suppfig:traffic,suppfig:mscoco,suppfig:mnist,suppfig:cifar10,suppfig:cifar100} by illustrating the nearest neighbors in all test datasets given a query image as in~\cite{li2021universal}.
It is clear that our method produces more correct neighbors than other methods. 
While other methods retrieve images with more similar colors, shapes and backgrounds, \eg in \cref{suppfig:traffic,suppfig:mscoco,suppfig:cifar10,suppfig:cifar100}, our method is able to retrieve semantically similar images.
More specifically, as shown in \cref{suppfig:birds}, our method correctly produces neighbors of the bird in the query image while other methods pick images with similar appearances or similar background, \eg images with twigs.
In \cref{suppfig:traffic}, other methods mainly retrieve the triangle sign while our method is able to retrieve the correct sign with illumination distortion. In \cref{suppfig:cifar100}, other methods including SUR, URT are distracted by the blue background but our method select the correct shark images. It again suggests that our method is able to quickly adapt the features for unseen few-shot tasks.

\end{document}